\documentclass{article}

\PassOptionsToPackage{numbers}{natbib}

\usepackage[final]{neurips_2022}

\usepackage[utf8]{inputenc} 
\usepackage[T1]{fontenc}   
\usepackage{hyperref}       
\usepackage{url}           
\usepackage{booktabs}      
\usepackage{amsfonts}       
\usepackage{nicefrac}      
\usepackage{microtype}      
\usepackage{xcolor}         

\usepackage{graphicx}
\usepackage{amsmath}
\usepackage{caption}   
\usepackage{subcaption}
\usepackage{enumitem} 
\usepackage{floatrow} 
\usepackage{wrapfig} 
\usepackage{algorithm}
\usepackage{algpseudocode}

\usepackage{amsthm} 
\theoremstyle{plain}
\newtheorem{thm}{Theorem}[section]
\newtheorem{lem}[thm]{Lemma}
\newtheorem{prop}[thm]{Proposition}
\newtheorem{cor}[thm]{Corollary}

\theoremstyle{definition}
\newtheorem{defn}{Definition}[section]

\theoremstyle{remark}

\title{Theory and Approximate Solvers for Branched Optimal Transport
with Multiple Sources}

\author{
  Peter Lippmann \hspace{0.7cm} 
  Enrique Fita Sanmartín \hspace{0.7cm} 
  Fred A. Hamprecht \\
  IWR at Heidelberg University, 69120 Heidelberg, Germany\\
  \texttt{\{peter.lippmann, enrique.fita.sanmartin, fred.hamprecht\}} \\
  \texttt{@iwr.uni-heidelberg.de}
}

\begin{document}

\maketitle

\begin{abstract}
Branched optimal transport (BOT) is a generalization of optimal transport in which transportation costs along an edge are subadditive. This subadditivity models an increase in transport efficiency when shipping mass along the same route, favoring branched transportation networks. We here study the \mbox{NP-hard} optimization of BOT networks connecting a finite number of sources and sinks in~$\mathbb{R}^2$. First, we show how to efficiently find the best geometry of a BOT network for many sources and sinks, given a topology. Second, we argue that a topology with more than three edges meeting at a branching point is never optimal. Third, we show that the results obtained for the Euclidean plane generalize directly to optimal transportation networks on two-dimensional Riemannian manifolds. Finally, we present a simple but effective approximate BOT solver combining geometric optimization with a combinatorial optimization of the network topology.
\end{abstract}

\section{Introduction}
\label{sec:intro}
Optimal transport (OT)~\cite{villani2009optimal,cuturi2013sinkhorn,peyre2019computational} stipulates transportation costs that increase linearly with the transported mass. However, in many systems of practical and theoretical interest, a \textit{diminishing cost} property is more realistic: it is more economic to jointly transport two loads with nearby destinations along the same route. The optimal transportation networks under diminishing costs exhibit branching; and indeed, nature and societies are using branched networks, e.g.~in blood circulation, gas supply or mail delivery. In this paper, we study the theory and practice of finding good or even optimal solutions in branched optimal transport (BOT).

More formally, we consider a finite set of \textit{sources} $S$ with supplies $\mu_S > 0$ and \textit{sinks} $T$ with demands $\mu_T > 0$, located at fixed positions $x_S$ and $x_T$ in $\mathbb{R}^2$. A possible transportation network is represented as a directed, edge-weighted graph $G(V,E)$ with nodes $V = S \cup T \cup B$. The edges $E \subset V \times V$ interconnect the terminals $S$ and $T$ with the help of a set of additional nodes $B$, so-called \textit{branching points} (BPs), with coordinates $x_B$. The edge direction indicates the direction of mass flow. The edge weights, denoted by $m_e$, specify the absolute flows. Gilbert first proposed the BOT problem~\cite{gilbert1967minimum} in which the objective is to solve for
\begin{align} \label{eq_cost}
\underset{B, E, x_B, m_E}{\arg \min} \
\sum_{(i,j) \, \in \, E} m_{ij}^\alpha &\left\|  x_i -  x_j \right\|_2 , \text{ subject to } 
\\ \nonumber
\text{supply } \ \mu_s = \sum\nolimits_{k} m_{sk}  &- \sum\nolimits_{k} m_{ks} \ \text{ at each source } s, 
\\ \nonumber
\text{demand } \ \mu_t = \sum\nolimits_{k} m_{kt}  &- \sum\nolimits_{k} m_{tk} \ \text{ at each sink } t, 
\\ \nonumber
\text{conservation } \ \sum\nolimits_{k} m_{kb}  &= \sum\nolimits_{k} m_{bk} \ \text{ at each BP } b,
\end{align} 
given a single parameter $\alpha \in [0,1]$. 
The problem of BOT is interesting in that it combines combinatorial optimization (over $B$, $E$) with continuous optimization (over $x_B$, $m_E$). \\

\begin{figure}[t]
         \captionsetup[subfigure]{justification=centering}
     \centering
     \begin{subfigure}[b]{0.22\textwidth}
         \centering
         \includegraphics[width=\textwidth]{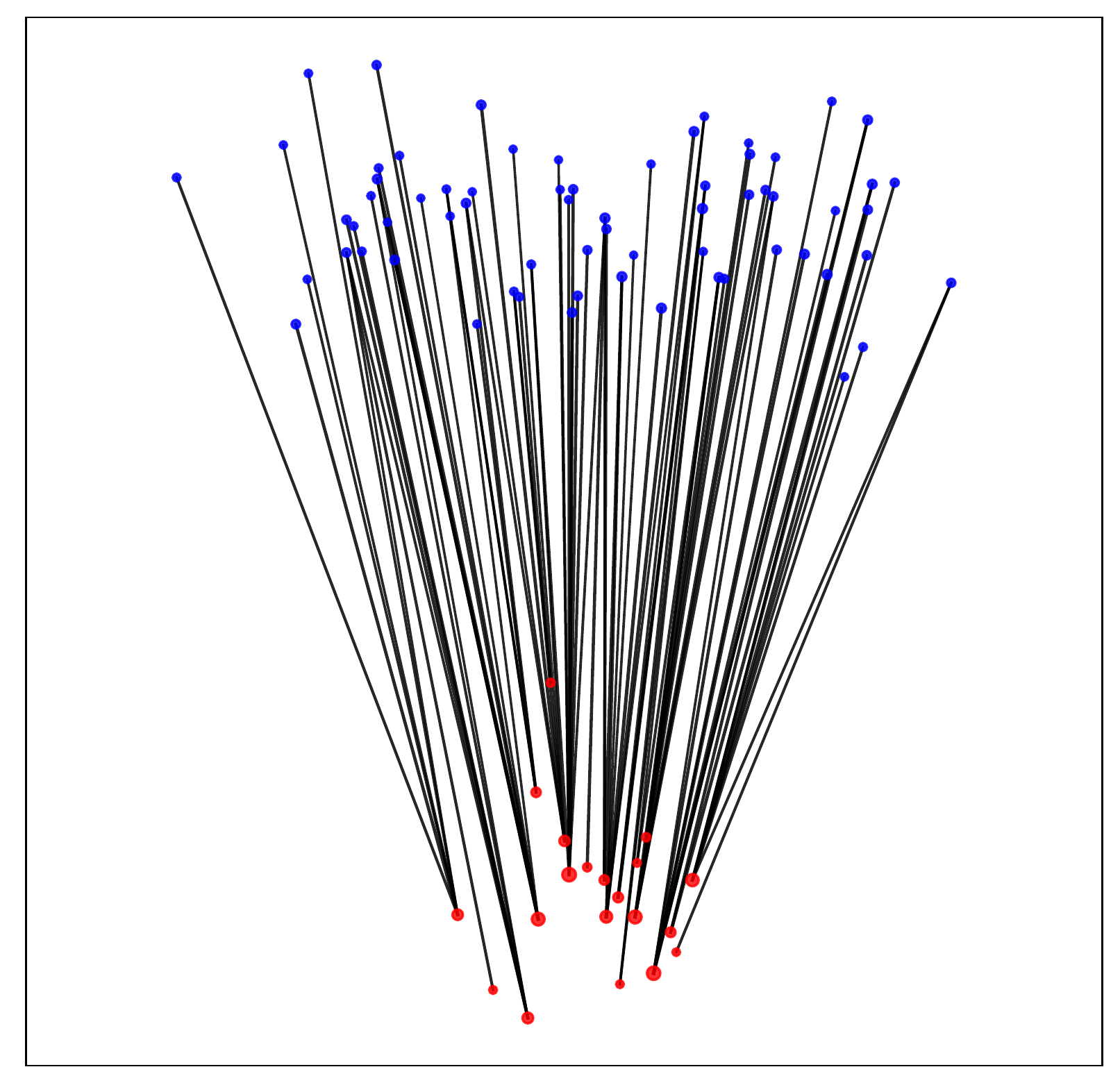}
         \captionsetup{labelformat=empty}
         \caption{$\ \alpha=1 \ $ \\ Optimal Transport}
         \label{al1}
     \end{subfigure}
     \hfill
     \begin{subfigure}[b]{0.22\textwidth}
         \centering
         \includegraphics[width=\textwidth]{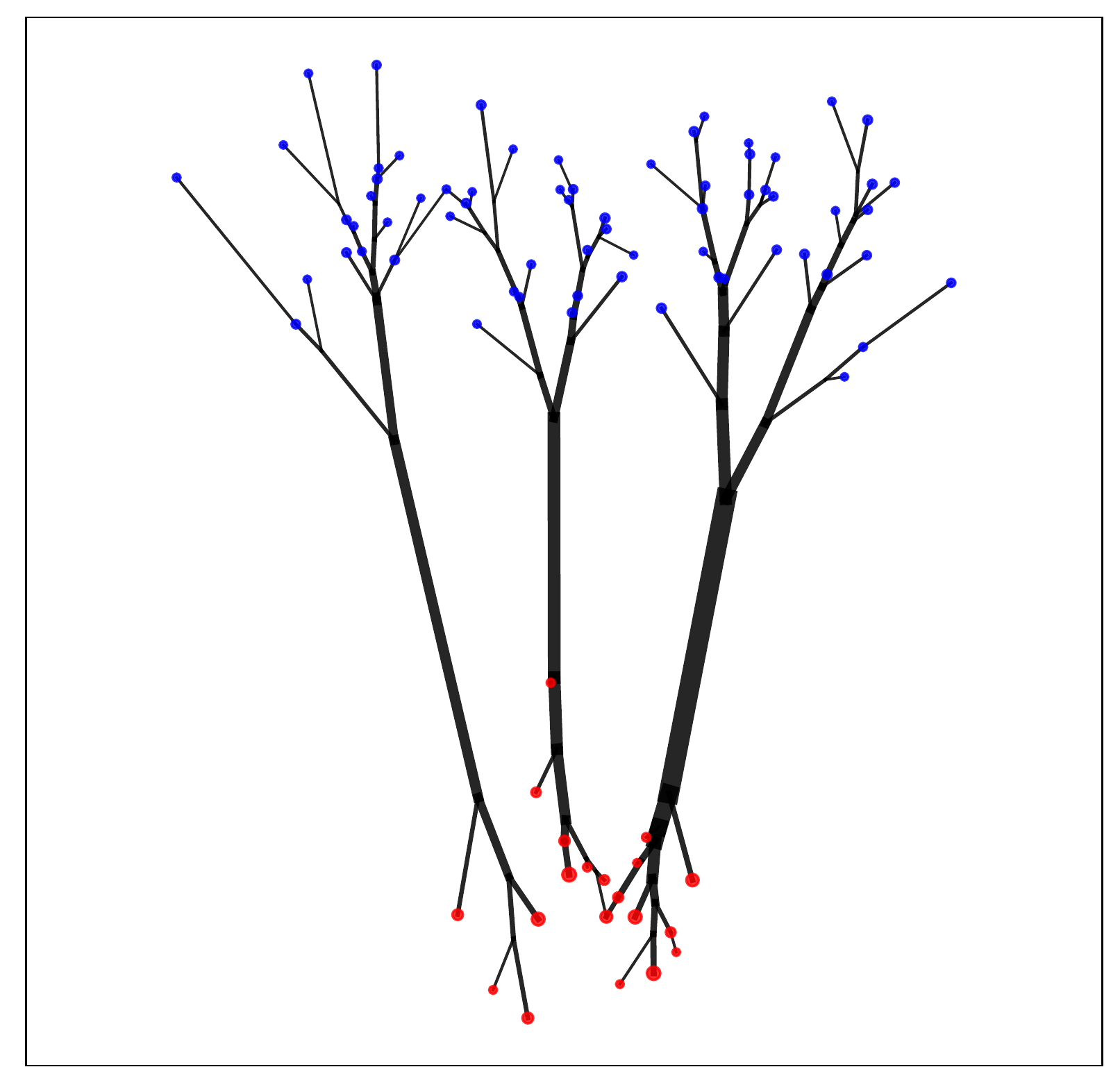}
         \captionsetup{labelformat=empty}
         \caption{$\ \alpha = 0.95 \ $ \\ \vphantom{Euclidean Steiner Tree}}
         \label{al05}
     \end{subfigure}
     \hfill
     \begin{subfigure}[b]{0.22\textwidth}
         \centering
         \includegraphics[width=\textwidth]{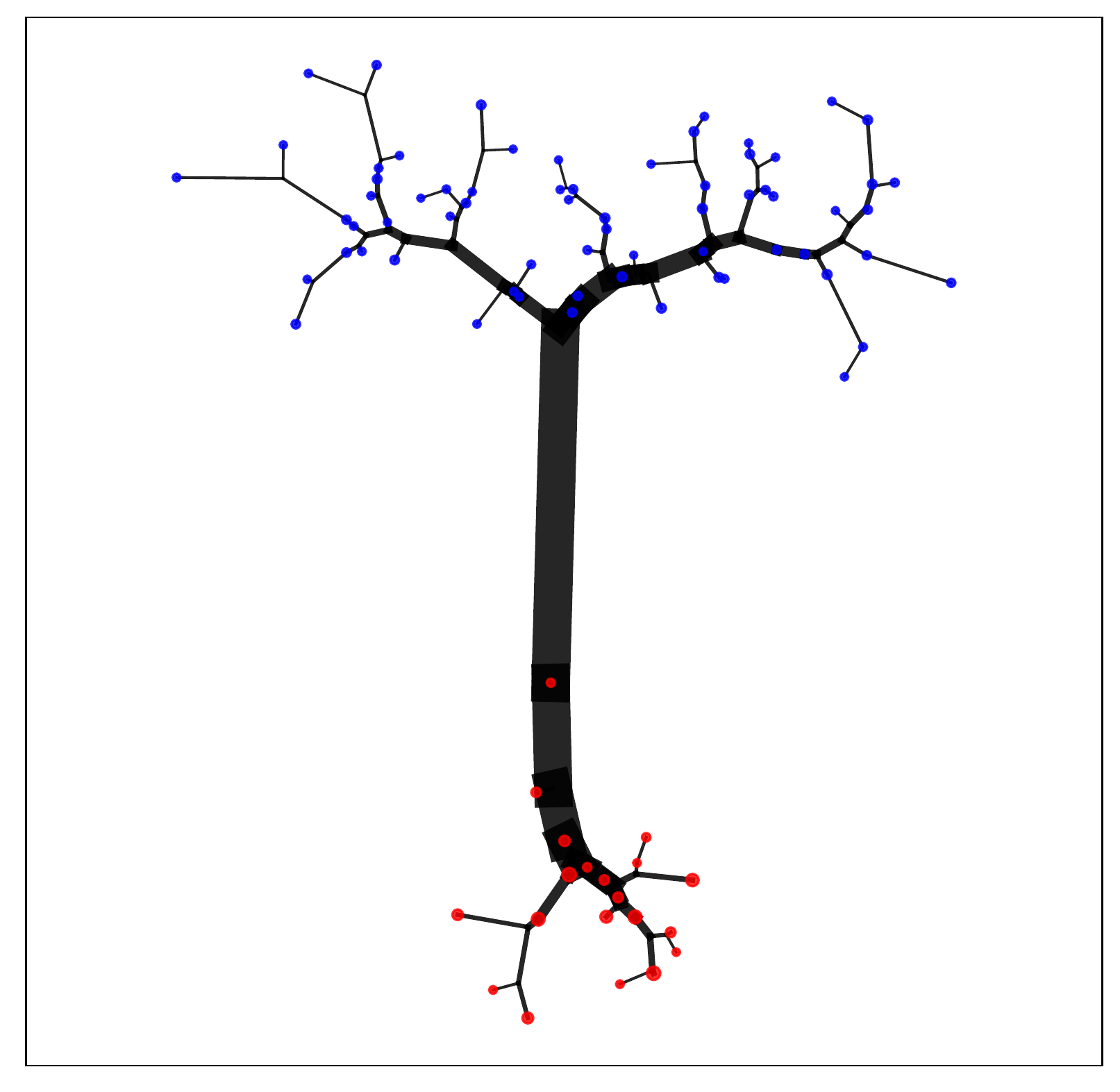}
         \captionsetup{labelformat=empty}
         \caption{$\ \alpha = 0.5 \ $ \\ \vphantom{Euclidean Steiner Tree}}
         \label{al095}
     \end{subfigure}
     \hfill
     \begin{subfigure}[b]{0.22\textwidth}
         \centering
         \includegraphics[width=\textwidth]{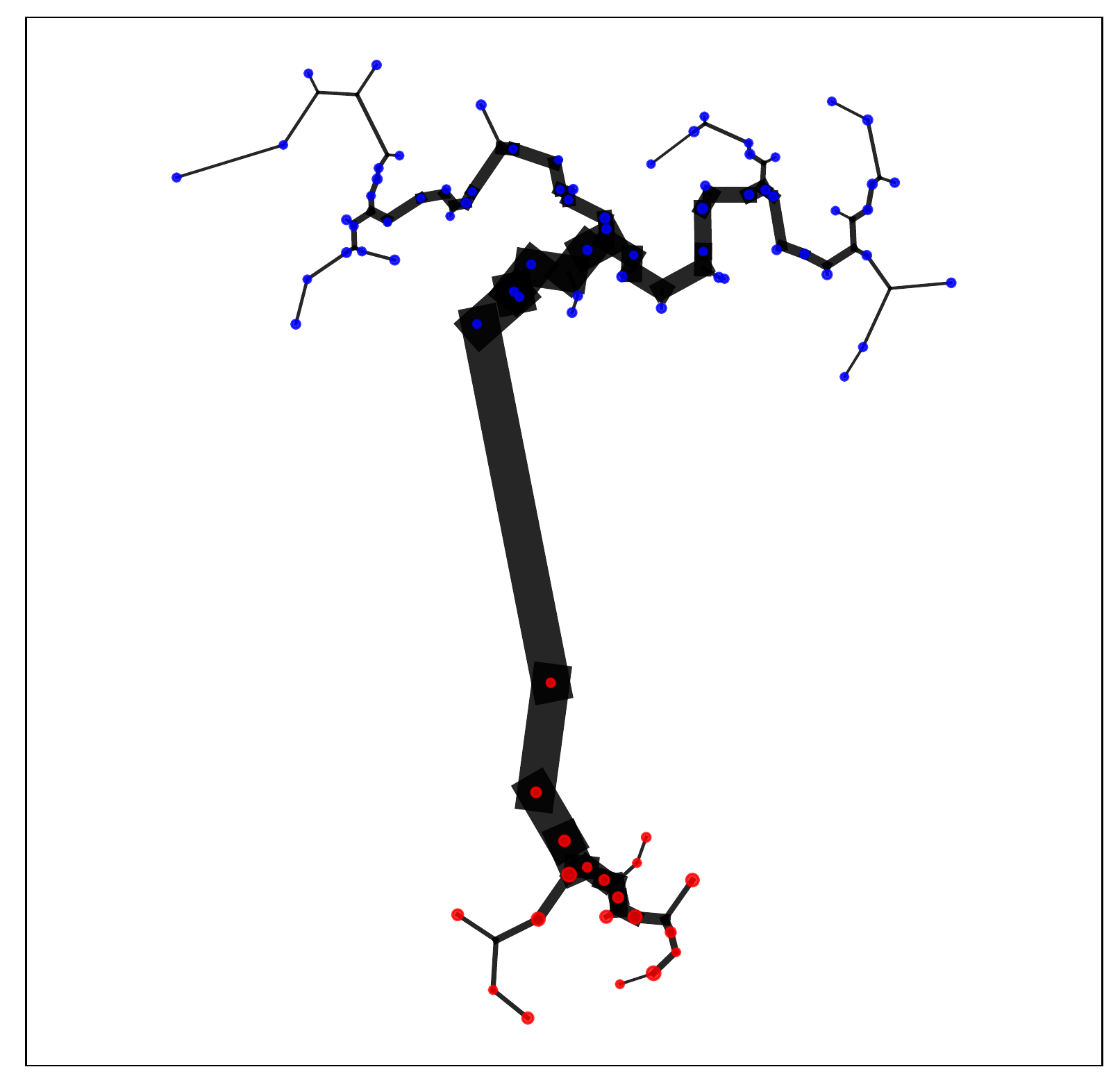}
         \captionsetup{labelformat=empty}
         \caption{$\alpha = 0 \ $ \\ Euclidean Steiner Tree }
         \label{al0}
     \end{subfigure}
        \caption{Branched optimal transport (Eq.~\ref{eq_cost}) interpolates between optimal transport and the Euclidean Steiner tree problem. On a toy example, shown are good BOT solutions (found by our approximate solver for $\alpha \neq 1$, see Sect.~\ref{sec:practical}) for the same set of sources (red) and sinks (blue). The disk sizes indicate the demands and supplies, the edge widths the mass transported along each edge.}
        \label{al_init}
\end{figure}

For $\alpha = 1$, the BOT problem is the discrete version of the famous optimal transport problem for which optimal solutions can be found efficiently~\cite{cuturi2013sinkhorn,peyre2019computational}. 
However, due to the linearity of the cost function, OT solutions do not exhibit any branching but consist of straight lines between sources and sinks, see Fig.~\ref{al1}. In contrast, for $\alpha \in [0,1)$, 
the subadditivity of $m \mapsto m^\alpha$ reflects the increased efficiency of transporting loads together, i.e.~$(m_1 + m_2)^\alpha < m_1^\alpha + m_2^\alpha$. Thus, for $\alpha \in [0,1)$, BOT solutions show a branched structure, see Fig.~\ref{al05}-\subref{al0}. Unlike OT, the optimization problem of BOT is NP-hard~\cite{guisewite1991algorithms}. In the special case of \mbox{$\alpha = 0$}, BOT turns into the well-studied Euclidean Steiner tree problem~(ESTP)~\cite{warme2000exact,hwang1992steiner}. In the ESTP, the objective is to find the overall shortest network that interconnects all terminals (with the help of BPs), independently of the edge flows, since $m^0 = 1$. For different values of $\alpha$, BOT interpolates between these two optimization problems, \mbox{see Fig.~\ref{al_init}.}

\paragraph{Connection of BOT to machine learning.} Optimal transport has emerged as an important tool in machine learning~\cite{arjovsky2017wasserstein,courty2016optimal,peyre2019computational}. BOT is a strict generalization, describing a more versatile concept and more challenging optimization problem. 

BOT offers a mathematical formalism that is deceivingly simple (cf.~Eq.~(\ref{eq_cost})) and yet engenders non-trivial structure. Many machine learning problems such as tracking of divisible targets (computer vision), skeletonization (image analysis), trajectory inference (bioinformatics) come with input that is essentially continuous (images, distributions) and require structured output that is discrete, e.g.~graphs. Arguably, this transition from continuous to discrete is one of the most interesting aspects (and an unsolved problem) in current machine learning research. It is also a problem that cannot be solved by a mere upscaling of standard deep learning architectures.

In addition, routing problems have become a popular problem to challenge machine learning and amortized optimization algorithms with difficult optimization problems~\cite{kool2018attention,cappart2021combinatorial,joshi2019efficient}. Combining combinatorial and continuous optimization, BOT is a highly instructive target for new machine learning approaches. In Sect.~\ref{sect:generalize_dim} we address the generalization of BOT to higher-dimensional Euclidean space, particularly relevant for applications in data science. \\

In this paper, 
we make the following contributions: 
We generalize an existing method for constructing BOT solutions with optimal geometry to the case of multiple sources. Based on this generalization, we present an analytical and numerical scheme to rule out $n$-degree branchings with $n > 3$. Further, we demonstrate how to extend geometric and topological properties of optimal BOT solutions to two-dimensional Riemannian manifolds. Lastly, we propose a more practical numerical algorithm for the geometry optimization together with a simple but compelling heuristic, addressing the optimization of the BOT topology. To the best of our knowledge, no readily accessible code for finding BOT solutions is publicly available. By making our code available at \url{https://github.com/hci-unihd/BranchedOT} we hope to aid the evolution of the field.

\section{Topology and geometry of BOT solutions}
\label{sec:topo-geom}
A BOT problem can be divided into the combinatorial optimization of the network topology, specified by the set of BPs $B$ and edges $E$ (see Sect.~\ref{sec:intro}), and the geometric optimization of the BP positions~$x_B$.
\citet{bernot2008optimal} showed that optimal BOT solutions can be assumed to be acyclic, which restricts the search for the optimal topology to trees. Given $n$ terminals, WLOG, the topology can be represented as a so-called \textit{full tree topology}, which has $n-2$ BPs, each of degree three. Higher-degree branchings may effectively form during the geometry optimization if multiple BPs settle at the same position. A set of such BPs is referred to as \textit{coupled} BP, cf.~Fig.~\ref{non-deg2}. The union set of all neighbors of the individual BPs (not including the BPs themselves) is referred to as set of \textit{effective neighbors}. Conversely, a BP configuration in which all BPs are uncoupled and located away from the terminals is called \textit{non-degenerate},~see Fig.~\ref{non-deg1}. 

The number of distinct full tree topologies interconnecting $n$ terminals is given by $(2n -  5)!! = (2n -  5) \cdot (2n  - 7) \cdot ... \cdot 3 \cdot 1$ and hence increases super-exponentially with the number of terminals~\cite{schroder1870vier}. Given 100 terminals, one would have to consider more than $10^{18}$ possible full tree topologies, making an exhaustive search computationally intractable already for problems of modest size. Fortunately, given a tree topology, the geometric optimization of the BP positions reduces to a convex optimization problem, as all edge flows $m_{ij}$ are already uniquely determined by the flow constraints in Eq.~(\ref{eq_cost}). The corresponding linear system can be solved in linear time by dynamic programming, called ``elimination on leaves of a tree'' in~\cite{smith1992find}.
Since the Euclidean norm, like any norm, is convex, given a fixed tree topology, the cost function in Eq.~(\ref{eq_cost}) becomes a convex function of the BP positions. 
Together with the independence of the individual BPs, this implies the following lemma on the optimal substructure of BOT solutions (see App.~\ref{sec:subopt}).

\noindent
\begin{minipage}{.51\textwidth}
\begin{defn} \label{defn:ros_gos}
For a chosen topology $T$, a BOT solution is called a \textit{relatively optimal solution}~(ROS of $T$) if its BP configuration has minimal cost. The overall best BOT solution, given by the optimal topology together with its ROS, is called the \textit{globally optimal solution}~(GOS)\label{glo-ros}.
\end{defn}   

\begin{lem} 
\label{lem:subprobs}
\textbf{(a)} For a given tree topology, a BOT solution is relatively optimal if and only if every (coupled) BP connects its (effective) neighbors at minimal cost. 
\textbf{(b)} In a globally optimal solution, every subsolution restricted to a connected subset of nodes solves its respective subproblem (cf.~App.~\ref{sec:subopt}) globally optimally.
\end{lem}
\end{minipage}%
\hfill
\begin{minipage}{.46\textwidth}
\vspace{-0.6cm}
\begin{figure}[H]
    \centering
     \begin{subfigure}[b]{0.41\textwidth}
         \centering
         \includegraphics[width=\textwidth]{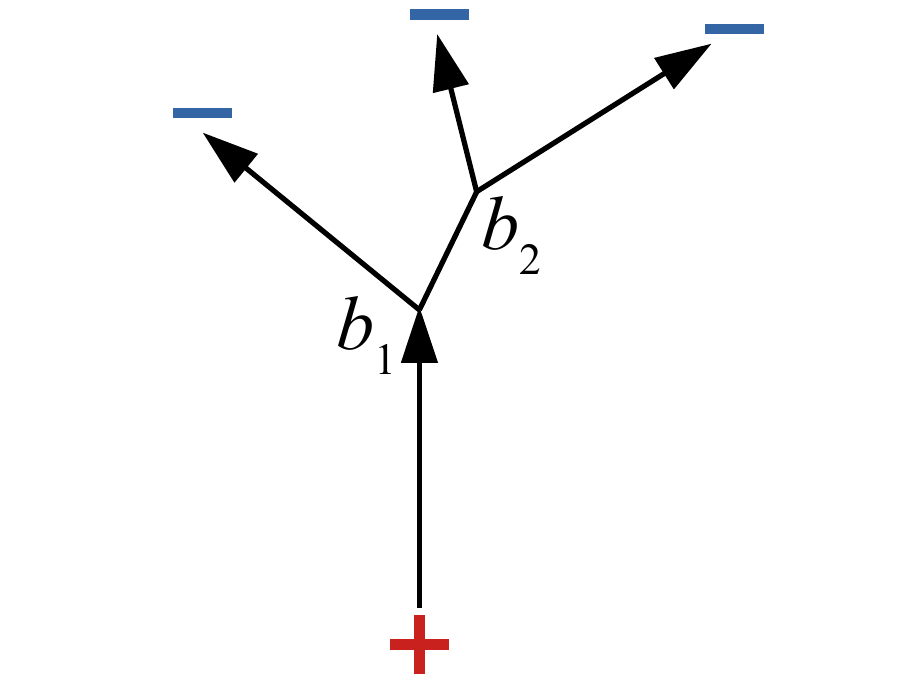}
         \caption{Non-degenerate BP configuration}
         \label{non-deg1}
     \end{subfigure}
     \hspace{0.6cm}
     \begin{subfigure}[b]{0.41\textwidth}
         \centering
         \includegraphics[width=\textwidth]{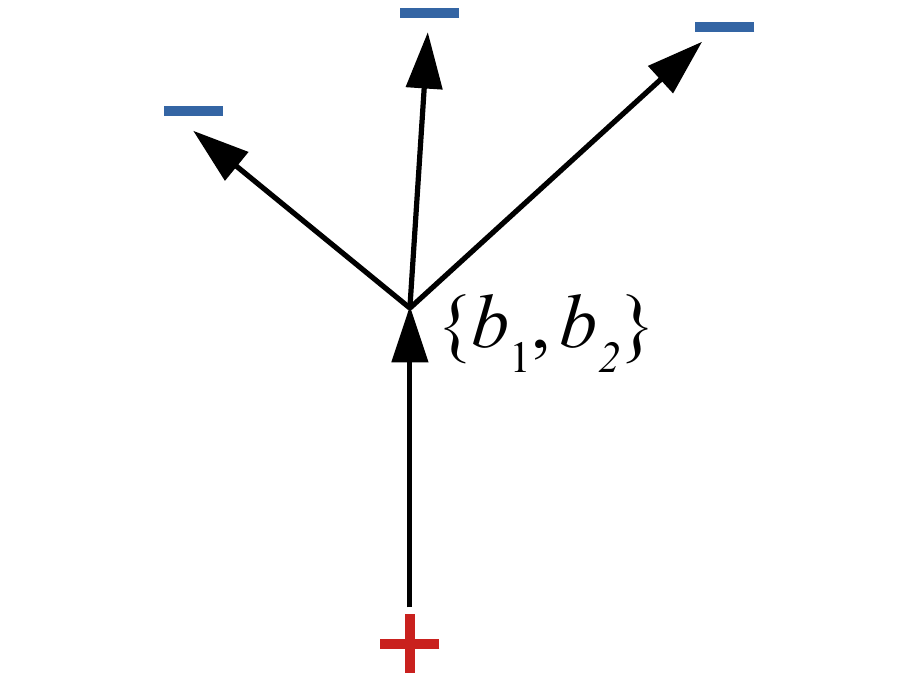}
         \caption{Coupled BP with effective degree 4}
         \label{non-deg2}
     \end{subfigure}
        \caption{Two BP configurations for the same full tree topology: A higher-degree branching may effectively form by coupling the two BPs at the same position.}
        \label{non-deg}
\end{figure}
\end{minipage}

\section{Geometric optimization of BOT solutions}
\label{sect:geom_opt}
Although the BP optimization for a given tree topology is a convex problem, as argued above, it is non-trivial, since the objective function is not everywhere differentiable. Here, we present a principled geometric approach, which was first suggested by Gilbert in~\cite{gilbert1967minimum} and previously developed in the context of the ESTP~\cite{melzak1961problem}. More recently, this approach was discussed in the comprehensive work by \citet{bernot2008optimal}, where it was applied exclusively to BOT problems with a single source. A generalization to the case of multiple sources was posed as an open problem by the authors (see Problem 15.11), for which we give the solution in this section.

\subsection{Geometric solution for one source and two sinks}
\begin{figure}
     \centering
     \begin{subfigure}[b]{0.15\textwidth}
         \centering
         \includegraphics[width=\textwidth]{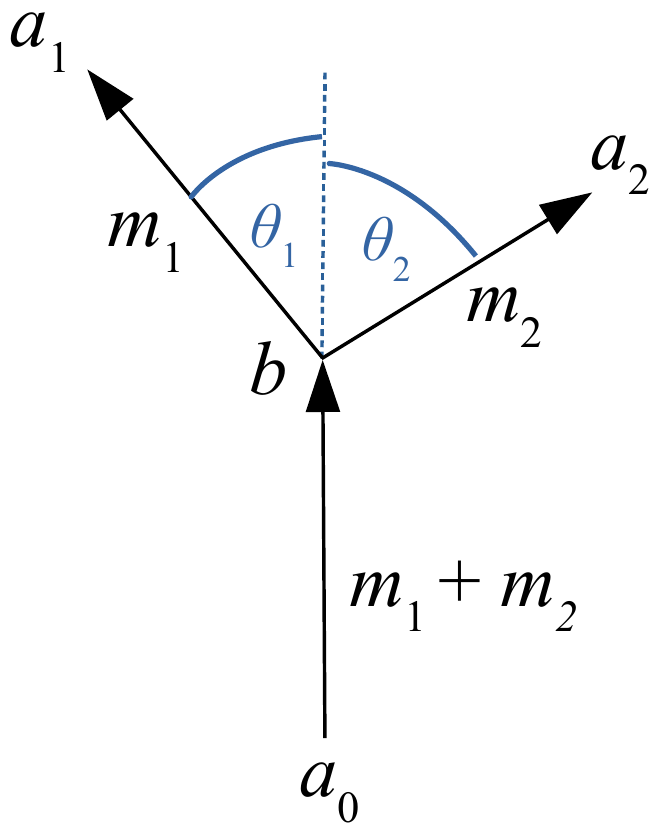}
         \caption{}
         \label{1-to-2mine}
     \end{subfigure}
     \hspace{0.8cm}
     \begin{subfigure}[b]{0.17\textwidth}
         \centering
         \includegraphics[width=\textwidth]{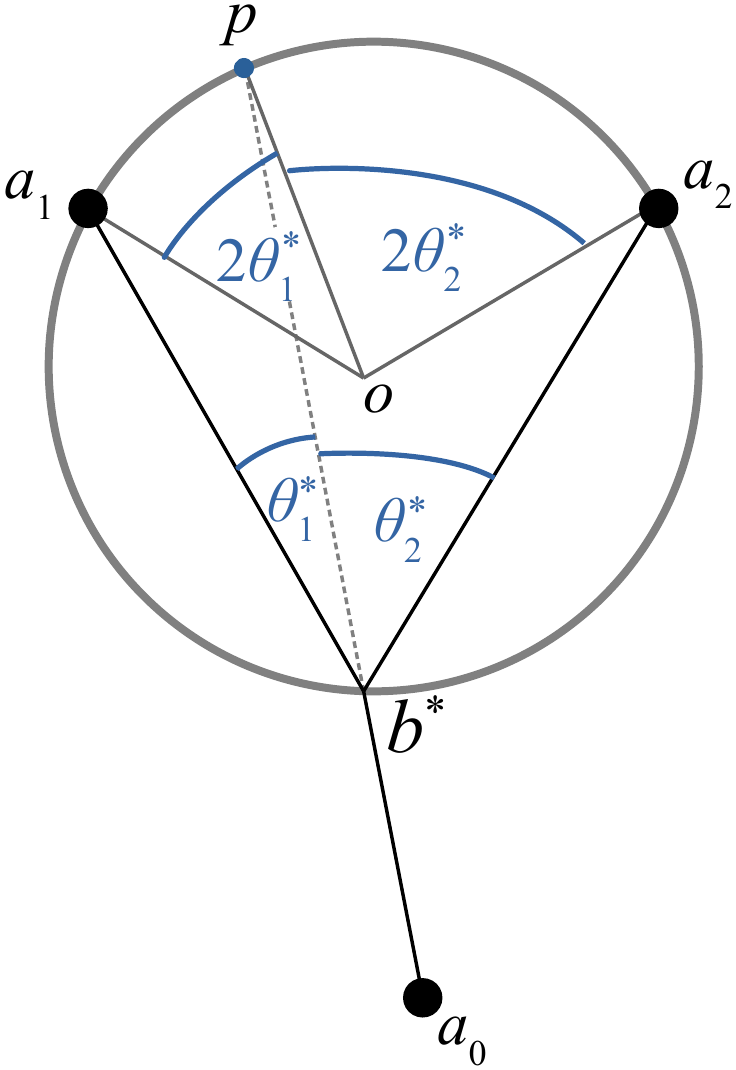}
         \caption{}
         \label{1-to-2ybranching}
     \end{subfigure}
     \hspace{0.8cm}
     \begin{subfigure}[b]{0.18\textwidth}
         \centering
         \includegraphics[width=\textwidth]{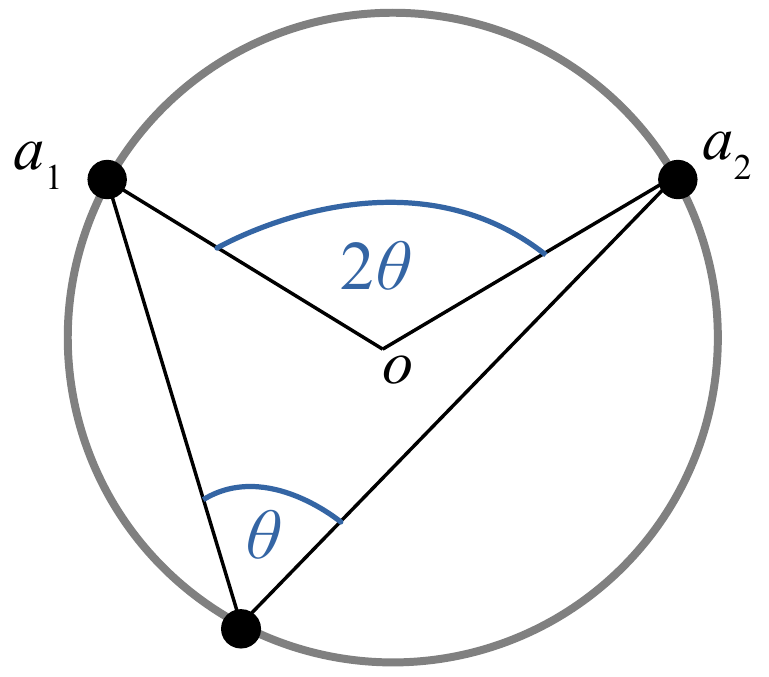}
         \caption{}
         \label{central-thm-here}
     \end{subfigure}
     \hspace{0.8cm}
     \begin{subfigure}[b]{0.16\textwidth}
         \centering
         \includegraphics[width=\textwidth]{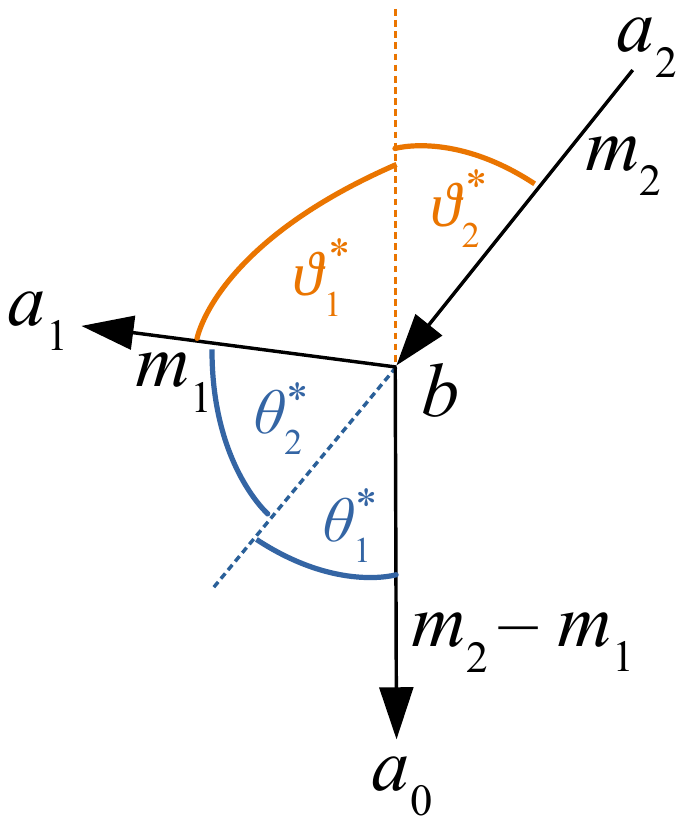}
         \caption{}
         \label{case2a}
     \end{subfigure}
        \caption{\textbf{(a)} Branching point $b$ connecting one source and two sinks with branching angles $\theta_1$ and~$\theta_2$. \textbf{(b)}~Construction of the optimal BP $b^*$ applying twice the central angle property illustrated in \textbf{(c)}. \textbf{(d)}~shows the relation of the optimal branching angles $\theta^*_i$ and $\vartheta^*_i$, relevant for the case of asymmetric branchings (see Sect.~\ref{subsec:constr}).}
        \label{1-to-2}
\end{figure}

Motivated by Lem.~\ref{lem:subprobs}, we start by considering a single BP $b$ in isolation (cf.~Fig.~\ref{1-to-2mine}), following~\cite{bernot2008optimal}. Given a source at position\footnote{We will often use the node label, e.g., $a_0$, to denote also the position of the node, instead of writing $x_{a_0}$.} 
$a_0$ and two sinks at positions $a_1$ and $a_2$, we aim to find the optimal position $b^*$ for the BP connecting the three terminals, i.e., the minimizer of 
\begin{align} \label{eq_cost1to2}
\mathcal{C}(b) = m_1^\alpha \vert a_1 - b \vert  
+ m_2^\alpha \vert a_2 - b \vert 
+ (m_1 + m_2)^\alpha \vert a_0 - b \vert, 
\end{align}
where $m_1$ and $m_2$ are the respective demands of the two sinks. Due to the convexity of $\mathcal{C}(b)$, the minimum must lie either at a stationary point at which $\nabla_{\! \!  b} \, \mathcal{C} = 0$ or at a non-differentiable point, where $b$ coincides with one of the~$a_i$.
\citeauthor{bernot2008optimal} showed that the gradient is equal to zero if and only if the branching angles $\theta_i$, see Fig.~\ref{1-to-2mine}, are given by
\begin{align}
&\theta^*_1 = \arccos \bigg( \frac{k^{2\alpha} + 1 - (1-k)^{2 \alpha }}{2 k^\alpha} \bigg) =: f(\alpha, k), \nonumber\\
&\theta^*_2 = \arccos \bigg( \frac{(1 \! - \! k)^{2\alpha} + 1 - k^{2 \alpha }}{2 (1 \! - \! k)^\alpha} \bigg) = f(\alpha, 1 \! -k), \label{eq_fandh} \\
&\theta^*_1 + \theta^*_2 = \arccos \bigg(  \frac{1 - k^{2 \alpha} - (1-k)^{2 \alpha}}{2 k^\alpha (1-k)^\alpha} \bigg) =: h(\alpha, k) \nonumber, 
\end{align}
where we have defined the flow fraction 
$k:= m_1/(m_1 + m_2)$
and the two functions $f$ and $h$, related via $h(\alpha, k) = f(\alpha, k) + f(\alpha, 1-k)$. If a BP exists that realizes the branching angles $\theta_i^*$, it can be constructed geometrically based on the \textit{central angle property}~(see App.~\ref{app-central}). It states that, given a circle through $a_1$ and $a_2$, the angle $\angle a_1 o a_2$ at the center $o$ is twice the angle enclosed with a point anywhere on the opposite circle arc, cf.~Fig.~\ref{central-thm-here}. In particular, let us construct the so-called \textit{pivot circle} with central angle $\angle a_1 o a_2 = 2\theta^*_1 + 2\theta^*_2$ and \textit{pivot point} $p$ as in Fig.~\ref{1-to-2ybranching}. Applying the central angle property twice (once for $\theta_1^*$ and once for $\theta_2^*$), a BP located at the intersection of the lower circle arc and the connection line $\overline{a_0p}$ realizes both angles $\theta^*_i$ and is therefore optimal.

However, given the pivot point and pivot circle, $\overline{a_0 p}$ may not intersect the lower circle arc, depending on the position of $a_0$. Accordingly, the lower half plane can be partitioned into a region for which the described construction yields an optimal Y-shaped branching and three other regions, see Fig.~\ref{vll1}. For $a_0$ located in one of these regions, the optimal BP position coincides with one of the terminals, resulting in a V-shaped branching ($b^* = a_0$) or an L-shaped branching ($b^* \in \{a_1,a_2\}$), cf.~Fig.~\ref{vllapp}~\cite{bernot2008optimal}. 

\subsection{Geometric construction of BOT solutions for a given topology}
\label{subsec:constr}
\begin{figure}
\centering
     \begin{subfigure}[b]{0.17\textwidth}
         \centering
         \includegraphics[width=\textwidth]{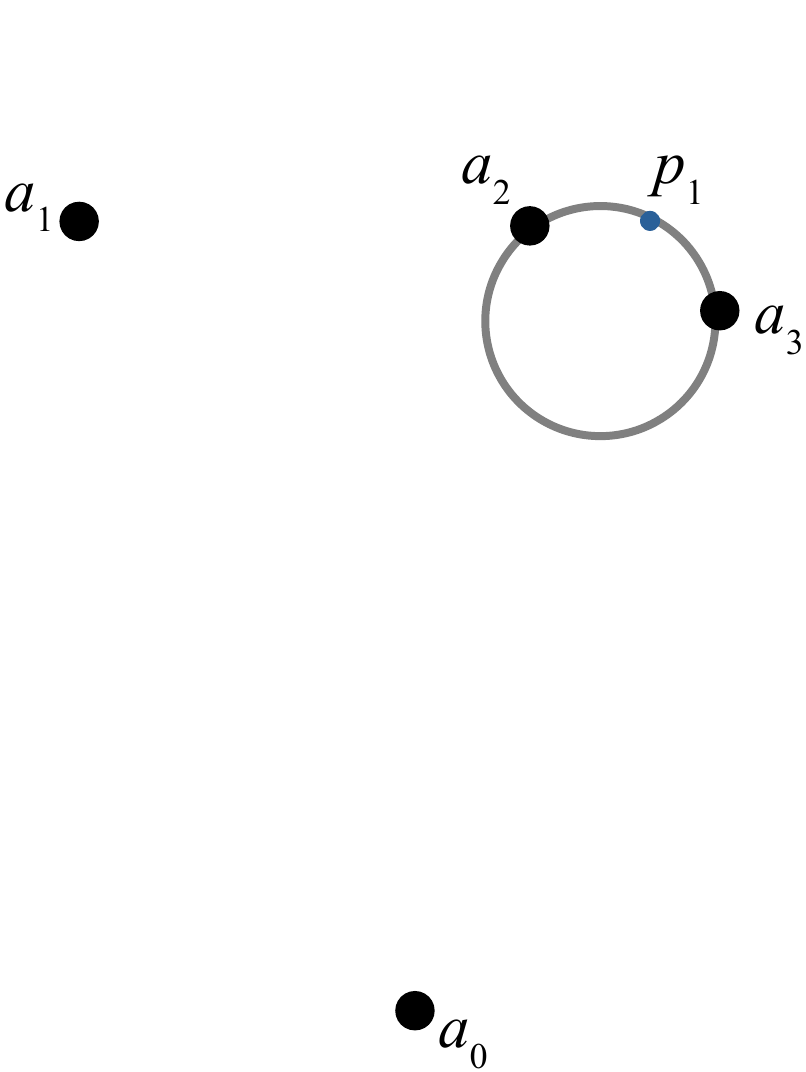}
         \caption{}
         \label{recur1}
     \end{subfigure}
     \hspace{1.cm}
     \begin{subfigure}[b]{0.17\textwidth}
         \centering
         \includegraphics[width=\textwidth]{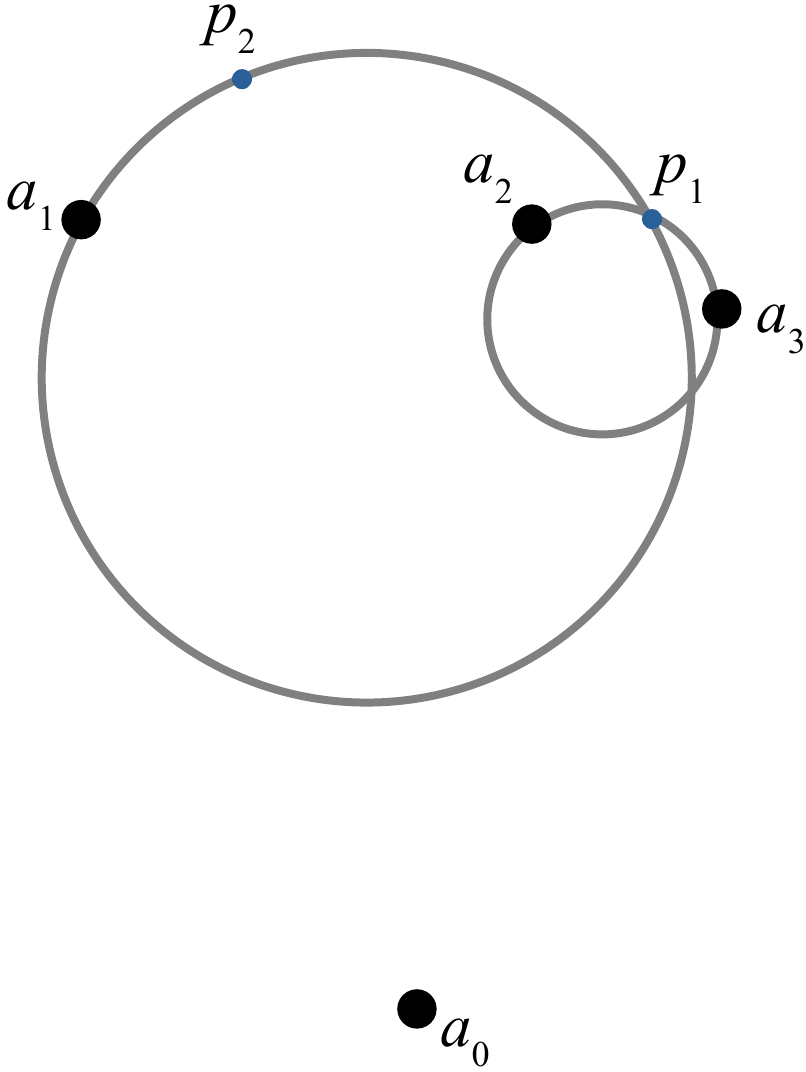}
         \caption{}
         \label{recur2}
     \end{subfigure} 
	\hspace{1.cm}
     \begin{subfigure}[b]{0.17\textwidth}
         \centering
         \includegraphics[width=\textwidth]{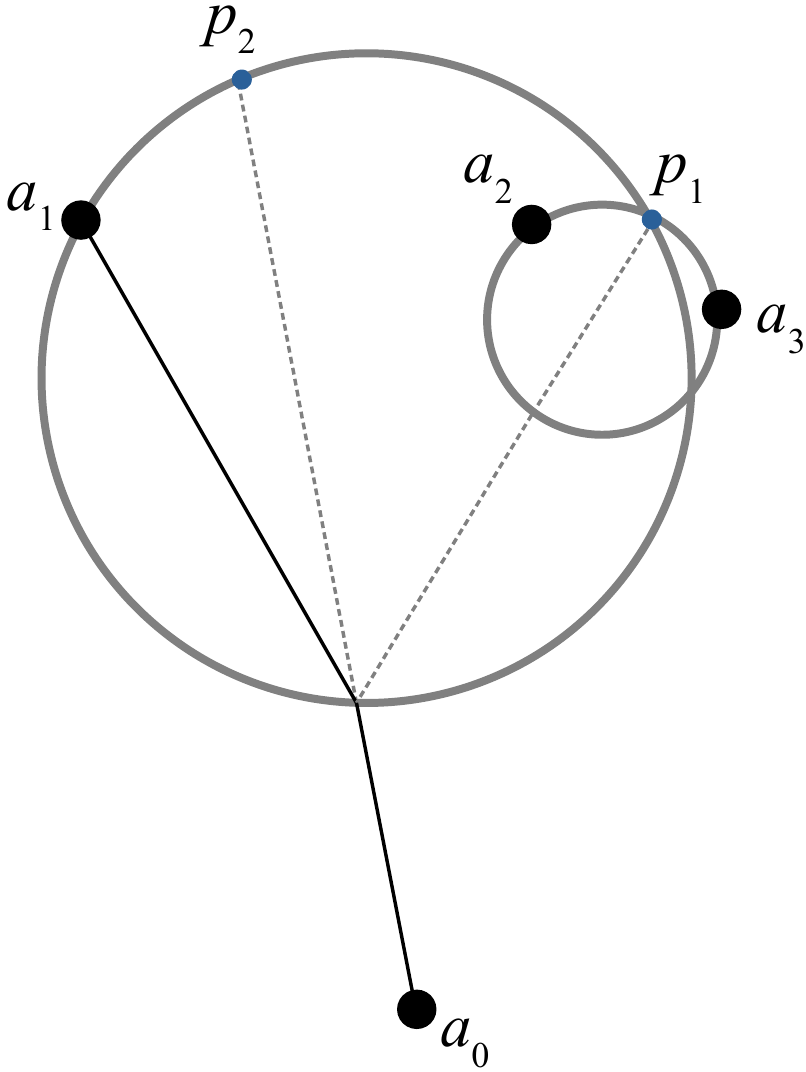}
         \caption{}
         \label{recur3}
     \end{subfigure}
     \hspace{1.cm}
     \begin{subfigure}[b]{0.17\textwidth}
         \centering
         \includegraphics[width=\textwidth]{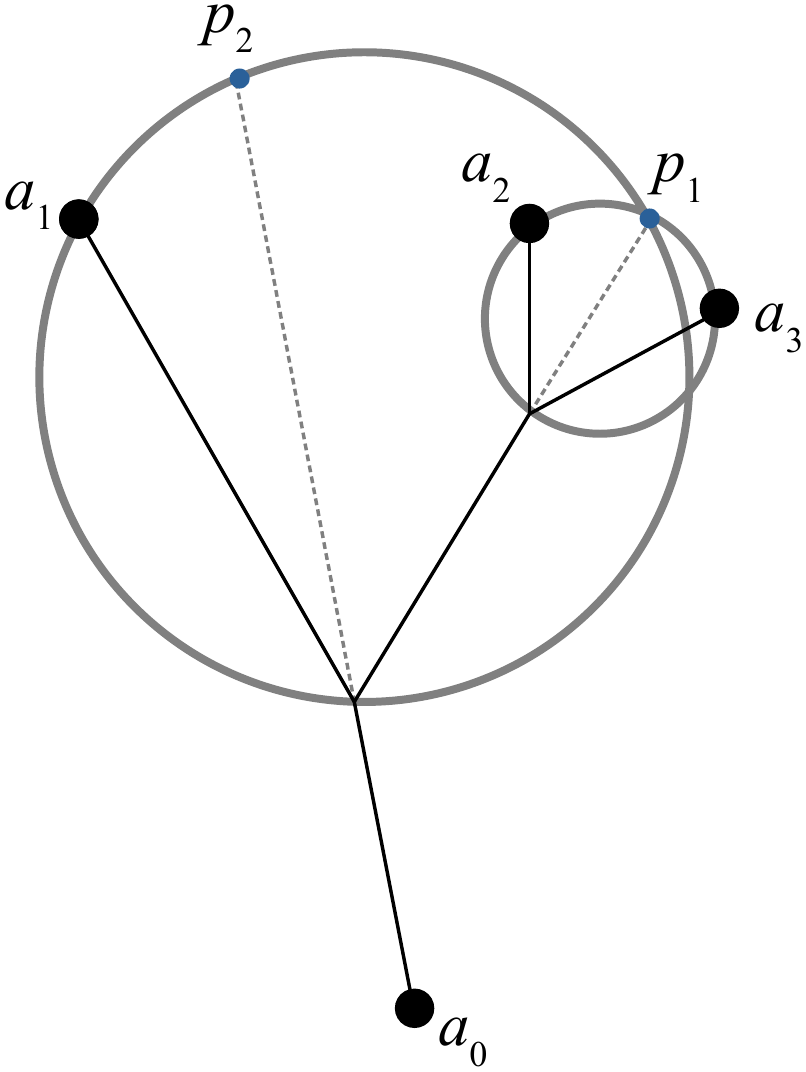}
         \caption{}
         \label{recur4}
     \end{subfigure} 
\caption{Recursive geometric construction of a relatively optimal solution, using one pivot point and pivot circle per branching point to collectively realize the optimal branching angles.}  
\label{recur}
\end{figure}

Applying the geometric construction from above in a recursive manner, one can construct the ROS~(see Def.~\ref{defn:ros_gos}) for larger BOT problems, as illustrated in Fig.~\ref{recur}. Given a full tree topology $T$, first, we determine all edge flows (see Sect.~\ref{sec:topo-geom}) and consequently the optimal branching angles. Then, a \textit{root node} is chosen, arbitrarily (here $a_0$), and all other nodes are sorted based on the number of edges to $a_0$ (ignoring edge directions and resolving ties arbitrarily). Starting from the furthest nodes and working towards the root, two nodes are recursively summarized by a pivot point, constructed from the optimal branching angles, see Fig.~\ref{recur1}-\subref{recur2}. Afterwards, in reversed order, the optimal BPs are placed iteratively, each as in the 1-to-2 case, see Fig.~\ref{recur3}-\subref{recur4}. In this manner, the optimal branching angles are realized at every BP and the resulting solution is a ROS of $T$ by Lem.~\ref{lem:subprobs}.  

The choice of the root node induces a node ordering as described above. Given this ordering, consider any BP $b$ and denote its children by $a_1$ and $a_2$ and its parent node by $a_0$. The construction of the pivot point now requires the positions of $a_1$ and $a_2$ and the optimal branching angles enclosed by the children edges $(b,a_1)$ and $(b,a_2)$. However, the branching angles do not only depend on the absolute flows $m_1$ and $m_2$ of the respective edges but also on the flow directions. Given that both flows point towards $b$ or given that both flows point away from $b$, as in Fig.~\ref{1-to-2mine}, the branching is referred to as \textit{symmetric} and the optimal branching angles of interest are given by $\theta_i^*$, cf.~Eq.~(\ref{eq_fandh}). Note that BOT problems and their solutions are fully symmetric under complete exchange of sinks and sources (up to reversal of all flow directions). On the contrary, in case of one flow pointing towards $b$ and one pointing away from $b$, referred to as \textit{asymmetric branching}, the optimal branching angles $\vartheta_i^*$ enclosed by the children edges are calculated differently, see Fig.~\ref{case2a}. However, the branching angles $\vartheta_i^*$ can be related geometrically to the known $\theta_i^*$. Using the functions $f$ and $h$ from Eq.~(\ref{eq_fandh}), we find that
\begin{equation}
\label{eq_case2_2}
\begin{aligned}
\vartheta^*_1 &= \pi - \theta^*_1 - \theta^*_2 = \pi - h\Big(\alpha, k=\frac{m_2 - m_1}{m_2}\Big)  \\ 
\vartheta^*_2 &= \theta^*_1 = f\Big(\alpha, k = \frac{m_2 - m_1}{m_2}\Big). 
\end{aligned}  
\end{equation}  
After determining the two angles $\vartheta_1^*$ and $\vartheta_2^*$ from the flows $m_1$ and $m_2$, the BP construction based on the central angle property works analogously to the symmetric case. Crucially, this distinction of symmetric and asymmetric branching makes the recursive construction applicable also to problems with multiple sources, where asymmetric branchings may be unavoidable, consider e.g.~Fig.~\ref{recur4} with $a_0$ and $a_3$ as sources and $a_1$ and $a_2$ as sinks (see App.~\ref{app:constr}). 
Further, note that the known conditions for optimal V- and L-branching can be transferred to the asymmetric case simply by relabelling $a_0 \to a_2$, $a_1 \to a_0$ and $a_2 \to a_1$, cf.~Fig.~\ref{1-to-2mine} and Fig.~\ref{case2a}. In terms of angular inequalities (derived in App.~\ref{sec_LV}), these conditions, for both branching types, are summarized in Table~\ref{tab_corner}.  

\begin{minipage}{\textwidth}
  \begin{minipage}[b]{0.37\textwidth}
    \begin{figure}[H]
    \centering 
    \includegraphics[width=\linewidth]{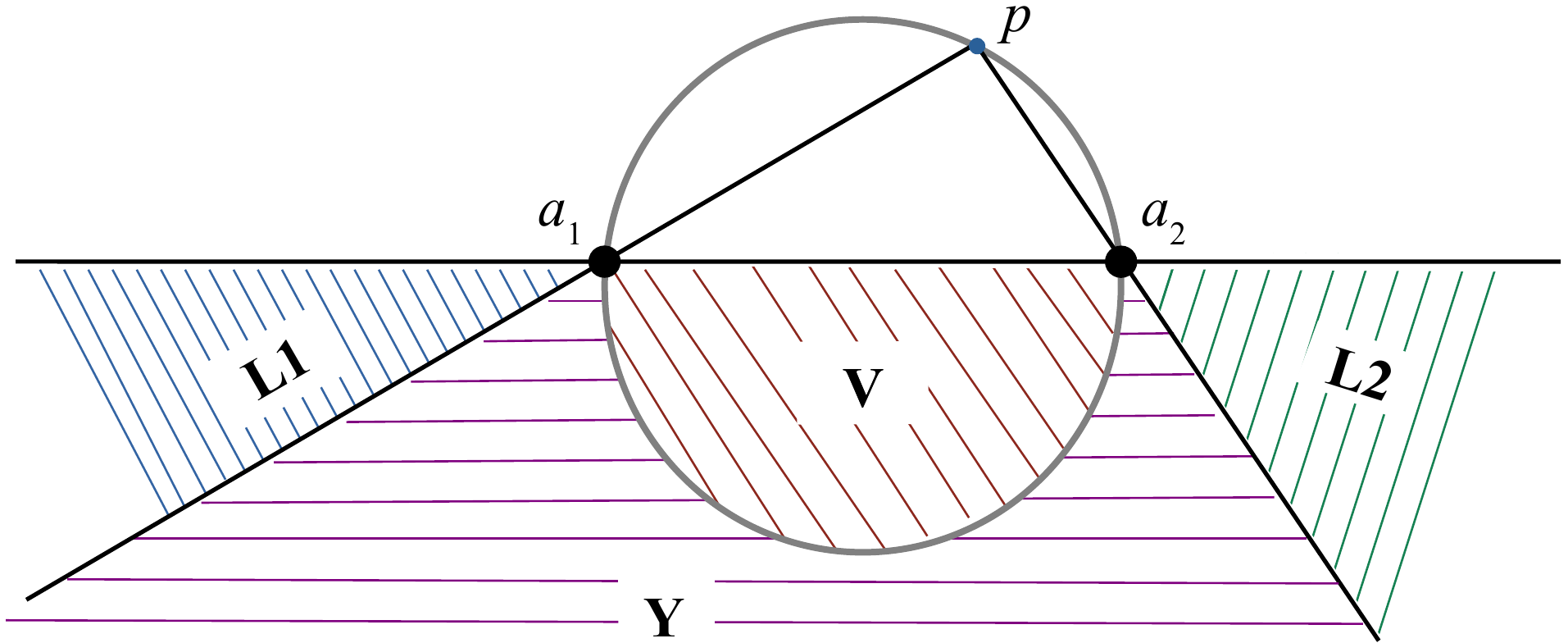} 
    \caption{Regions of optimal Y-, V- and L-branching.}  
    \label{vll1}
    \end{figure} 
  \end{minipage}
  \hfill
  \begin{minipage}[b]{0.58\textwidth}
  {\renewcommand{\arraystretch}{1.4}
    \centering
    \begin{tabular}{cc}\hline
      symmetric branching & asymmetric branching \\
     \hline 
     $\ $ V: \scalebox{1}{$\  \angle a_1a_0a_2 \ge \theta^*_1 + \theta^*_2$} & L2: \scalebox{1}{$\angle a_0a_2a_1 \ge \theta^*_1 + \theta^*_2$} \\ 
     L1: \scalebox{1}{$\angle a_2a_1a_0 \ge \pi - \theta^*_2$}  & 
     $\ $ V: \scalebox{1}{$\  \angle a_1a_0a_2 \ge \pi - \theta^*_2$} \\ 
     L2: \scalebox{1}{$\angle a_0a_2a_1 \ge \pi - \theta^*_1$} & L1: 
     \scalebox{1}{$\angle a_2a_1a_0 \ge \pi - \theta^*_1$} \\ \hline
      \end{tabular}
      \captionof{table}{Relations between the L- and V-branching conditions.}
      \label{tab_corner}
    }
    \end{minipage}
\vspace{0.3cm}
\end{minipage}

In principle, given a full tree topology, the described method efficiently constructs the ROS in linear time. However, as already pointed out by \citet{gilbert1967minimum}, the approach has some practical limitations, even after our generalization. Figure~\ref{1-to-2ybranching} shows how the pivot point is constructed only from the positions of two children $a_1$ and $a_2$ and the corresponding optimal branching angles. However, a priori there are two possible pivot point locations, one in the upper and one in the lower half plane with respect to~$\overline{a_1a_2}$. Hence, the construction relies on knowing in which half plane the third node $a_0$ lies. For larger trees, the topological parent $a_0$ may itself be a BP whose position is not yet determined. In the worst case, one would thus have to try all $2^{n-2}$ possible pivot point combinations to find the ROS. 
This pivot point degeneracy gets substantially worse in higher dimensions, making the recursive construction applicable only in $\mathbb{R}^2$. 
Secondly, the geometric construction only produces solutions which are non-degenerate, i.e., solutions without edge contractions. For now, the geometric construction is therefore primarily of theoretical interest; and indeed, it forms the basis of our following arguments. Note that both of the aforementioned problems could be solved elegantly in the special case of $\alpha = 0$~\cite{hwang1986linear,hwang1992shortest}.

\section{Properties of optimal BOT topologies}
\label{sec:topo}
Let us now consider topological modifications in order to improve the transportation cost of a BOT solution. In particular, we intend to show that a topology $T$ can be improved if its ROS contains coupled BPs. Let us start by considering a general BOT solution which contains a coupled \mbox{4-BP}, i.e., a coupled BP with four effective neighbors, as in Fig.~\ref{non-deg2}. Lemma~\ref{lem:subprobs} states that a solution is not globally optimal if any subsolution is not globally optimal. It will therefore suffice to study the coupled BP as an isolated subproblem.

\subsection{Non-optimality of coupled branching points}
\label{subsec:nonopt}
Given two sources and two sinks, there are two possible configurations in which the terminals can be arranged, cf.~Fig.~\ref{fork1},\subref{fork2}. First we address the case in which the two sources are at opposite corners of the terminal quadrilateral, as in Fig.~\ref{fork1}. Based on Lem.~\ref{lem:subprobs}, a necessary condition for the existence of a globally optimal 4-BP is that all four V-branchings between neighboring terminals are optimal. This puts a lower bound on each of the angles $\gamma_i$, see Tab.~\ref{tab_corner}. The general idea, also regarding the other 4-branching scenarios, is to show that the angular sum of these lower bounds already exceeds~$2 \pi$. This will immediately imply that not all V-branchings can be optimal simultaneously and thus a coupled 4-BP cannot be globally optimal. Given a 4-BP as in Fig.~\ref{fork1}, all V-branchings are asymmetric (i.e.~neighboring flows point in opposite directions). Hence, all four lower bounds (in Tab.~\ref{tab_corner}) are of the form $\gamma_i \ge \pi - \theta^*_2 = \pi - f(\alpha, 1  -  k)$ and indeed $\pi - f(\alpha, 1  -  k) > \pi / 2$, see Lem.~\ref{lem:monot-f-k}, so that their sum exceeds $2 \pi$. 

\noindent
\begin{minipage}{.48\textwidth}
Next, let us consider the scenario in Fig.~\ref{fork2} with two sources at neighboring corners. WLOG, we use the normalization $m_1 + m_2 = 1 = m_3 + m_4$ and assume that $m_1 > m_3$ and $m_2 < m_4$. In this case, the four conditions for optimal V-branching in Tab.~\ref{tab_corner} read: \\[0.15cm]
$\gamma_1 \ge \pi - f\Big(\alpha, 1-\frac{m_1 - m_3}{m_1} \Big) = \pi - f\Big(\alpha, \frac{m_3}{m_1} \Big)$, \\[0.15cm]
$\gamma_2 \ge h\Big(\alpha, \frac{m_1}{m_1 + m_2} \Big) = h(\alpha, m_1)$, \\[0.15cm]
$\gamma_3 \ge \! \pi  -  f\Big(\alpha, 1 \! - \frac{m_4 - m_2}{m_4}\Big) \! = \pi - f\Big(\alpha, \frac{1- m_1}{1 - m_3} \Big)$, \\[0.15cm] 
$\gamma_4 \ge h\Big(\alpha, \frac{m_3}{m_3 + m_4} \Big) = h (\alpha, m_3)$,\\[0.15cm]
\end{minipage}%
\hfill
\begin{minipage}{.48\textwidth}
\vspace{-1.3cm}
\begin{figure}[H]
     \centering
     \begin{subfigure}[b]{0.29\linewidth}
         \centering
         \includegraphics[width=\textwidth]{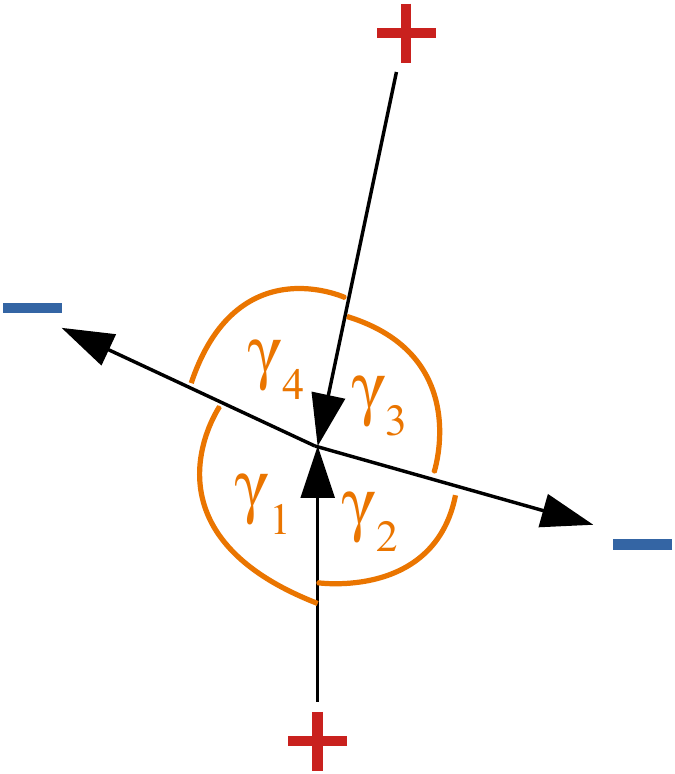}
         \caption{}
         \label{fork1}
     \end{subfigure}
     \hspace{-0.cm}
     \begin{subfigure}[b]{0.30\linewidth}
         \centering
         \includegraphics[width=\textwidth]{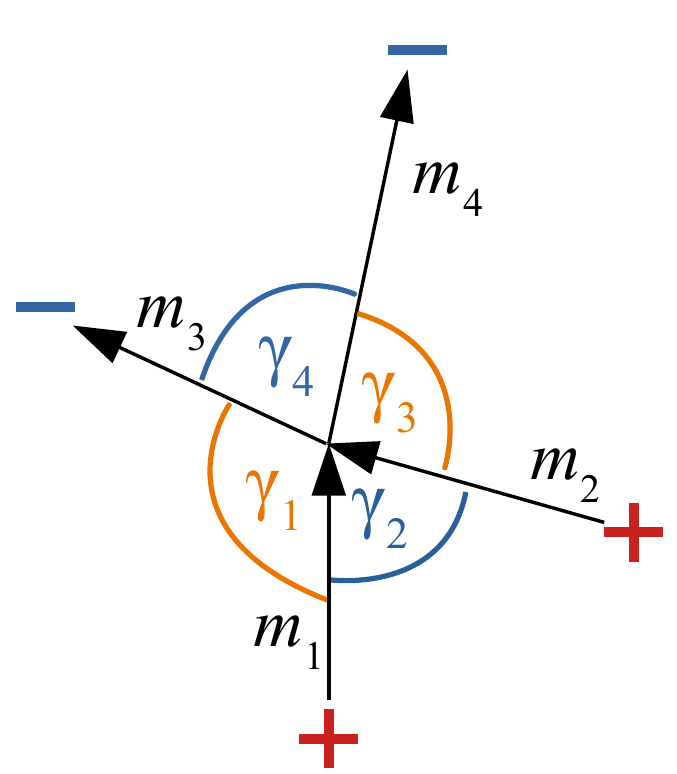}
         \caption{}
         \label{fork2}
     \end{subfigure}
     \hspace{-0.cm}
     \begin{subfigure}[b]{0.31\linewidth}
         \centering
         \includegraphics[width=\textwidth]{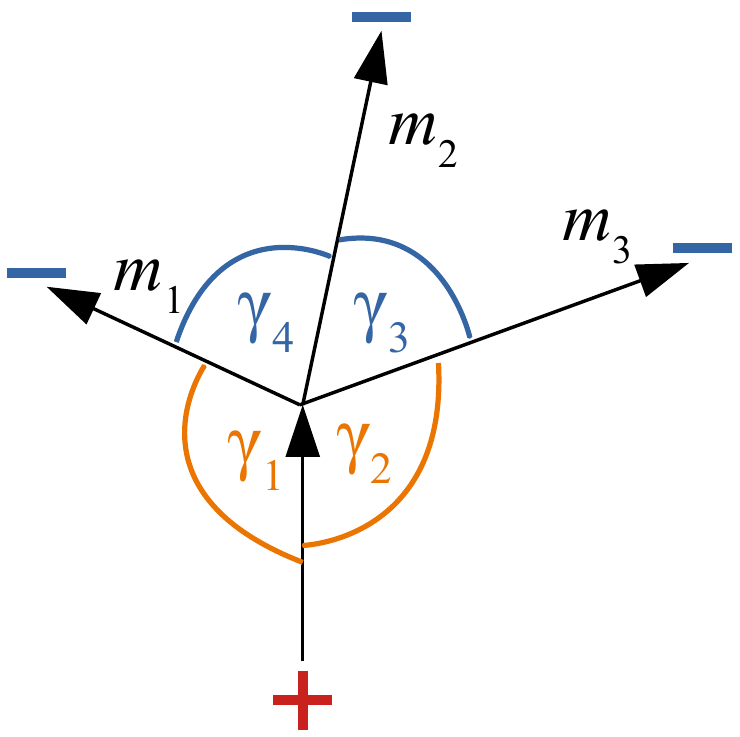}
         \caption{}
         \label{fork3}
     \end{subfigure}
        \caption{Different scenarios of coupled 4-BPs with symmetric branching angles in blue, asymmetric ones in orange.}
        \label{fork-cases}
\end{figure}
\end{minipage}
\vspace{-0.3cm}

where the expressions~(\ref{eq_fandh}) were plugged into the V-branching conditions in Tab.~\ref{tab_corner} for symmetric and asymmetric branching respectively, as indicated by the colors in Fig.~\ref{fork-cases}. Let us show that in fact for all combinations of $\alpha$, $m_1$ and $m_3$ the sum of the lower bounds already exceeds $2 \pi$. Indeed, summing the lower bounds and subtracting $2\pi$ yields 
\begin{align*}
&\underbrace{h(m_1)}_{= \ h(1-m_1)} + \ h(m_3) - f\Big(\underbrace{\frac{m_3}{m_1}}_{> \, m_3} \Big) - f\Big(\underbrace{ \frac{1 - m_1}{1 - m_3} }_{> \, 1 - m_1} \Big) >
h(1-m_1) + h(m_3) - f(m_3) - f(1-m_1)  \\
&= f(m_1) + f(1-m_3) > 0,
\end{align*}
using $h(\alpha,k) = f(\alpha, k) + f(\alpha, 1-k)$ and the fact that $f(\alpha, k)$ is strictly decreasing with respect \mbox{to $k$}, see Lem.~\ref{lem:monot-f-k}. To summarize, we have arrived at the following lemma:

\begin{lem}
A coupled 4-BP not coincident with a terminal connecting two sources and two sinks is never globally optimal.
\end{lem}%

Exactly the same logic applies for a coupled 4-BP connecting one source and three sinks (or equivalently 3 sources and 1 sink), as in Fig.~\ref{fork3}. WLOG, in the following, we normalize the flows so that $m_1 + m_2 + m_3 =1$. We then determine the necessary conditions under which all V-branchings are optimal. We again intend to show that such a 4-BP can never be globally optimal by showing that for any combination of $\alpha$ and ${m_i}$ the sum of the lower bounds exceeds $2 \pi$. This is equivalent to proving the following inequality (see App.~\ref{app:gamma_deriv}):
\begin{align*} 
h \Big(\frac{m_1}{m_1 + m_2} \Big) &- f( m_1) + h \Big(\frac{m_3}{m_3 + m_2} \Big)  -  f(m_3) > 0.  
\end{align*}
Assuming a globally optimal 4-BP existed, one could continuously displace a terminal in a way such that for the resulting BOT problem a coupled 4-BP is still globally optimal. Choosing different such displacements four additional inequalities can be derived (see App.~\ref{app:transient}): 
\begin{prop} \label{prop:inequalities}
Given a BOT problem with one source and three sinks, with demands $m_1, m_2, m_3$ as in Fig.~\ref{fork3}, a coupled 4-BP away from the terminals cannot be globally optimal if at least one of the following inequalities holds true:
\begin{align*}
\Gamma &= h \Big(\frac{m_1}{m_1 + m_2} \Big) - f(m_1) + h \Big(\frac{m_3}{m_3 + m_2} \Big) - f( m_3) > 0, \\
\Gamma_{1,*} &= f(1 -m_*) + f\Big(1 - \frac{m_2}{1-m_*} \Big) - f(1-m_*-m_2) > 0, \\
\Gamma_{2,*} &= h \Big(\frac{m_*}{m_* + m_2} \Big) + f\Big(\frac{m_2}{1 - m_*} \Big) - h(m_*) > 0 
\end{align*}
where $*=1,3$. Note that {$\Gamma = \Gamma_{1,1} + \Gamma_{2,1} = \Gamma_{1,3} + \Gamma_{2,3}$}.
\end{prop}
In App.~\ref{app:ineq_ana}, we prove the inequalities analytically for a large subset of the parameter space. For the remainder we present a numerical argument (see App.~\ref{app:numer_ineq}). In addition, we show by induction how, given that coupled 4-BPs are never globally optimal, one can further rule out coupled $n$-BPs (with $n$ effective neighbors) for all~$n > 4$.

\begin{thm} \label{thm:degree3}
Given a BOT problem in the Euclidean plane and assuming that coupled 4-BPs are never globally optimal, in a globally optimal BOT solution each branching point not coincident with a terminal must have degree three. 
\end{thm}

\section{Generalization of BOT to Riemannian manifolds}
\label{chap:mf}
In this section, we extend the BOT problem together with many of the previous results to two-dimensional Riemannian manifolds
$\mathcal M$ embedded into $\mathbb{R}^3$~\cite{lee2018introduction}. This includes the sphere as important special case, particularly relevant for global transportation networks. In the generalized BOT cost function~\eqref{eq_cost_mf} we replace the Euclidean metric by the geodesic distance $d: \mathcal{M} \times \mathcal{M} \to \mathbb{R}^+$, i.e.
\begin{align} \label{eq_cost_mf}
\mathcal{C}_M = \sum_{(i,j) \, \in \, E} m_{ij}^\alpha \, d(x_i,x_j).
\end{align} 
As we assume the manifold to be embedded, the length of a geodesics can be measured in $\mathbb{R}^3$. 
First, we generalize the non-optimality of cyclic solutions. The corresponding proof in~\cite{bernot2008optimal} readily applies also to two- and higher-dimensional manifolds. As before, solving a BOT problem on a curved surface can thus be separated into the combinatorial topology optimization and the continuous optimization of the BP configuration.

\subsection{Linear approximation of BOT solutions on manifolds}
\label{sec:approx-mf}

Intuitively speaking, a two-dimensional Riemannian manifold locally looks like the Euclidean plane. If we zoom in on a sufficiently small region, geodesics again resemble straight lines and the geodesic distance approaches the Euclidean one. This can be used to show that the branching angles which were optimal for Y-branchings in the Euclidean plane are also optimal on Riemannian manifolds. Below, we summarize the main steps of the proof. All details can be found in App.~\ref{app:mf}.

\begin{wrapfigure}{r}{0.38\textwidth}
\centering 
\vspace*{0.2cm}
\includegraphics[width=\linewidth]{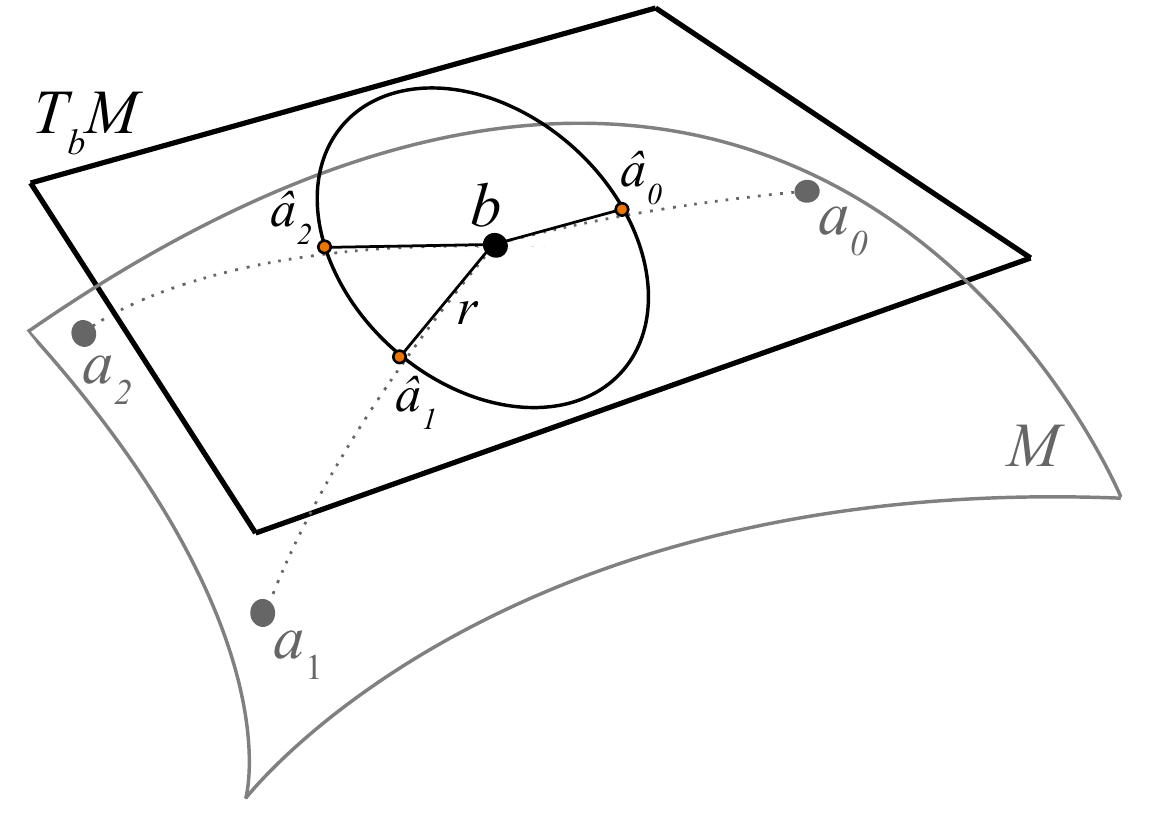} 
\caption{On the flat disk \mbox{$D(r) \subset T_b \mathcal{M}$}, the BP $b$ does \textit{not} solve the problem with terminals $\hat a_i$ optimally if the angles between the dotted geodesics are not optimal.}  \label{tangent-plane}
\end{wrapfigure}

Given a Y-branching on a manifold, we measure the angles between the three geodesics in the tangent space $T_b \mathcal{M}$ at the BP $b$.
We now zoom in on a small neighborhood $U$ around $b$ and consider only the subsolution in $\mathcal M \cap U$. The terminals of the corresponding subproblem are projected orthogonally onto the tangent space, more specifically onto a small disk of radius $r$, denoted by $D(r)$, see Fig.~\ref{tangent-plane}. Let us denote the cost of the subsolution on the manifold by $\mathcal C_M(b)$ and the cost of the corresponding subproblem in the flat disk by $\mathcal C(b)$. Now, assuming that the angles between the geodesics deviate from the optimal branching angles, the same holds true for the projected subsolution. Consequently, there exists an alternative BP $b^*$ in the disk with cheaper cost $\mathcal C(b^*)$. Note that the radius of this disk becomes smaller the smaller we choose the region $\mathcal M \cap U$ of the subproblem.

Crucially, the cost difference between a subsolution on the manifold and its projection onto the plane tends to zero quadratically in the limit of $r \to 0$. The intuitive reason for this is that the tangent space $T_b \mathcal{M}$ locally approximates the manifold to linear order. On the contrary, the costs $\mathcal C(b)$ and $\mathcal C(b^*)$ in the disk scale linearly in $r$ and so does the cost improvement \mbox{$\mathcal C(b) - \mathcal C(b^*) = M \, r$}, for some fixed $M > 0$. To conclude the proof, one projects $b^*$ onto the manifold and evaluates the cost difference of the two subsolutions there. The difference is of the form $M \, r + O(r^2)$, with second order differences due to the projection from $D(r)$ to the manifold. Consequently, a finite radius $r > 0$ must exist for which the cost difference is truly positive. A BOT solution on the manifold for which the Y-branching angles deviate from the optimal branching angles can thus be improved and is not relatively optimal. The logic of the proof outlined here can easily be extended to the V- and L-branching conditions as well as our results regarding the non-optimality of coupled BPs. 
Again, improving the BOT solution locally in the tangent plane (w.r.t.~its geometry or topology) and projecting back to the manifold results in an improved solution on the manifold (see App.~\ref{app:mf_other}).  

\begin{thm} \label{thm-mf-ang}
Consider the solution to a generalized BOT problem on a two-dimensional Riemannian manifold embedded into $\mathbb{R}^3$. For the solution to be relatively optimal, it is a necessary condition that each BP satisfies the optimal angle conditions for Y-, V- and L-branching, which apply for BOT in the Euclidean plane. For it to be globally optimal, assuming that coupled 4-BPs are not optimal in the plane, it is a necessary condition that BPs not coincident with a terminal have degree three.    
\end{thm}

Though there is no readily available algorithm to solve BOT on embedded surfaces, we discuss some possible approaches in App.~\ref{app:mf_practical}.

\section{Heuristics and numerical optimization}
\label{sec:practical}
In this section, we present a simple but effective algorithm for the geometry optimization, followed by a compelling heuristic for the topology optimization. 
As pointed out earlier, the difficulty of solving a BOT problem stems from the super-exponentially growing number of possible full tree topologies. Obtaining an exact solution by brute-force is almost always computationally infeasible and hence fast heuristic solvers are needed. For BOT problems with a single source, a branch-and-bound method is applicable~\cite{xue1999computing}, enabling exact solutions for up to 16 nodes. However, this method does not generalize directly to the case of multiple sources. While some literature exists on heuristics for BOT problems with a single source~\cite{xia2010numerical}, we are not aware of heuristics for multiple sources, except~\cite{piersa2014ramification}. The authors of~\cite{piersa2014ramification} present a simulated annealing based optimization strategy for BOT, based on hand-crafted geometrical and topological modifications, which may require user supervision. Furthermore, continuous approaches to solve BOT exist which do not rely on a subdivision into geometry and topology optimization. The authors of~\cite{oudet2011modica} phrase BOT as a limit of functional minimization problems. Since their algorithm discretizes the plane and the BOT cost function, their output is however not sparse but a discretized function.   

\subsection{Numerical branching point optimization for a given topology} 
\label{sec:smith} 
Brute-force and heuristic BOT solvers alike typically rely on the geometry optimization of many different topologies. A fast and reliable BP optimization routine is therefore essential, as it determines the computational bottleneck of these algorithms. 
For a given tree topology $T$, all edge flows $m_{ij}$ are known (see Sect.~\ref{sec:topo-geom}). The objective is thus to minimize the following convex cost function: 
\begin{align} \label{eq_cost_smith}
\mathcal{C}(\{x_i\}) = \sum_{(i,j) \, \in \, T} m_{ij}^\alpha 
\left \| x_i - x_j  \right \|_2,
\end{align} 
where, for $1 \le i \le n$, the $x_i$ hold the fixed coordinates of the terminals and, for $n+1 \le i \le n+m$, the variable BP positions. Since the cost function is not everywhere differentiable, we suggest the following generalization of Smith's algorithm developed for geometry optimization in the ESTP~\cite{smith1992find}. It is an effective algorithm specifically for minimizing the sum of Euclidean norms in two- and higher-dimensional Euclidean space. Unlike the geometric construction in Section~\ref{subsec:constr}, it is applicable to all (not necessarily full) tree topologies.

Starting from a non-optimal, non-degenerate BP configuration, e.g.~from a random initialization, the gradient with respect to each BP position $x_i$ is set to zero for $n+1 \le i \le n+m$, resulting in the following non-linear system of $m$ equations:
\begin{align*}
x_i = \sum_{j \, : \, (i,j) \in T} m_{ij}^\alpha 
\frac{x_j}{\vert x_i - x_j \vert} 
 \  \Bigg  /  \sum_{j \, : \, (i,j) \in T} \frac{m_{ij}^\alpha}{\vert x_i - x_j \vert}.
\end{align*}
This system can be solved approximately, by iteratively solving the following \textit{linearized} system
\begin{align} \label{eq:iter}
&x_i^{(k + 1)} = \!  \! \sum_{j \, : \, (i,j) \in T}  m_{ij}^\alpha 
\frac{x_j^{(k+1)}}{\vert x_i^{(k)} - x_j^{(k)}  \vert} 
 \  \Bigg  /  \sum_{j \, : \, (i,j) \in T}  \frac{m_{ij}^\alpha}{\vert x_i^{(k)} - x_j^{(k)}  \vert}, \hspace{0.4cm}\text{for } n+1 \le i \le n+m.
\end{align} 
Note that $x_i^{(k)} = x_i$ is fixed for $1 \le i \le n$.
For each iteration, the solution can be found in linear time, again by ``elimination on leaves of a tree'', similar to determining all edge flows from the flow constraints. The algorithm is easily parallelized over $d$ spatial dimensions of a BOT problem so that a single iteration is of order $O(nd)$. In essence, this is an iteratively reweighted least squares (IRLS) approach~\cite{chartrand2008iteratively}. The connection is made explicit in App.~\ref{app:smith}. 
Details on the proof of convergence, the empirical runtime of the algorithm and suitable convergence criteria can be found in App.~\ref{app:smith} and in~\cite{smith1992find}.
The arguments in~\cite{smith1992find} readily apply to our generalization. Besides our method, other techniques may be used for the geometry optimization, for instance the interior point method presented in~\cite{xue1997efficient}.

\subsection{A greedy randomized algorithm for the topology optimization}
\label{sec:sim-ann}

\begin{wrapfigure}{r}{0.42\textwidth}
\centering 
\includegraphics[width=\linewidth]{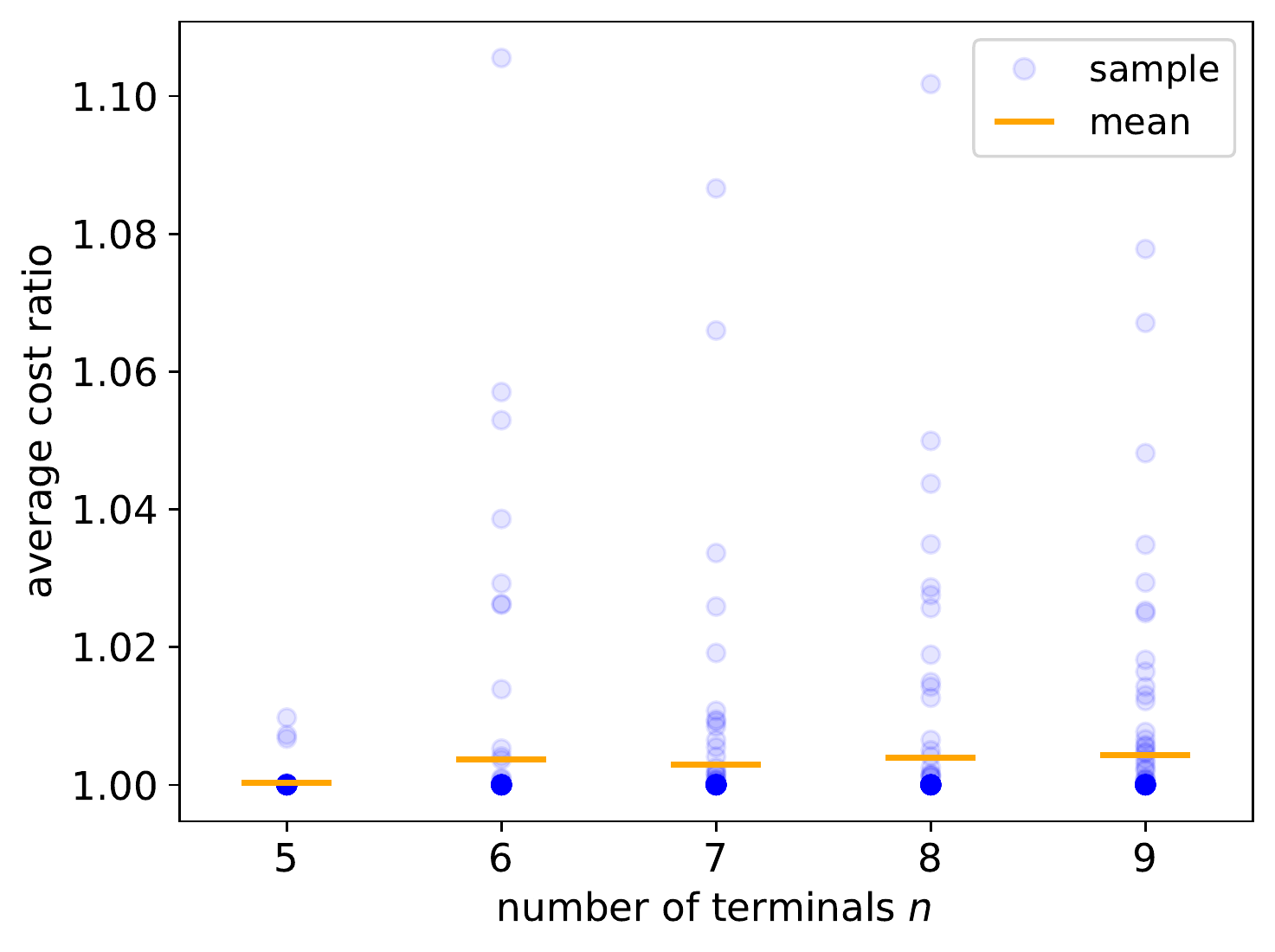} 
\caption{Cost ratios of our greedy heuristic and brute-force solutions (the closer to 1 the better) for different number of terminals $n$. For each $n$, we uniformly sampled 100 different BOT problems. Most runs ended up close to the global optimum (see dark blue assembly near~1.0).}  
\label{MCratio}
\end{wrapfigure}

Our heuristic for the optimization of the BOT topology is inspired by the idea of simulated annealing~\cite{kirkpatrick1983optimization}, which has been applied in different variants to combinatorial problems such as the Traveling Salesman Problem~\cite{malek1989serial} or the ESTP~\cite{grimwood1994euclidean}. 
In our heuristic, the BOT topology is iteratively modified by randomly deleting an edge and replacing it with a new one. At each step, the new solution is accepted according to a criterion, which typically depends on the cost difference between the solutions and a user-chosen hyperparameter, the temperature, used to mimic a physical cooling process. However, because in practice it works already sufficiently well (see Fig.~\ref{MCratio}), we refrained from designing an elaborate cooling scheme. Instead, we apply the heuristic most greedily, i.e., in the zero-temperature limit, where a new state is accepted \textit{only} if it decreases the cost.

Starting from an initial tree topology $T$, e.g.,~the minimum spanning tree~(mST) or the OT solution\footnote{In particular in the regime $\alpha \approx 1$, our BOT solver benefits from existing efficient OT solvers by using their solution as initial guess.}, we uniformly sample an edge $\hat e \in E$ and remove it from $T$. Let the incident node of $\hat e$ which ended up in the smaller connected component be $\ell$. Then, one calculates the distance $d(e, \ell)$ between $\ell$ and every edge $e=(i,j)$ in the larger component
and samples one of these edges with probability $p(e) \propto \exp(-d(e, \ell)^2 / d_{min}^2)$, where $d_{min}$ is the distance to the closest considered edge. The node $\ell$ is then connected to the sampled edge via a new BP to produce a new tree topology. For this topology, we optimize the geometry (as described in Sect.~\ref{sec:smith}) and compare costs with the previous solution. If the new state is rejected, start the next iteration by sampling $\hat e$ \textit{without replacement} until either a move is accepted and all  above steps are repeated; or until no accepted move is found, upon which the search terminates. 

Experiments for small BOT problems suggest that even in the greedy zero-temperature limit the algorithm often finds the globally optimal solution, after comparatively few iterations. For this, the greedy heuristic (using the mST as initialization) was compared against exact solutions with up to nine terminals, obtained by brute-force. For each $n$, 100 BOT problems were sampled uniformly with respect to $\alpha$, the terminal positions and demands and supplies, cf.~Alg.~\ref{alg:random}. The ratios of the heuristic's cost divided by the cost of the exact solution are plotted in Fig.~\ref{MCratio}. On average the heuristic solution is less than 0.5\% worse than the brute-force solution. This is impressive, considering the fact that for $n=9$ the brute-force solver requires over $10^5$ BP optimizations, whereas the simulated annealing heuristic on average required $29 \pm 10$ iterations to converge. Additional experiments (also for larger BOT problems) suggest that the number of BP optimizations until convergence scales better than~$O(n^2)$, see App.~\ref{app:heuristic}. 
Further, the cost ratios in Fig.~\ref{MCratio} stay roughly constant as $n$ increases. Additional experiments for BOT in higher dimensions (see Fig.~\ref{bf_all}) indicate that the average quality of the heuristic solution decreases only very slightly with $n$.   
Unfortunately, one can only speculate how this trend extends to larger BOT problems, where brute-force solutions are no longer feasible. Figure~\ref{al_init} shows heuristic solutions of a larger example problem for different values of $\alpha$. In particular, we find that the greedy heuristic is very effective at removing higher-degree branchings and undesirable edge crossings.

\section{Generalization to higher-dimensional BOT}
\label{sect:generalize_dim}

Optimal BOT solutions are acyclic also in $\mathbb{R}^d$~\cite{bernot2008optimal}. Thus, for a given topology, the edge flows are known, the optimal substructure property of Lemma~\ref{lem:subprobs} generalizes and the convex geometry optimization can be separated from the combinatorial topology optimization. Though, the optimal angle conditions for Y-, V- and L-branching (see Sect.~\ref{sect:geom_opt}) hold also in $\mathbb{R}^d$, the results on the degree limitation do not generalize, as the arguments rely on the fact that the angles between edges meeting at a higher-degree branching point sum up to $2\pi$ (cf. Sect.~\ref{subsec:nonopt}). 
The numerical geometry optimization as well as the greedy algorithm for the topology optimization presented in Sect.~\ref{sec:practical} are readily applicable to BOT problems in $\mathbb{R}^d$ (see also App.~\ref{app:smith} and App.~\ref{app:heuristic}).

\section{Conclusions}
We have studied branched optimal transport in $\mathbb{R}^2$ from a theoretical and practical perspective. First, we have tackled the geometric optimization of BOT solutions, given a tree topology. We generalized the existing exact method presented in~\cite{bernot2008optimal,gilbert1967minimum} to the case of multiple sources. Based on theory developed in the process of this generalization, we formulated a catalog of necessary and sufficient conditions for optimal BOT solutions and argued that $n$-degree branching points for $n>3$ are never optimal. Moreover, we showed that these conditions also apply for BOT on two-dimensional manifolds. Lastly, we presented a greedy randomized algorithm, which optimizes the tree topology, combined with an efficient numerical branching point optimization method. We compared our algorithm to the optimal solution for small examples, obtaining compelling results. 

BOT provides a unifying framework for optimal transport and the Euclidean Steiner tree problem and is itself of great theoretical and practical interest. The emergent branching in BOT can be used to simulate and study the myriad of efficient transportation systems which exhibit subadditive costs. Moreover, BOT combines both combinatorial and convex optimization and could be an inspiring problem to be solved by machine learning techniques. The number of optimality criteria derived in this paper can guide further research in this area and the presented approximate solvers may serve as competitive baseline for new ML-based approaches. 

\begin{ack}
We would like to thank Jarosław Piersa for sharing his code with us for a comparison to his work. Further, we thank Edouard Oudet for helpful hints on the comparison to his related work and Fabian Egersdoerfer for his improved C++ implementation of the geometry optimization.

This work is supported by Deutsche Forschungsgemeinschaft (DFG) under Germany’s Excellence Strategy EXC-2181/1 - 390900948 (the Heidelberg STRUCTURES Excellence Cluster), by Informatics for Life and by SIMPLAIX funded by the Klaus Tschira Foundation.
\end{ack}

\newpage
\bibliography{references}
\bibliographystyle{abbrvnat}

\newpage
\appendix

\section*{Appendix overview}
The appendix is subdivided into the following seven topics:
\begin{enumerate}[label=\textbf{\Alph*}, leftmargin=0.7cm]
    \item \textbf{\nameref{app-central}}: Quick proof of the central angle property, used in Sect.~\ref{subsec:constr} in the recursive construction of relatively optimal solutions based on the optimal branching angles. 
    \item \textbf{\nameref{sec_LV}}: Derivation of the conditions listed in Tab.~\ref{tab_corner} under which V- or L-branching are optimal.
    \item \textbf{\nameref{sec:subopt}}: Formal proof of Lemma~\ref{lem:subprobs} on the optimal substructure of BOT solutions.
    \item \textbf{\nameref{app_fh}}: Collection of small lemmas on the monotonicity and other properties of the analytical expression for the branching angles (cf.~Eq.~(\ref{eq_fandh})).
    \item \textbf{\nameref{appsec:nonopt}}: Technical proofs and numerical scheme to show the non-optimality of higher-degree branchings discussed in Sect.~\ref{subsec:nonopt}.
    \item \textbf{\nameref{app:mf}}: Formal proof of Theorem~\ref{thm-mf-ang}, which generalizes the optimal branching conditions and other properties from the Euclidean plane to embedded surfaces. A sketch of the proof can be found in Sect.~\ref{sec:approx-mf}. 
    \item \textbf{\nameref{appsec:algo}}: Additional details and experiments for the different algorithms presented in the main paper. Section~\ref{app:smith} focuses on the numerical geometry optimization and Sect.~\ref{app:heuristic} on the greedy algorithm for the topology optimization. Section~\ref{app:constr} holds a few examples of the recursive geometric construction of relatively optimal solutions for BOT problems with multiple sources.
\end{enumerate}

\section{Central angle property}
\label{app-central}
\begin{figure}[h]
\centering
     \begin{subfigure}[b]{0.17\textwidth}
         \centering
         \includegraphics[width=\textwidth]{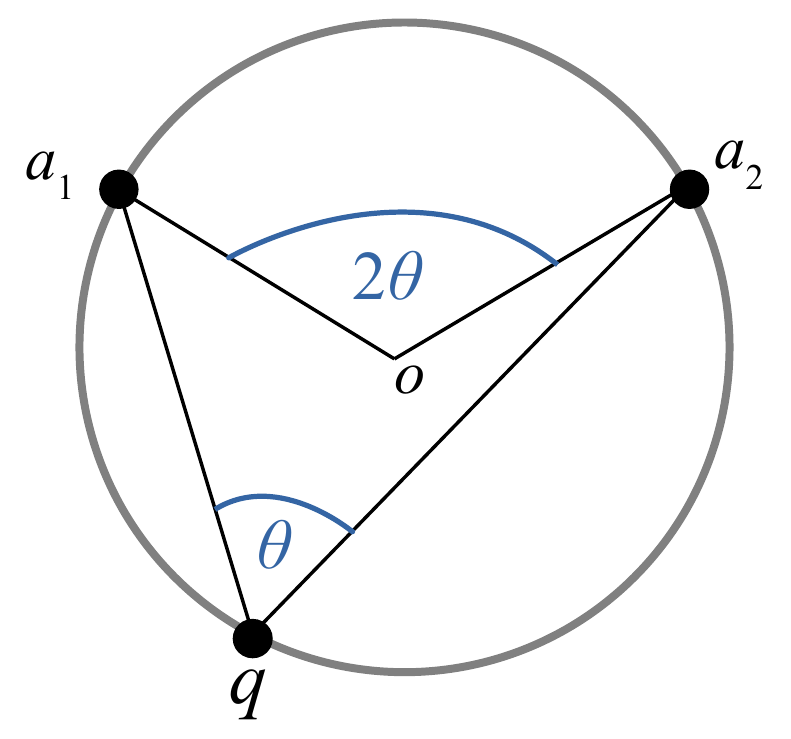}
         \caption{}
         \label{central-thm}
     \end{subfigure}
     \hspace{1.7cm}
     \begin{subfigure}[b]{0.17\textwidth}
         \centering
         \includegraphics[width=\textwidth]{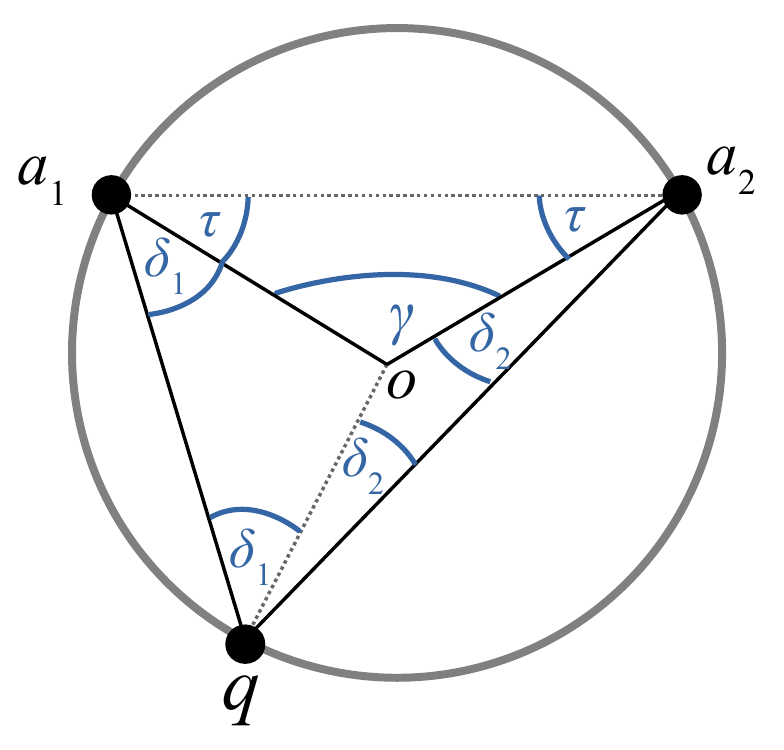}
         \caption{}
         \label{central-proof}
     \end{subfigure}
\caption{\textbf{(a)} Illustration of the central angle theorem. \textbf{(b)} Isosceles triangles used in the geometric proof.}  
\label{central}
\end{figure}

In this section, we present a geometric proof of the central angle property used in the geometric construction of relatively optimal solutions for a given full tree topology (see Sect.~\ref{subsec:constr}). It states that for a circle, as in Fig.~\ref{central-thm}, the central angle $\angle a_1 o a_2$ is twice the angle $\angle a_1 q a_2 = \theta$ for all $q$ on the lower circle arc.

Let us start by constructing the three isosceles triangles $\triangle o q a_1$, $\triangle o q a_2$ and $\triangle o a_1 a_2$ with angles as denoted in Fig.~\ref{central-proof}. Now consider the angular sums in the following triangles:
\begin{align*}
\triangle q a_1 a_2 : \ 2 \delta_1 + 2 \delta_2 + 2 \tau = 180^\circ \ \text{ and } \
\triangle o a_1 a_2 : \gamma + 2 \tau = 180^\circ \ .
\end{align*} 
Subtracting the two equations immediately reveals that $\gamma =  2 \delta_1 + 2 \delta_2 = 2 \theta$ and the proof is complete.

\section{Optimal L- and V-shaped branching}
\label{sec_LV}
Below, we formally derive the conditions listed in Tab.~\ref{tab_corner} under which V- or L-branching provide the optimal solution to a BOT problem with one source and two sinks. The proof is inspired by the approach in~\cite{mordukhovich2013fermat}, where subdifferentials are applied to the related Fermat-Torricelli problem.

\begin{defn}[Subgradient and subdifferential] A vector $v \in \mathbb{R}^n$ is called a \textit{subgradient} of a convex scalar function $g: \mathbb{R}^n \to \mathbb{R}$ at a specific point $y$ if for all $x \in \mathbb{R}^n$ it satisfies
\begin{align}
g(x) \ge g(y) + \langle  v,  x -  y \rangle
\end{align}
The set of all subgradients of the function $g$ at a given point $y$ is called the \textit{subdifferential} of $g$ at $y$ and is denoted by $\partial g(y)$. \end{defn}

From a geometric point of view, the subdifferential of $g$ at $y$ is the set of gradients of all straight lines which cross $g(y)$ and lie below the image of $g$. The subdifferential rule of Fermat follows immediately from the definition and states that $g$ achieves an absolute minimum at $y$ if and only if $0 \in \partial g(y)$. Now, for $g(b) = c \cdot \left\| b - a \right\|_2$ with $c \in \mathbb{R}$ and $a \in \mathbb{R}^n$ the subdifferential is given by
\begin{align*}
\partial g(b) = \begin{cases}
\mathbb{B}_{c} \, , \quad \quad \quad  \ \text{ for }  b = a \\
\bigg \{ c \frac{b - a}{\left\| b - a \right\|_2} \bigg \} \,  , \text{ elsewhere, }
\end{cases}  
\end{align*}
with $B_r = \{ v \in \mathbb{R}^n : \left \| v \right \|_2 \le r \}$, the ball of radius $r$.
Furthermore, it can be easily shown that for \mbox{$g(x) = g_1(x) + g_2(x)$}, one has $\partial g(y) = \partial g_1(y) + \nabla g_2(y)$, given that both $g_1$ and $g_2$ are convex functions and $g_2$ is differentiable.\footnote{The sum of a vector and a set of vectors, as in $\partial g_1(y) + \nabla g_2(y)$, is known as the Minkowski sum.} Using this, we calculate the subdifferentials of the cost function of the 1-to-2 branching in Eq.~(\ref{eq_cost1to2}). The subdifferentials at $b = a_i$ are of the form
\begin{align*}
\partial \mathcal{C}(a_i) = \mathbb{B}_{m_i^\alpha} + \sum_{j \neq i}  
m_j^\alpha \frac{ a_i - a_j}{ \left\| a_i - a_j \right\| } \ .
\end{align*}  
Based on the rule of Fermat, the cost function achieves an absolute minimum at $b = a_i$ if and only if
\begin{align} \label{eq_subdiff_cond}
\left\| \ \sum_{j \neq i} m_j^\alpha \frac{ a_i - a_j}{\left\| a_i - a_j \right\|} \ \right\| \le m_i^\alpha \ . \\[-0.4cm]
\nonumber
\end{align}

\paragraph{V-branching.} We square condition~(\ref{eq_subdiff_cond}) and evaluate it for $i=0$ in order to determine under which condition a V-shaped branching with $b^* = a_0$ is optimal:
\begin{align*}
\left\| \  m_1^\alpha \frac{a_0 - a_1}{\left\| a_0 - a_1 \right\|} 
+  m_2^\alpha \frac{a_0 - a_2}{\left\| a_0 - a_2 \right\|}
 \ \right\|^2 = 
m_1^{2 \alpha} + m_2^{2 \alpha} + 2 m_1^\alpha m_2^\alpha \cos(\psi)  
\le m_0^{2\alpha} = (m_1 + m_2)^{2 \alpha} \, 
\end{align*}
where $\psi$ denotes the angle of the terminal triangle at $a_0$, i.e.,~\mbox{$\psi = \angle a_1a_0a_2$}. The condition in terms of $\psi$ reads
\begin{align} \label{eq_psi_cond}
\psi \ge \arccos \bigg(  \frac{1 - k^{2 \alpha} - (1-k)^{2 \alpha}}{2 k^\alpha (1-k)^\alpha} \bigg) = h(\alpha, k) = \theta^*_1 + \theta^*_2 \ ,
\end{align} 
where we have used the flow fraction $k = m_1/(m_1 + m_2)$. We immediately recognize the expression for the optimal branching angle $\theta^*_1 + \theta^*_2$, cf.~Eq.~(\ref{eq_fandh}). 

\begin{figure}[H]
\centering 
\hspace*{-0.cm}
\includegraphics[scale=0.3]{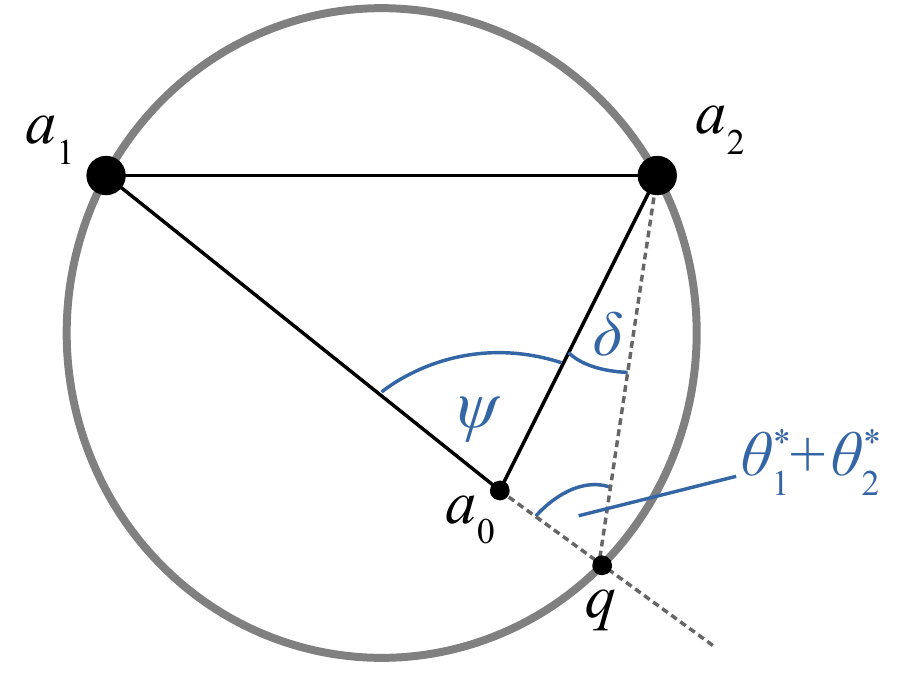} 
\caption{Sketch to show that $\psi > \theta^*_1 + \theta^*_2$.}  
\label{v-proof}
\end{figure}

We already know that $\psi = \theta^*_1 + \theta^*_2$ on the lower circle arc of the pivot circle by construction. And indeed one can easily check that $\psi > \theta^*_1 + \theta^*_2$ if an only if the source $a_0$ lies inside the lower half of the pivot circle, as in Fig.~\ref{v-proof}. For that, we construct a line through $a_1$ and $a_0$ and the intersection of $\overline{a_1a_0}$ with the lower pivot circle we denote by $q$. By construction of the pivot circle, $\angle a_1 q a_2 = \theta^*_1 + \theta^*_2$. Using the angular sum in the triangle $\triangle a_0 q a_2$, one immediately obtains: 
\begin{align*}
180^\circ = \delta + ( 180^\circ - \psi) + (\theta^*_1 + \theta^*_2) \ \to \ \psi - (\theta^*_1 + \theta^*_2) = \delta > 0 \ .
\end{align*}

Hence, for $a_0$ located inside the lower half of the pivot circle indeed $\psi > \theta^*_1 + \theta^*_2$. A similar argument can show that for any point outside the pivot circle the optimal V-branching condition is not fulfilled.

\begin{figure}[H]
     \centering
     \begin{subfigure}[b]{0.35\textwidth}
         \centering
         \includegraphics[width=\textwidth]{figures/vll1new.pdf}
         \caption{ Regions of optimal Y-, V- and \mbox{L-branching}}
         \label{vll1app}
     \end{subfigure}
     \hspace{0.7cm}
     \begin{subfigure}[b]{0.35\textwidth}
         \centering
         \includegraphics[width=\textwidth]{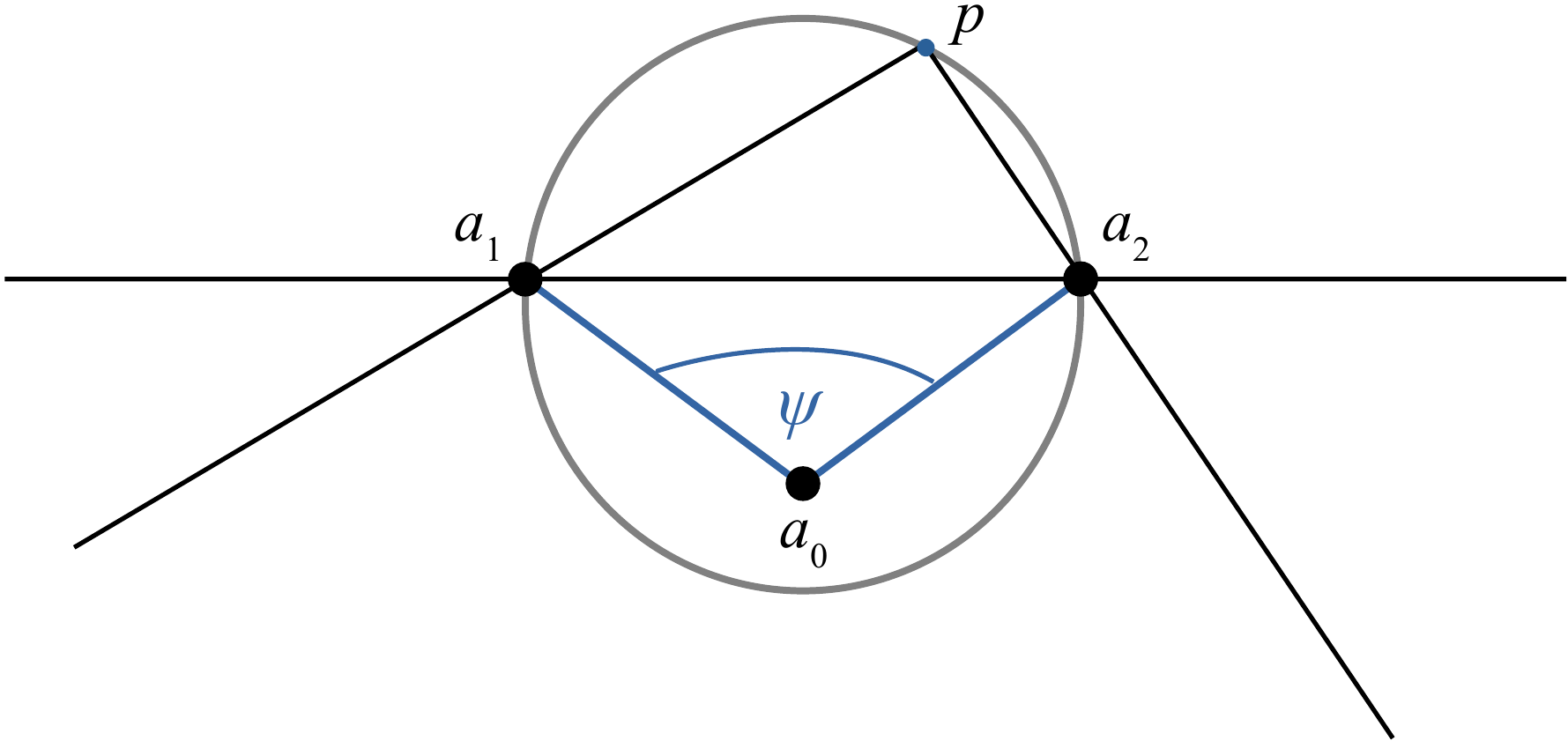}
         \caption{V-branching \\ }
         \label{vll2}
     \end{subfigure} \\
     \vspace{0.7cm}
	 \centering
     \begin{subfigure}[b]{0.35\textwidth}
         \centering
         \includegraphics[width=\textwidth]{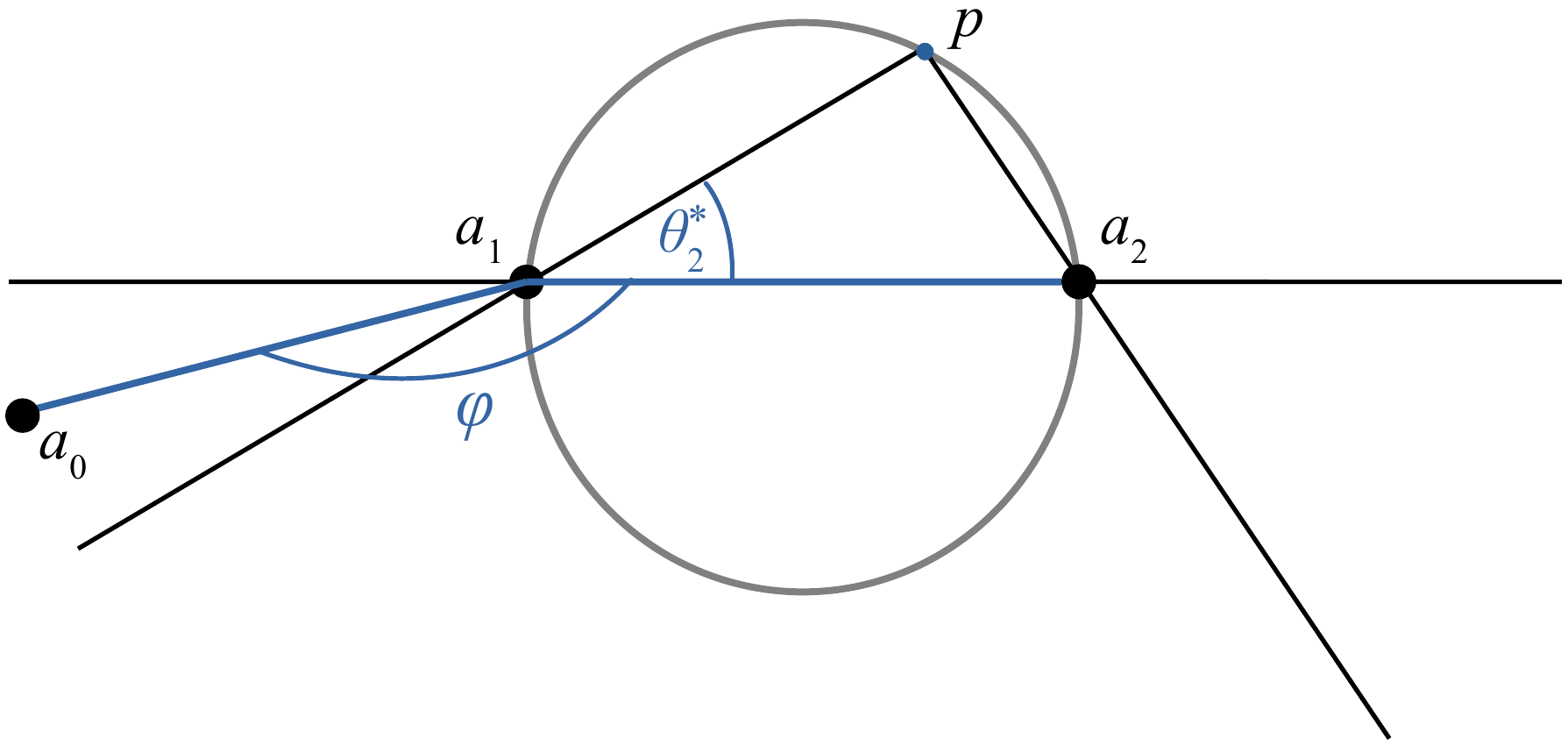}
         \caption{L1-branching}
         \label{vll3}
     \end{subfigure}
     \hspace{0.7cm}
     \begin{subfigure}[b]{0.35\textwidth}
         \centering
         \includegraphics[width=\textwidth]{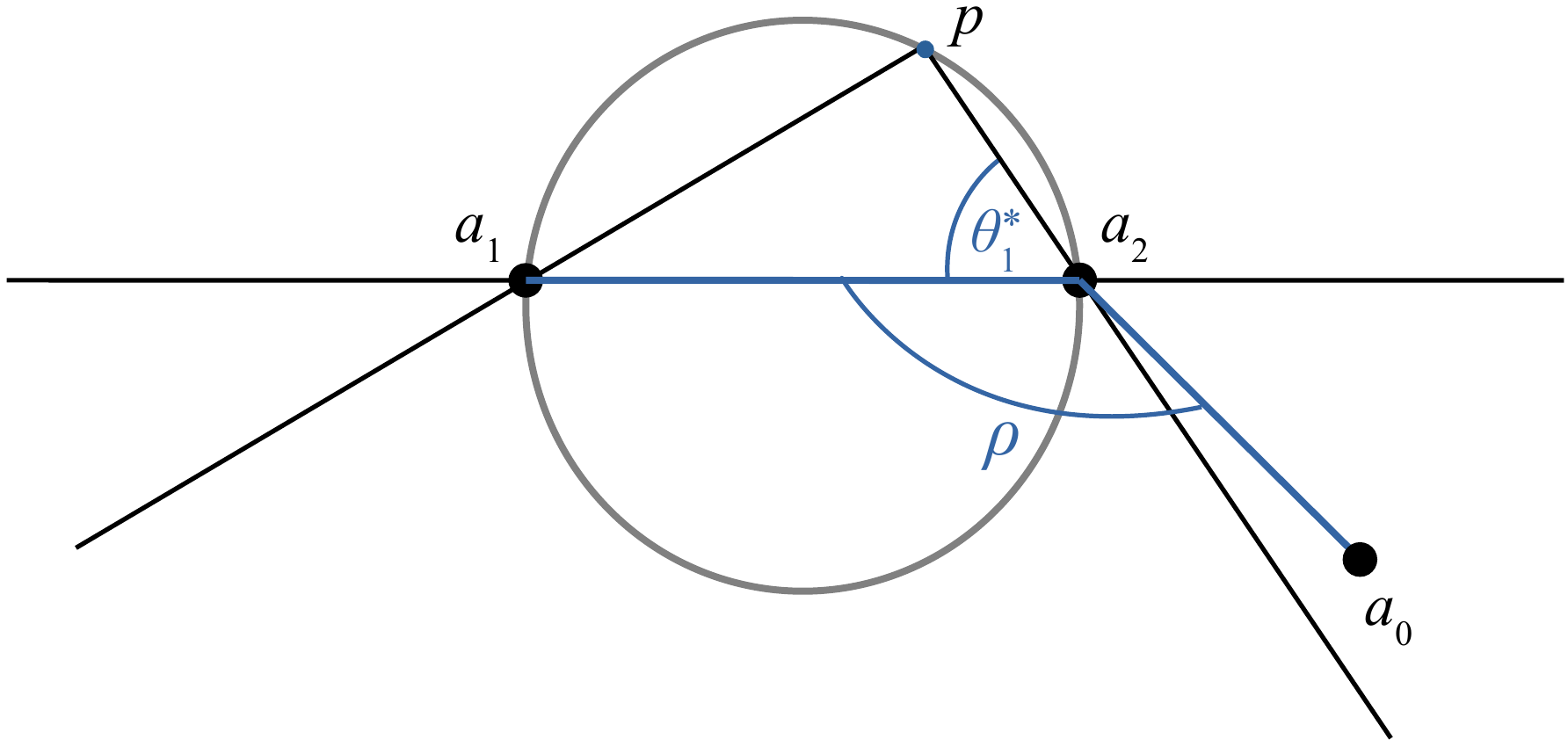}
         \caption{L2-branching}
         \label{vll4}
     \end{subfigure}
        \caption{After constructing the pivot point the position of the source $a_0$ determines whether a Y-,V- or L-shaped branching is optimal.}
        \label{vllapp}   
        
\end{figure}

\paragraph{L-branching.} Analogous to the above steps, one obtains conditions for optimal L1- and L2-branching, where $b^*= a_1$ and $b^*= a_2$ respectively. Again squaring the general condition~(\ref{eq_subdiff_cond}), now for $i=1,2$, one eventually finds that
\begin{align}
\varphi &\ge \pi - f(\alpha, 1-k) = \pi - \theta^*_2 \ , \\
\varrho &\ge \pi - f(\alpha, k) = \pi - \theta^*_1  \label{eq_l2_cond} \ ,
\end{align} 
where the angle $\varphi$ and $\varrho$ denote the angles of the terminal triangle located at $a_1$ and $a_2$, i.e.~\mbox{$\varphi = \angle a_2a_1a_0$} and $\varrho = \angle a_0a_2a_1$ (see also Fig.~\ref{vll3},\subref{vll4}). Let us now demonstrate that these conditions are indeed fulfilled if and only if the source is located in the L1- and L2-region, as marked also in Fig.~\ref{vll1}. The pivot point $p$ is constructed such that $\angle a_1 o p = 2 \theta^*_1$ and $\angle p o a_2 = 2 \theta^*_2$, cf.~Fig.~\ref{1-to-2ybranching}. Besides that, by construction, we have that $\angle p a_1 a_2 = \theta^*_2$ and $\angle a_1 a_2 p = \theta^*_1$ (as shown in Fig.~\ref{vll3},\subref{vll4}). Then, looking at Fig.~\ref{vll3}, it is evident that indeed for any source $a_0$ inside the L1-branching sector the condition $\varphi = \angle a_2a_1a_0  \ge \pi - \theta^*_2$ holds. The respective condition~(\ref{eq_l2_cond}) for L2-branching holds true exactly inside the highlighted L2-region. 

\begin{defn}[Transient and strict V- and L-branchings] \label{defn:trans}
A V- or L-branching for which the inequality conditions (\ref{eq_psi_cond})-(\ref{eq_l2_cond}) hold as equality is referred to as \textit{transient} \label{glo-trans} V- or L-branching. The reason for this is that, in such a case, one of the terminal positions may be perturbed infinitesimally, so that the condition is no longer fulfilled and the optimal solution transitions to a Y-shaped branching. On the contrary, if an L- or V-branching condition is fulfilled as strict inequality, we call the L- or V-branching \textit{strict}.
\end{defn}

\subsection{Relation of V- and L-conditions between symmetric and asymmetric branchings}
Let us briefly show that the corner cases of Y-, V- and L-branching work analogously for both flow scenarios (described in Fig.~\ref{1-to-2mine},\subref{case2a}). The conditions under which L- and V-branching are optimal in the asymmetric branching case could be again determined straightforwardly by plugging into the subdifferential condition~(\ref{eq_subdiff_cond}), as before. However, looking at Fig.~\ref{1-to-2mine},\subref{case2a}, we notice that the well-known symmetric branching case can be transformed into the asymmetric case by relabeling $a_0 \to a_2$, $a_1 \to a_0$ and $a_2 \to a_1$. This relation provides a direct correspondence of the L- and V-branching conditions. The conditions for the corner case are transferred according to Table~\ref{tab_cornerapp}. Note that the L- and V-branching conditions for both cases are of the exact same form, only that for the asymmetric case the stationary branching angles are $\vartheta^*_i$ instead of $\theta^*_i$. Moreover, for fixed children positions $a_1$ and $a_2$, the position of the source again distinguishes between optimal Y-, V- and L-branching. The partitioning of the lower half plane into the respective regions is completely analogous to Fig.~\ref{vll1app}.

\begin{table}[h]
{\renewcommand{\arraystretch}{1.4}
\begin{center}
\vspace{-0.15cm}
\caption{Relations between the L- and V-branching conditions (see Fig.~\ref{vllapp}) for symmetric and asymmetric branchings.}
\vspace{0.15in}
\begin{footnotesize}
\begin{tabular}{ c c } 
 \hline
 symmetric branching & asymmetric branching \\
 \hline 
 $\ $ V: \scalebox{1}{$\  \angle a_1a_0a_2 \ge \theta^*_1 + \theta^*_2$} & L2: \scalebox{1}{$\angle a_0a_2a_1 \ge \theta^*_1 + \theta^*_2 = \pi - \vartheta^*_1$} \\ 
 L1: \scalebox{1}{$\angle a_2a_1a_0 \ge \pi - \theta^*_2$}  & 
 $\ $ V: \scalebox{1}{$\  \angle a_1a_0a_2 \ge \pi - \theta^*_2 = \vartheta^*_1 + \vartheta^*_2$} \\ 
 L2: \scalebox{1}{$\angle a_0a_2a_1 \ge \pi - \theta^*_1$} & L1: 
 \scalebox{1}{$\angle a_2a_1a_0 \ge \pi - \theta^*_1 = \pi - \vartheta^*_2$}
 \\ 
\hline
\end{tabular}
\end{footnotesize}
\label{tab_cornerapp}
\end{center}}
\end{table}

\section{Optimal substructure of BOT solutions (Proof of Lem.~\ref{lem:subprobs})}
\label{sec:subopt}
In this section, we provide the formal proof to Lemma~\ref{lem:subprobs}, repeated below for completeness:

\begin{lem}
\textbf{(a)} For a given tree topology, a BOT solutions is a relatively optimal if and only if every (coupled) BP connects its (effective) neighbors at minimal cost. \textbf{(b)} In a globally optimal solution, every subsolution restricted to a connected subset of nodes solves its respective subproblem globally optimally.
\end{lem}

Let us start with the following definition, which divides a BOT problem and its possible solutions into subproblems and corresponding subsolutions. 

\begin{figure}[!htp]
\centering 
\hspace*{0cm}
\includegraphics[scale=0.32]{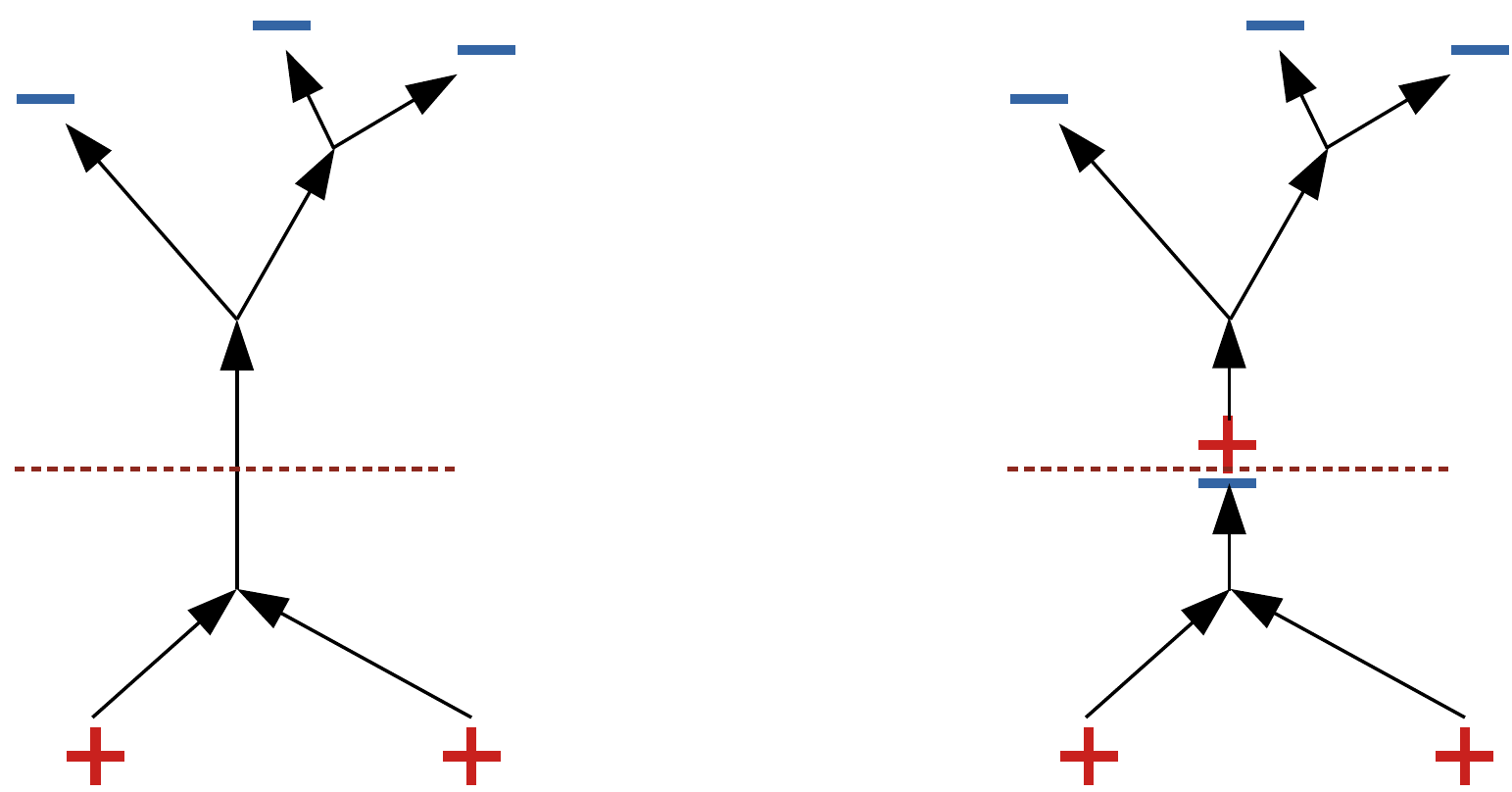} 
\caption{A BOT solution (left) is split into two subsolutions (right) by cutting an edge as indicated by the dashed line.}  
\label{splitting}
\end{figure}

\begin{defn}[Subproblems and subsolutions] \label{defn:split} 
A given BOT solution may be split into two subsolutions, by choosing a number of edges $\{e_i\}$ and cutting them at points~$\{x_i\}$, so that the topology is split into two connected components. This procedure induces two subproblems and two subtopologies. Each \textit{subproblem} consist of the terminals contained in the respective component plus additional terminals at the positions $x_i$. The demands or supplies of the additional terminals at $x_i$ are equal to the amount of flow through the corresponding edge $e_i$ that was cut. The terminal becomes a sink in one subproblem and a source in the other according to the direction of flow through $e_i$. The two \textit{subtopologies} are given by the induced subgraph on all terminals contained in one component. The \textit{subsolutions} to the created subproblems are given by the subtopologies and the BP configurations of the respective subsets of branching points contained in each subproblem. Note that using a number of such splits a given solution may be divided into several subsolutions, each solving their respective subproblem. An illustrative example can be found in Fig~\ref{splitting}.
\end{defn}

\paragraph{Proof of the optimal substructure as necessary condition for optimality.} The optimal substructure as necessary optimality condition follows immediately from the independence of the subproblems. Given that a solution is minimal in cost (in the relative or global sense), each subproblems itself must be minimal in that sense. Otherwise the cost of this subproblem could be decreased by (a) improving the BP configuration or (b) the topology. In that case, also the full solution could be improved in cost and could thus not be optimal.

\paragraph{Proof of the optimal substructure as sufficient condition for optimality.} The optimal substructure as sufficient optimality condition stems from the fact that, given a tree topology, the BOT cost function is convex with respect to the BP positions. As a first step, let us prove the following lemmata: 

\begin{lem} \label{lem_just} Let $b$ be a branching point with neighbors $\{a_i \}$. Let the optimal position of $b$ be at one of its neighbors with the condition~(\ref{eq_subdiff_cond}) fulfilled as equality, i.e.~the branching is transient. 
Then, for every  $\delta >0 $ there exists a BP location $b_\delta \in \mathbb{R}^2$ away from all neighbors with $\left \| b - b_\delta \right \|_2 < \delta$ and a continuous function $f:\mathbb{R^+}\to \mathbb{R}$ with $\lim_{x\to 0}f(x)=0$, such that the gradient $\left \| \nabla_{\! \! b} \, \mathcal{C}(b_\delta) \right \|_2 < \epsilon$ for $0<\epsilon<f(\delta)$.
\end{lem}

\noindent  \textit{Proof.} 
Let $\mathcal C(b)$ be the cost of the subproblem with terminals $a_i$. WLOG, the condition~(\ref{eq_subdiff_cond}) is fulfilled as equality for terminal $a_0$, meaning that 
\begin{align} 
\left \| v \right \| = m_0^{\alpha} \  \text{ with } \ v:= \sum_{i \neq 0} m_i^\alpha \frac{ a_0 - a_i}{\left\| a_0 - a_i \right\|} \ . 
\end{align}
The optimal branching point is hence located at $a_0$. Consider the alternative branching point position $b_\delta = a_0 - \delta v$ with sufficiently small $\delta >0$, such that $b_\delta$ is located away from all terminals. The gradient with respect to the branching point position is well defined at $b_\delta$ and reads:
\begin{align}
\nabla_{\! \! b} \, \mathcal C(b_\delta) = - m_0^\alpha \frac{v}{\left \| v \right \|} +  
\sum_{i \neq 0} m_i^\alpha \frac{ a_0 - a_i - \delta v}{\left\| a_0 - a_i -\delta v \right\| } \ .
\end{align}
Clearly, $\left \| \nabla_{\! \! b} \, \mathcal C(b_\delta) \right \|$ can be brought arbitrarily close to zero as $\delta \to 0$, since
\begin{align}
\left \| \nabla_{\! \! b} \, \mathcal C(b_\delta) \right \|^2 = \underbrace{m_0^{2 \alpha} + m_0^{2 \alpha} - 2 m_0^\alpha \Big \langle v, \frac{v}{\left \| v \right \|} \Big \rangle }_{= \, 0} + O(\delta)  \ ,  
\end{align}
comprising the terms which go to zero as $\delta \to 0$ in $O(\delta)$. The function absorbed in $O(\delta)$ therefore provides the function $f$ and $b_\delta$ is the desired branching point with arbitrarily small gradient. The argument works for both coupled and uncoupled branching points. \hfill\rlap{\hspace*{-1.em}$\qedsymbol$} \\

\begin{lem} \label{lem_large_transient}
For a given full tree topology $T$, let $B=\{b_i\}$ be the ROS of a BOT problem with terminals $A = \{a_i\}$. Let $B$ contain Y-branchings and transient V- or L-branchings. Then, for every  $\delta >0 $ there exists a non-degenerate BP configuration $B_\delta$ and a continuous function $f:\mathbb{R^+}\to \mathbb{R}$ with $\lim_{x\to 0}f(x)=0$, such that $\left \| B - B_\delta  \right \|_2  < \delta$ and the gradient norm fulfills $\left \| \nabla_{\! \! B} \, \mathcal{C}(B_\delta) \right \|_2 < \epsilon$ for $0<\epsilon<f(\delta)$.
\end{lem}

\noindent  \textit{Proof.} 
Let us split the solutions into subsolutions at branching points that exhibit optimal Y-branchings. They are kept fixed $\delta_i = 0$ and have zero gradients. The resulting subproblems can then be considered independently. It suffices to show that, for every individual subproblem, individual displacements of the BPs exist that yield a non-degenerate BP configuration with arbitrarily small gradients. If a subproblem consist of a single branching point, Lem.~\ref{lem_just} is directly applicable and the infinitesimal displacement is chosen as described there. If multiple branching points are coupled in a transient branching we proceed as follows: For the given topology, define a root node and apply the recursive geometric construction with pivot points and pivot circles, as illustrated in Fig.~\ref{recur}. In the construction, all pivot circles will meet at the position of the coupled branching point, which is the geometric equivalent of condition~(\ref{eq_subdiff_cond}) being fulfilled as equality, cf.~Fig.~\ref{just-angles}. Then, starting from the branching point furthest from the root node, take an infinitesimal step of size $\sim O(\delta)$ towards its corresponding pivot point, exactly as in Lem.~\ref{lem_just}. Thereby, the optimal branching angles will almost be realized and consequently the gradient of the resulting uncoupled BP will be arbitrarily small (as shown explicitly in Lem.~\ref{lem_just}). For the next BP in topological order (w.r.t.~the chosen root node) repeat the procedure with a step size even smaller of size $\sim O(\delta^2)$, again the optimal branching angles are almost realized and the resulting gradient vanishes as $\delta \to 0$. Repeating, this procedure for every of the finitely many branching points with smaller and smaller step sizes produces a non-degenerate BP configuration with arbitrarily small gradient. 
 \hfill\rlap{\hspace*{-1.em}$\qedsymbol$} \\

Based on Lem.~\ref{lem_large_transient}, we now prove that the optimality of each individual branching point is a sufficient condition for relatively optimal solutions. Similar to the proof above we do not explicitly distinguish between coupled and uncoupled branching points. Given a fixed tree topology, the BOT cost function $\mathcal{C}(\{b_i \})$ is a convex function of the branching point positions $\{b_i \}_{1 \le i \le m}$. Let us summarize all branching point coordinates in the vector $B \in \mathbb{R}^{2m}$ and denote the configuration in which every BP connects its neighbors at minimal cost by $B^* = \{b_i^* \}$. Due to the convexity it suffices to show that the BP configuration $\{b_i^* \}$ is a local minimum of the cost function. We may distinguish the following three cases:
\begin{enumerate}[label=(\alph*)]
\item If the BP configuration $\{b_i^* \}$ is non-degenerate, the cost function is differentiable and for each BP one has $\nabla_{\! \! b_i} \, \mathcal C(b_i^*) = 0$ and thus also $\nabla_{\! \! B} \, \mathcal{C}(B^*) = 0$.

\item Let the BP configuration additionally contain BPs which are strictly anchored at one of their neighbors (referred to as anchor), meaning that the inequality condition~(\ref{eq_subdiff_cond}) holds strictly (see also Def.~\ref{defn:trans}). Denote the subset of these BPs by $\{ \tilde b_j \}$. We intend to show that a neighborhood around $B^*$ exists such that $B^*$ is the minimal cost configuration in it. If the neighboring branching points of $\{ \tilde b_j \}$ are moved away from $B^*$ only by a sufficiently small $\delta$, the optimal position for $\{ \tilde b_j \}$ stays at their anchors, since the condition~(\ref{eq_subdiff_cond}) will still hold. Thus, we can find a sufficiently small neighborhood $U$ around $B^*$, so that WLOG, the coordinates of $\{ \tilde b_j \}$ are fixed to be equal to the coordinates of their respective anchors, as other configurations would be suboptimal. If a branching point is anchored at an external node, its position is fixed to the terminal position and the cost function restricted to $U$ no longer depends on them. So, by choosing $U$ sufficiently small, the cost function depends only on BPs away from terminals. Summarize the remaining free branching points in the vector $B_{f}$ for which the gradient at $B_{f}^* \subset B^*$ is zero by assumption, i.e.~$\nabla_{\! \! B_f} \, \mathcal{C}(B_f^*) = 0$. Thus, $B^*$ is a local minimum in $U$ and due to the convexity an absolute minimum.

\item Let the BP configuration additionally contain transient branchings, i.e.~branching points for which the condition~(\ref{eq_subdiff_cond}) holds as equality. Denote them by $\{\hat b_k \}$. Again, the sufficiently small neighborhood around $B^*$ is constructed, as before, to remove the dependency of $\mathcal C$ on $\{\tilde b_j \}$. However, the cost function still depends on the position of the $\{\hat b_k \}$, and $C(B_f)$ is not differentiable at $\{\hat b_k^* \} \subset B_f^*$. Let us assume a BP configuration $B_{f,0}$ existed with cheaper cost than $B^*_{f}$, i.e. $\mathcal{C}(B^*_{f}) = \mathcal{C}(B_{f,0}) + \Delta$ for some $\Delta > 0$. Lemma~\ref{lem_large_transient} states that we can find an arbitrarily small $\delta$ so that $B^*_{f} + \delta$ is non-degenerate and the gradient $\nabla_{\! \! B_f} \, \mathcal{C}(B^*_{f} + \delta)$ is arbitrarily small. Due to the convexity of $\mathcal C$, the following inequality holds:
\begin{align*}
\mathcal{C}(B_{f,0}) &\ge \mathcal{C}(B^*_{f} + \delta) + \langle \nabla_{\! \! B_f} \, \mathcal{C}(B^*_{f} + \delta), B_{f,0} - (B^*_{f} + \delta) \rangle \\
&= \mathcal{C}(B^*_{f}) + \underbrace{O( \left \| \delta \right \|_2) + \langle \nabla_{\! \! B_f} \, \mathcal{C}(B^*_{f} + \delta), B_{f,0} - (B^*_{f} + \delta) \rangle}_{=: \ K(\delta)} 
\end{align*}
where we have used that $\mathcal{C}(B^*_{f} + \delta) = \mathcal{C}(B^*_{f}) + O( \left \| \delta \right \|_2)$, using Big-O notation. All terms which tend to zero as $\delta \to 0$ have been summarized in $K(\delta)$. $K(\delta)$ can be brought arbitrarily close to zero by choosing a sufficiently small $\delta$. In particular, there exists a $\delta > 0$ so that $\vert K(\delta) \vert < \Delta / 2$. Using the assumption $\mathcal{C}(B^*_{f}) = \mathcal{C}(B_{f,0}) + \Delta$, this leads to: 
\begin{align*}
\mathcal{C}(B_{f,0}) &\ge  \mathcal{C}(B^*_{f}) + K(\delta) = \mathcal{C}(B_{f,0}) + \Delta + K(\delta) \ge  \mathcal{C}(B_{f,0}) + \frac{\Delta}{2} \ ,
\end{align*} 
which is a contradiction. Hence $B^*$, is again a local minimum. \hfill\rlap{\hspace*{-1.em}$\qedsymbol$}
\end{enumerate}

\section{Properties of the functions $f$ and $h$ describing the optimal branching angles}
\label{app_fh}
\begin{figure}[H]
     \centering
     \begin{subfigure}[b]{0.45\textwidth}
         \centering
         \includegraphics[width=\textwidth]{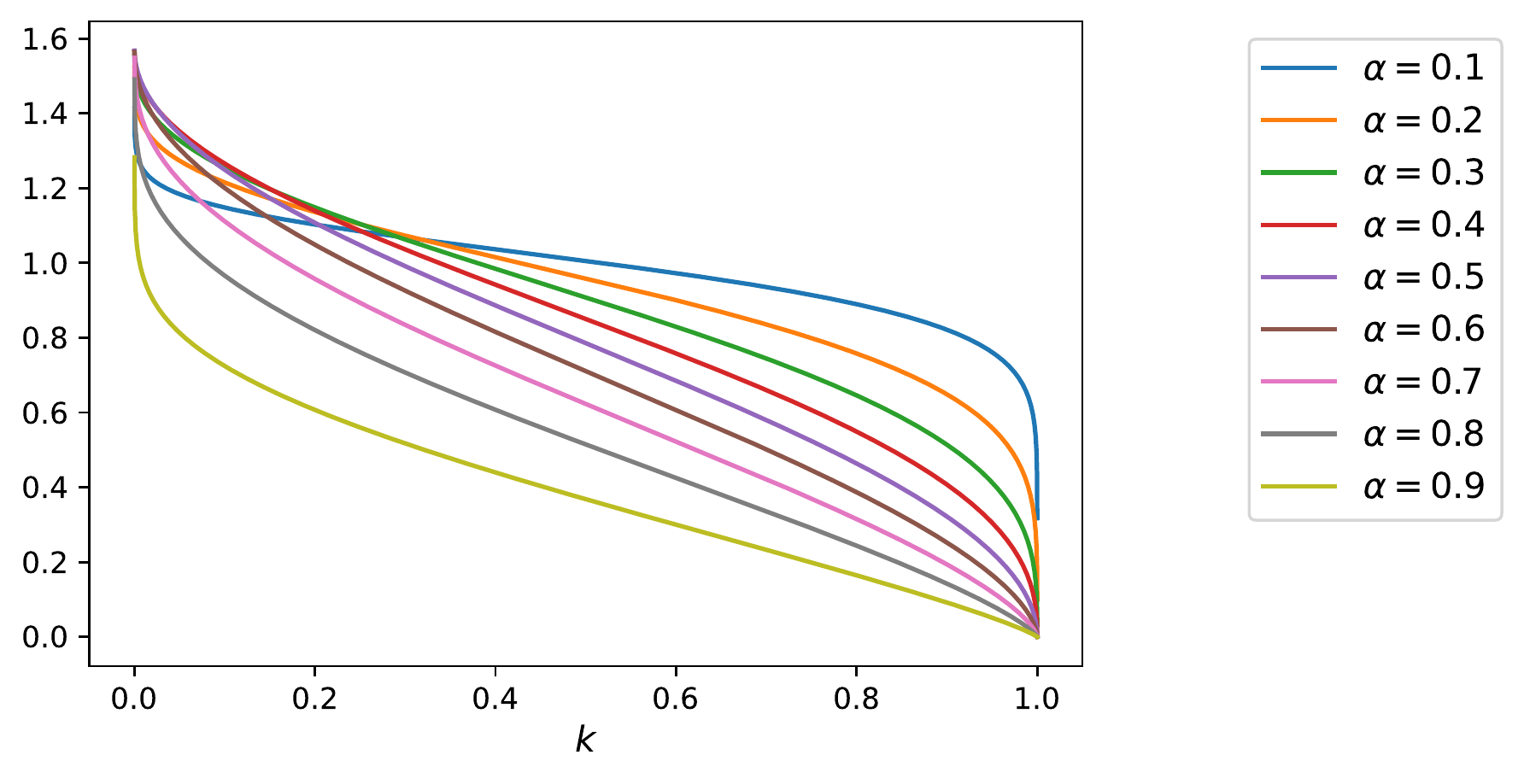}
         \caption{$f(\alpha, k )$ \phantom{blaaaaaaaaaaaaaaa}}
         \label{f-plot}
     \end{subfigure}
     \hspace{0.5cm}
     \begin{subfigure}[b]{0.33\textwidth}
         \centering
         \includegraphics[width=\textwidth]{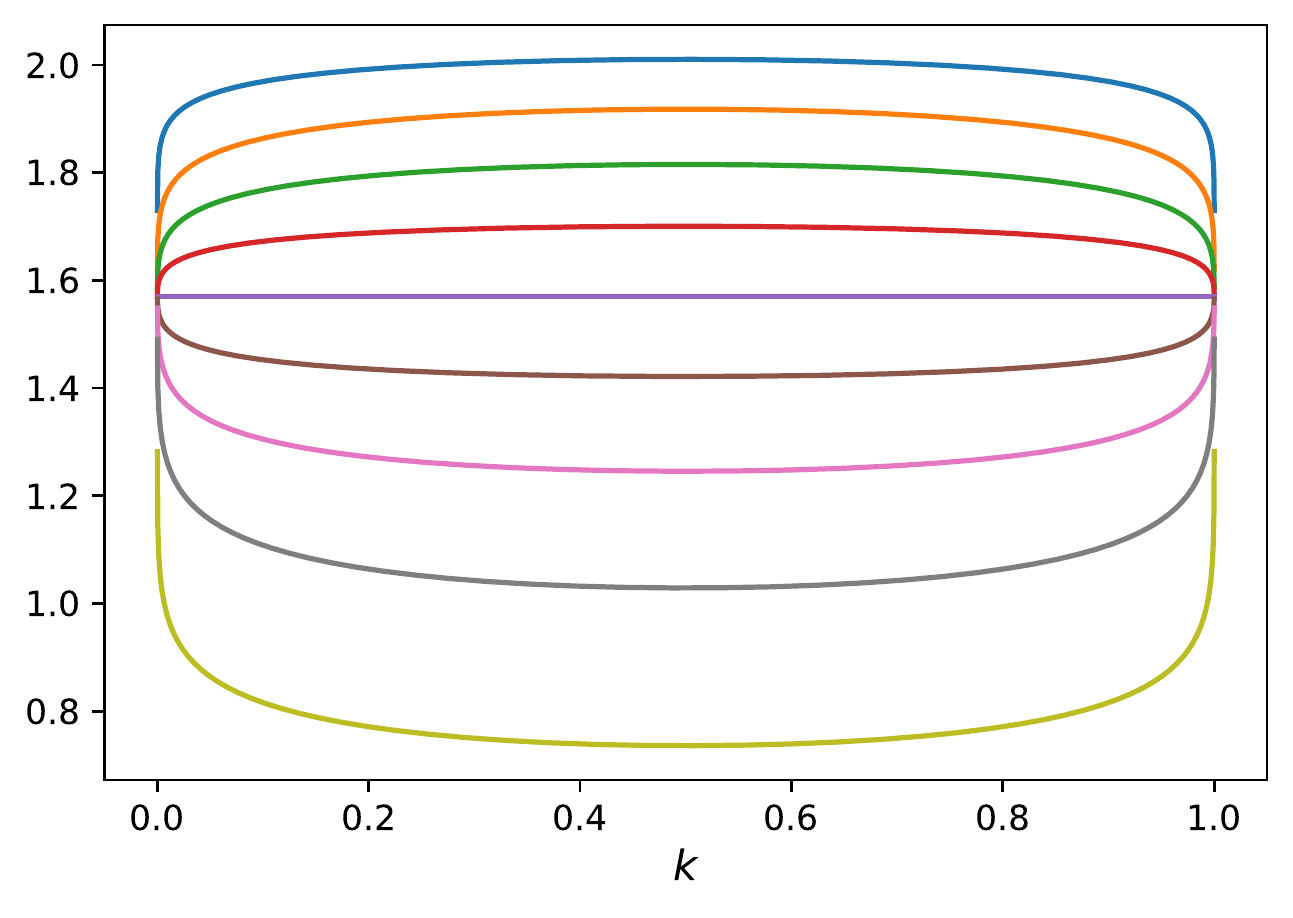}
         \caption{$h(\alpha, k )$}
         \label{h-plot}
     \end{subfigure}
        \caption{The functions $f$ and $h$ vs. $k$ for different values of $\alpha$.}
        \label{fh-plot}
\end{figure}

The optimal branching angles are expressed in terms of the following functions $f(\alpha, k)$ and $h(\alpha, k)$, defined for $\alpha \in [0,1]$ and $k \in (0,1)$, cf.~Eq.~(\ref{eq_fandh}):
\begin{align*}
f(\alpha, k) &= \arccos \bigg( \frac{k^{2\alpha} + 1 - (1-k)^{2 \alpha }}{2 k^\alpha} \bigg) \ , \\
h(\alpha, k) &= \arccos \bigg(  \frac{1 - k^{2 \alpha} - (1-k)^{2 \alpha}}{2 k^\alpha (1-k)^\alpha} \bigg) \ .
\end{align*} 
Figure~\ref{fh-plot} shows $f$ and $h$ as functions of $k$ for a number of different values of $\alpha$. The two functions are related by $h(\alpha, k) = f(\alpha, k ) + f(\alpha, 1-k)$, so that $h$ is symmetric around $k=1/2$. Both functions are defined for inputs $\alpha \in [0,1]$ and $k \in (0,1)$. For $\alpha = 1$, we have $f(\alpha=1, k) = 0 = h(\alpha=1, k)$ for all $k$, reflecting the fact that Y-shaped branchings are never optimal in the optimal transport case. On the other hand, we have $f(\alpha=0, k) = \pi/3$ and $h(\alpha=0, k)= 2 \pi / 3$ for all $k$. This limit corresponds to the Euclidean Steiner tree problem where in an optimal Y-branching all edges meet at $120^\circ$. Moreover, considering the limits of $k \to 0$ and $k \to 1$, one finds that $f(\alpha, k \to 0) \to \pi/2$ and $f(\alpha, k \to 1) \to 0$. Consequently, $h(\alpha, k) \to \pi/2$ for both $k \to 0$ and for $k \to 1$. In the following three lemmas, we investigate the monotonicities of the two functions.

\begin{lem} $f(\alpha, k)$ is strictly decreasing in $k$ for all $\alpha \in (0,1)$ and $f(\alpha,k) < \pi/2$ for all $\alpha \in [0,1]$ and $k \in (0,1)$. 
\label{lem:monot-f-k}
\end{lem}

\noindent \textit{Proof.}       
Since the inverse cosine function decreases monotonically, it is sufficient to show that the argument is monotonically increasing. Let us consider its derivative 
\begin{align*}
\frac{\partial}{\partial k} \bigg( \frac{k^{2\alpha} + 1 - (1-k)^{2 \alpha }}{2 k^\alpha} \bigg) 
 \sim k^{2 \alpha} + (1-k)^{2 \alpha} +  2 k (1-k)^{2 \alpha - 1}  - 1 =: s(\alpha, k) \ .  
\end{align*} 
The $\sim$ indicates that we have dropped overall factors which were clearly positive, as we are only interested in the sign of the derivative. First, note that \mbox{$s(\alpha = 0, k) = \frac{1+k}{1-k} > 0$} and $s(\alpha = 1, k) = 0$ for all $k$. Secondly, $s(\alpha , k)$ is strictly decreasing with respect to $\alpha$:
\begin{align*}
\frac{\partial s}{\partial \alpha} = 
2 k^{2\alpha} \log(k) + 2(1-k)^{2\alpha} \log(1-k) + 4k (1-k)^{2\alpha -1} \log(1-k) <  0 \ .
\end{align*}
Each of the above terms are negative due to the fact that $k, (1-k) \in  (0,1)$. Together it follows that $s(\alpha, k)$ is positive for any combination of $\alpha$ and $k$ and the proof of the monotonicity is complete. The fact that $f(\alpha, k \to 0) \to \pi/2$ for all $\alpha$ then immediately implies that $f(\alpha,k) < \pi/2$.  \hfill\rlap{\hspace*{-1.em}$\qedsymbol$} \\

\begin{lem} \label{lem:dhdk} For $\alpha \le 0.5$, $h(\alpha, k)$ increases monotonically in $k$ for $k \in (0,1/2]$ and decreases for larger values of $k$. Conversely, for $\alpha \ge 0.5$, $h(\alpha, k)$ increases first until in reaches a maximum at $k=1/2$ and decreases afterwards. \end{lem} 

\noindent \textit{Proof.} The argument here follows the proof of Lemma 12.14 in~\cite{bernot2008optimal}. Due to the monotonicity of the inverse cosine function, it is again sufficient to investigate the expression in the argument. We rewrite this expression trivially, so that it becomes a function of the fraction $r := \frac{k}{1-k}$ and consider the following derivative:
\begin{align*}
\frac{\partial }{\partial r} \bigg( \frac{(r+1 )^{2\alpha} - r^{2 \alpha} - 1}{2 r^\alpha} \bigg) 
\sim -(r+1)^{2\alpha} - r^{2\alpha} + 1 + 2 r (r+1)^{2\alpha - 1} =: t(\alpha, r)
\end{align*}    
We first note that $t(\alpha = 0, r) = \frac{r-1}{r+1} \le 0$ and that $t(\alpha = 0.5, r) = 0 $ as well as $t(\alpha = 1, r) = 0$ for all $r$. Next, we show that $t(\alpha, r)$ is a concave function with respect to $r$ for $r \le 1$, which corresponds to $k \in (0,1/2]$:
\begin{align*}
\frac{\partial^2  t}{\partial \alpha^2} = 4 (r -1) (r+1)^{2\alpha -1} \log^2(r+1) - 4 r^{2\alpha} \log^2(r) \le 0 \ ,
\end{align*} 
since $r \le 1$. Taken together, we conclude that $t(\alpha,r) \le 0$ for all $\alpha \in [0,0.5]$ and that $t(\alpha,r) \ge 0$ for all $\alpha \in [0.5,1]$. Moreover, since $r$ increases monotonically w.r.t.~$k$, the monotonicity of $h(\alpha, k)$ holds for $k \in (0,1/2]$. Due to the symmetry of $h$ around $k = 1/2$, the proof is complete.
\hfill\rlap{\hspace*{-1.em}$\qedsymbol$}  \\

\begin{lem} $f(\alpha, k)$ decreases monotonically in $\alpha$ for $\alpha \in [0.5,1]$ and $h(\alpha, k)$ decreases monotonically in $\alpha$ for all $\alpha \in [0,1]$. \end{lem} 

\noindent \textit{Proof.}   
As the inverse cosine function decreases monotonically, it is again sufficient to investigate the derivative of the function in the $\arccos$-argument. For $f(\alpha, k)$, we consider 
\begin{align*}
 \frac{\partial}{\partial \alpha} \bigg( \frac{k^{2\alpha} + 1 - (1 \! - \! k)^{2\alpha}}{2 k^a} \bigg) \!  
= \frac{1}{2} k^{-\alpha} \Big( \underbrace{ \log(k)}_{< \ 0} [ \,\underbrace{  k^{2 \alpha} + (1 \! - \! k)^{2 \alpha} - 1  }_{\le \  0} \, ] - 2 (1 \! - \! k)^{2 \alpha} \underbrace{\log(1 \!- \!k)}_{< \ 0} \Big) > 0  .
\end{align*}
To see that in fact the expression in the square bracket is smaller or equal to zero, we exploit that for $\alpha \ge 0.5$ the function $k \mapsto k^{2\alpha}$ is superadditive, so that $1 = 1^{2 \alpha} = ( k + (1-k))^{2 \alpha} \ge k^{2 \alpha} + (1-k)^{2 \alpha}$. For $h(\alpha, k)$, we consider 
\begin{align*}
\frac{\partial}{\partial \alpha} \bigg(  \frac{1 - k^{2 \alpha} - (1-k)^{2 \alpha}}{2 k^\alpha (1-k)^\alpha} \bigg)  = - \frac{1}{2} (1-k)^{-\alpha} k^{-\alpha} \Big(
&\underbrace{\log(k)}_{< \ 0} \,  [ k^{2 \alpha} + 1 - (1-k)^{2 \alpha} ]  + \\
 &\underbrace{ \log(1- k)}_{< \ 0} \, [ (1-k)^{2 \alpha} + 1 - k^{2 \alpha} ]
 \Big) > 0 \ .
\end{align*}
Since $k, (1-k) \in (0,1)$ clearly the expressions in square brackets are positive so that the overall expression is positive too. In the respective regions, $f(\alpha, k)$ and $h(\alpha, k)$ are thus monotonically decreasing in $\alpha$.  \hfill\rlap{\hspace*{-1.em}$\qedsymbol$}

\section{Non-optimality of higher-degree branchings}
\label{appsec:nonopt}

This section supplements Sect.~\ref{subsec:nonopt} of the paper. First, we address the third and most involved 4-branching scenario in which a coupled 4-BP connects one source and three sinks (or equivalently 3 sources and 1 sink), see Fig.~\ref{fork3app}. We derive the inequalities listed in Proposition~\ref{prop:inequalities} and prove them analytically for a large subset of the parameter space. For the remainder we present a numerical argument (App.~\ref{app:numer_ineq}). Lastly, we show by induction how, given that coupled 4-BPs are never globally optimal, one can further rule out coupled $n$-BPs (with $n$ effective neighbors) for all~$n > 4$~(App.~\ref{app_higher}).

\subsection{Non-optimality of coupled 4-BPs between one source and three sinks}
\label{app:degree4}
\begin{figure}[b]
         \centering
         \includegraphics[width=0.18\textwidth]{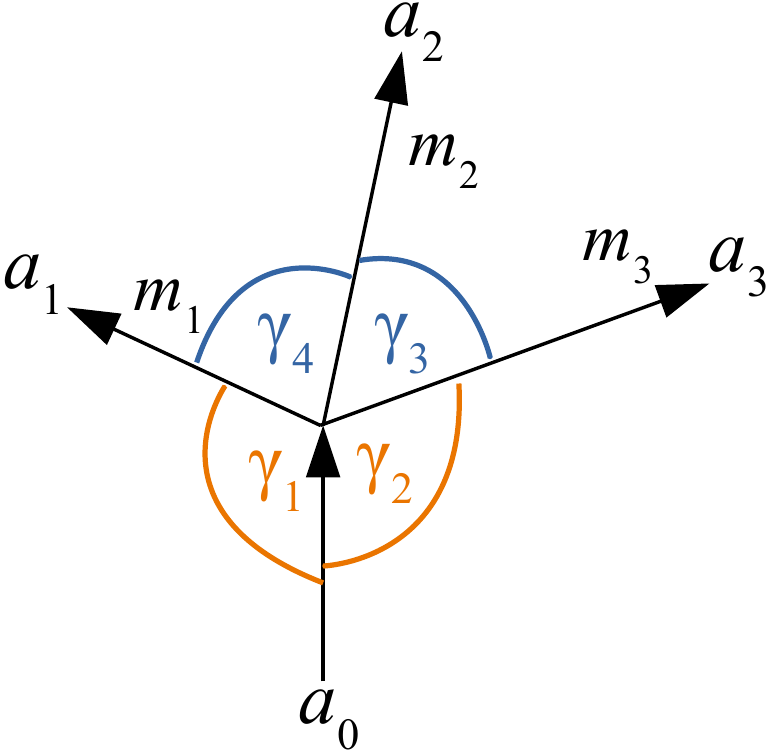}
        \caption{Coupled 4-BP between one source and three sinks.}
        \label{fork3app}
\end{figure}

Let us start by providing the derivation of Proposition~\ref{prop:inequalities}, which we repeat here for completeness:

\begin{prop}
Given a BOT problem with one source and three sinks, with demands $m_1, m_2, m_3$ as in Fig.~\ref{fork3app}, a coupled 4-BP away from the terminals cannot be globally optimal if at least one of the following inequalities holds true:
\begin{align*}
\Gamma &= h \Big(\frac{m_1}{m_1 + m_2} \Big) - f(m_1) + h \Big(\frac{m_3}{m_3 + m_2} \Big) - f( m_3) > 0, \\
\Gamma_{1,*} &= f(1 -m_*) + f\Big(1 - \frac{m_2}{1-m_*} \Big) - f(1-m_*-m_2) > 0, \\
\Gamma_{2,*} &= h \Big(\frac{m_*}{m_* + m_2} \Big) + f\Big(\frac{m_2}{1 - m_*} \Big) - h(m_*) > 0 
\end{align*}
where $*=1,3$. Note that {$\Gamma = \Gamma_{1,1} + \Gamma_{1,3} = \Gamma_{2,1} + \Gamma_{2,3}$}.
\end{prop}

Note that it is an important specification that we consider coupled 4-BPs \textit{away from the terminals}. For instance, in the OT case where $\alpha = 1$, all BPs are located at the terminals and coupled BPs with arbitrary number of neighbors may be globally optimal. For all following considerations, we assume that $\alpha \in [0,1)$.

\subsubsection{Derivation of the $\Gamma$-inequality}
\label{app:gamma_deriv}
WLOG, let us normalize the masses so that $m_1 + m_2 + m_3 =1$ and determine the necessary conditions under which all V-branchings are optimal:
\begin{align}
\gamma_1 &\ge \pi - f\Big(\alpha, 1- \frac{1-m_1}{1}\Big) = \pi - f(\alpha, m_1) \ , \nonumber \\
\gamma_2 &\ge \pi - f \Big( \alpha, 1 - \frac{1-m_3}{1}\Big) = \pi - f(\alpha, m_3) \ , \nonumber \\
\gamma_3 &\ge  h \Big(\alpha, \frac{m_3}{m_3 + m_2} \Big) \ ,  \nonumber \\
\gamma_4 &\ge h \Big(\alpha, \frac{m_1}{m_1 + m_2} \Big) \label{eq_cond4} \ . 
\end{align}
We intend to show that such a 4-BP can never be globally optimal by showing that for any combination of $\alpha$ and the masses $m_i$ the sum of the lower bounds is already larger than $2 \pi$. This is equivalent to proving that the following inequality holds true for all parameter combinations:
\begin{align}
\Gamma :=  h \Big(\alpha, \frac{m_1}{m_1 + m_2} \Big) - f(\alpha, m_1)  +  h \Big(\alpha, \frac{m_3}{m_3 + m_2} \Big)  -  f(\alpha, m_3) > 0 \label{eq_gamma} \ .
\end{align}
The inequality reflects that the problem setup is inherently symmetric under exchange of $m_1$ and $m_3$. WLOG, we assume that $m_1 \le m_3$. 

\subsubsection{Derivation of the $\Gamma_{1,*}$- and $\Gamma_{2,*}$-inequalities}
\label{app:transient}
\begin{figure}[b]
     \centering
     \begin{subfigure}[b]{0.1\textwidth}
         \centering
         \includegraphics[width=\textwidth]{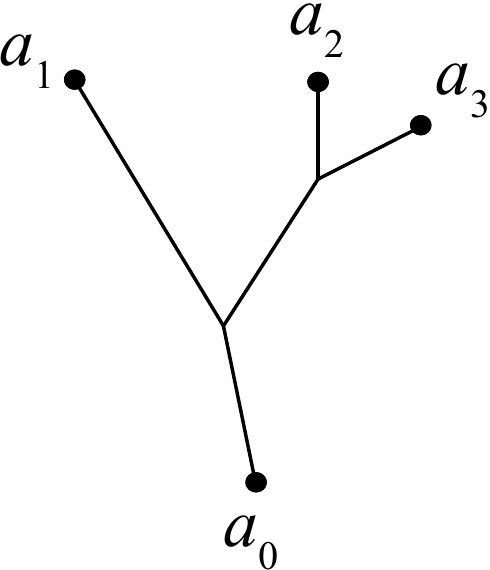}
         \captionsetup{labelformat=empty}
         \caption{Topology $T_1$}
         \label{topo1}
     \end{subfigure}
     \hspace{1.cm}
     \begin{subfigure}[b]{0.1\textwidth}
         \centering
         \includegraphics[width=\textwidth]{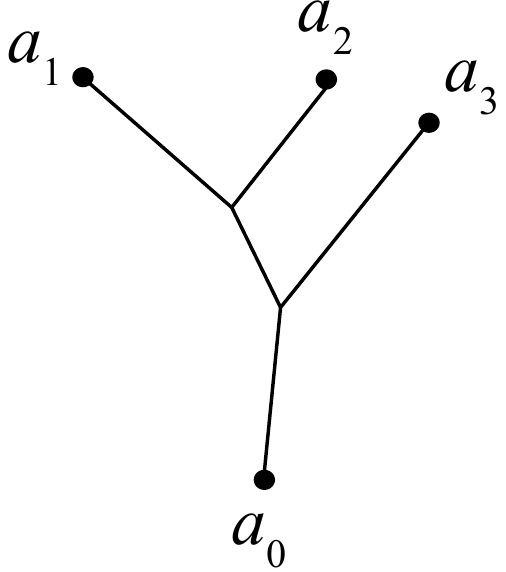}
         \captionsetup{labelformat=empty}
         \caption{Topology $T_2$}
         \label{topo2}
     \end{subfigure}
     \hspace{1.cm}
     \begin{subfigure}[b]{0.1\textwidth}
         \centering
         \includegraphics[width=\textwidth]{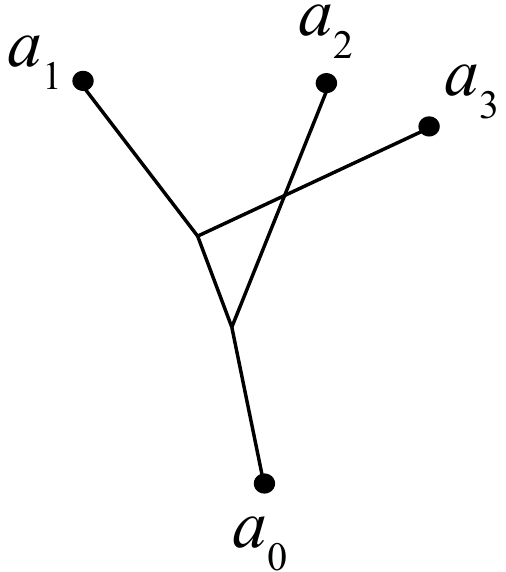}
         \captionsetup{labelformat=empty}
         \caption{Topology $T_3$}
         \label{topo3}
     \end{subfigure}
        \caption{The three distinct full tree topologies connecting four terminals.}
        \label{topo123}
\end{figure}

For a BOT solution to be globally optimal it means that it is the cheapest relatively optimal solution of all possible full tree topologies. In our case of four terminals, there are three distinct topologies, see Fig.~\ref{topo123}. Let us assume that a globally optimal 4-BP away from the terminals exists and denote the terminal positions by $a_i$. Then, for all three topologies $T_1$, $T_2$ and $T_3$, this branching point configuration is the ROS, since clearly for all $T_i$ a coupled 4-BP configuration can be realized by coupling the two branching points. 

\begin{figure}[!htp]
\centering 
\hspace*{-0.cm}
\includegraphics[scale=0.4]{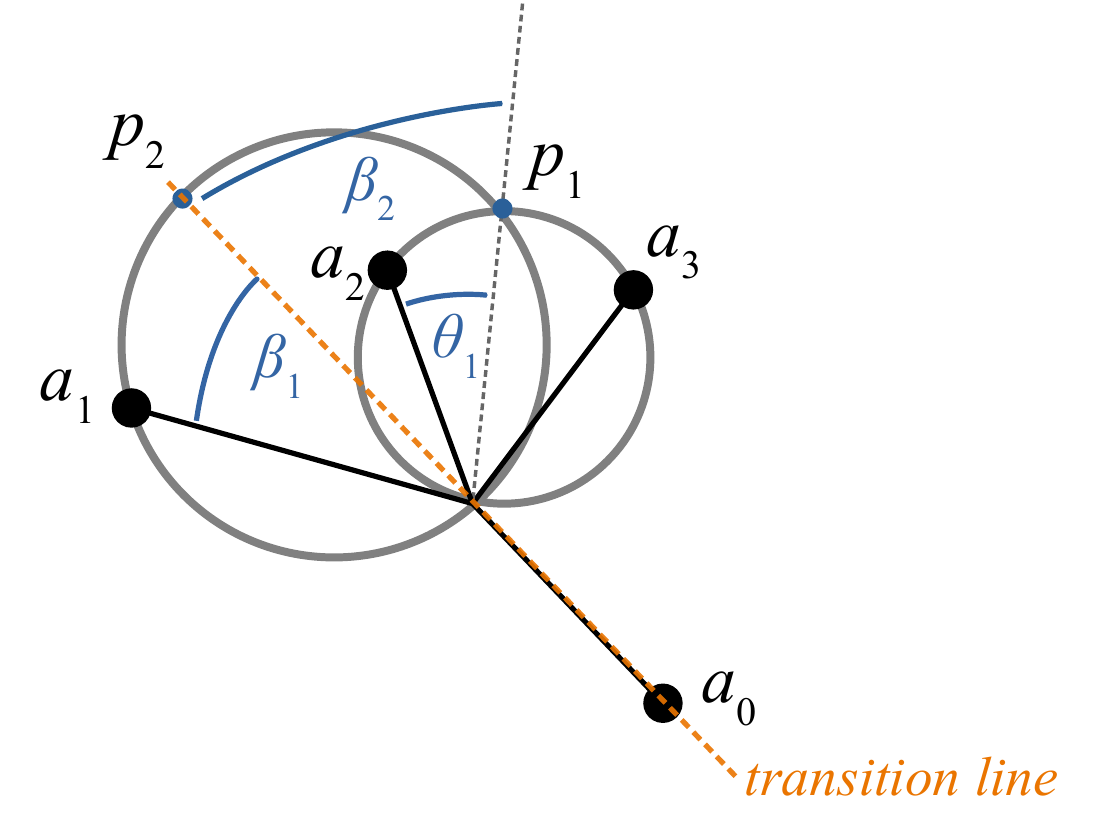} 
\caption{Transient coupled 4-BP with $a_0$ placed on the transition line of topology~$T_1$.}  
\label{just-angles}
\end{figure}

Let us investigate graphically under which conditions a coupled 4-BP provides the relatively optimal solution for the different topologies. Figure~\ref{just-angles} shows the pivot circle and pivot point construction for topology $T_1$, where $a_0$ has been chosen as root node. Let us refer to the line through $p_2$ and the intersection of the two pivot circles as \textit{transition line}, for the following reason: If $a_0$ was positioned to the left of the transition line, the ROS of $T_1$ would be non-degenerate, as shown in Fig.~\ref{recur4} for instance. For $a_0$ to the right of the transition line, the ROS of $T_1$ is given by a coupled 4-BP. Hence, the transition line marks the transition between a non-generate ROS of $T_1$ and a coupled 4-BP solution. Consequently, $a_0$ must lie to the right of the transition line of $T_1$. But, $a_0$ must simultaneously also lie on the appropriate side of the transition lines of the two other topologies $T_2$ and $T_3$. One can now argue that the root node $a_0$ can be moved along a continuous path onto the transition line of topology $T_1$, without crossing any of the other transition lines. In doing so, the coupled 4-branching stays relatively optimal for all three topologies. Most importantly, it thereby stays globally optimal. During this procedure the terminals $a_1$, $a_2$ and $a_3$ stay fixed so that the pivot points and pivot circles  as well as the transition lines stay exactly the same. For topology $T_1$ we have now arrived at a special case of coupled 4-branching, in which the V- and L-branchings are transient (see Def.~\ref{defn:trans}). Let us refer to such a BP configuration as \textit{transient} 4-branching. As a consequence, the angle $\gamma_4$ may be expressed in terms of the following branching angles (see Fig.~\ref{just-angles})
\begin{align} \label{eq_gamma4}
\gamma_4 = \beta_1 + \beta_2 - \theta_1 = h(\alpha, m_1) - f\Big(\alpha, \frac{m_2}{1-m_1} \Big) \ .
\end{align} 

All in all, the above argument shows that the existence of a globally optimal coupled 4-BP necessarily implies the existence of a globally optimal transient \mbox{4-branching}, which can be constructed by changing only the coordinates of one of the terminals. Clearly, for the globally optimal and transient 4-BP of topology $T_1$ all four necessary conditions for optimal V-branching must still apply. In order to generally rule out globally optimal coupled 4-BPs, it is therefore sufficient to show that at least one of the following two conditions is fulfilled for all parameter combinations $\alpha$ and $m_i$:
\begin{enumerate}
\item the sum of the lower bounds on $\gamma_i$ always exceeds $2 \pi$, 
\item condition~(\ref{eq_cond4}) is always incompatible with Eq.~(\ref{eq_gamma4}).  
\end{enumerate}
The first condition is equivalent to the following inequality, obtained by substituting the lower bound for $\gamma_4$ in (\ref{eq_gamma}) by (\ref{eq_gamma4}):
\begin{align} \label{eq_gamma1}
\Gamma_{1,1}(\alpha, m_1,m_2) := f(\alpha, 1 -m_1) + f\Big( \alpha, 1 - \frac{m_2}{1-m_1} \Big) - f(\alpha, 1-m_1-m_2) > 0   \ .
\end{align}
The second condition can be expressed as inequality simply by combining (\ref{eq_cond4}) and (\ref{eq_gamma4}):
\begin{align}
\Gamma_{2,1}(\alpha, m_1, m_2) := h \Big(\alpha, \frac{m_1}{m_1 + m_2} \Big) + f\Big( \alpha, \frac{m_2}{1 - m_1} \Big) - h(\alpha, m_1) > 0 \ .
\end{align}
Proving one of these two inequalities already suffices to rule out globally optimal 4-branching between one source and three sinks. Note that the three inequalities presented so far are not completely independent but are related via $\Gamma = \Gamma_{1,1} + \Gamma_{2,1}$. The above procedure of moving the root node onto the transition line of topology $T_1$ can be repeated exactly analogously for $T_2$. This results in the inequalities $\Gamma_{1,3}$ and $\Gamma_{2,3}$, which are of the exact same form except that $m_3$ appears in all places instead of $m_1$. Note that, in both cases, we have used that $a_0$ can be moved onto the transition lines of $T_1$ and $T_2$ without crossing the transition line of topology $T_3$. A justification for this is given in form of the following lemma.

\begin{lem} Starting from a globally optimal coupled 4-BP connecting the terminals $\{a_i\}$ and located away from all terminals, one terminal node may be moved along a continuous path onto the transition lines of topology $T_1$ and $T_2$ without crossing the transition line of topology $T_3$ first. \end{lem}

\noindent \textit{Proof.} Let us give a proof by contradiction. Assume that that $a_0$ could actually be moved along a continuous path onto the transition line of topology $T_3$ without touching any of the other two transition lines. Then, one may also move $a_0$ infinitesimally further across the transition line of topology $T_3$, such that the ROS of $T_3$ becomes non-degenerate. At the same time, for topology $T_1$ and $T_2$ the coupled 4-branching configuration stays relatively optimal. Now, since the ROS of $T_3$ deviates from the coupled 4-branching configuration, the 4-branching can no longer be globally optimal. Note that a coupled 4-BP can only be globally optimal if \textit{all three} topologies agree. Consequently, the ROS of $T_3$ must be globally optimal. However, the non-degenerate solution of $T_3$ will necessarily contain a cycle. To see this, let $b_1$ be the branching point connected to $a_0$ and $b_2$ the other branching point in $T_3$. After crossing the transition line of $T_3$, the edge $(b_1, b_2)$ at first has finite but infinitesimal length, so that either one of the edges $(b_2,a_1)$ or $(b_2,a_3)$ must intersect with $(b_1,a_2)$, thereby creating a cycle in the ROS of $T_3$ (cf. Fig.~\ref{topo3}). However, it was proven in~\cite{bernot2008optimal} that for $\alpha \in [0,1)$ cyclic BOT solution cannot be globally optimal, so that we have arrived at a contradiction. \hfill\rlap{\hspace*{-1.em}$\qedsymbol$}\\

To summarize, we have shown the following lemma:

\begin{lem} \label{lem:transient} 
The existence of any globally optimal 4-BP connecting one source and three sinks and located away from the terminals implies the existence of two globally optimal transient 4-branchings, one for topology $T_1$ and one for $T_2$. 
\end{lem}

\subsubsection{Analytical treatment of the inequalities}
\label{app:ineq_ana}
\paragraph{$\boldsymbol{\Gamma > 0 } \ $ for $\ \boldsymbol{\alpha \le 0.5}$:} For $\alpha \le 0.5$ we have $h \ge \pi/2$ and $f < \pi/2$ for all combinations of $m_i$ (see App.~\ref{app_fh}) and thus clearly $\Gamma > 0$ if $\alpha \le 0.5$. For all following arguments, we therefore assume that $\alpha >0.5$. The following analytical arguments all rely on the properties of $h$ and $f$, which are listed and proven in App.~\ref{app_fh}.

\paragraph{$\boldsymbol{\Gamma > 0} \ $ for $ \ \boldsymbol{m_1 \ge 1/4}$:} Using that $f(\alpha, k)$ is monotonically decreasing in $k$, the following loose lower bound suffices to demonstrate that $\Gamma > 0$ if $m_1 \ge 1/4$:
\begin{align*}
\Gamma &\ge 2 \ \min_k \, h(\alpha, k) - 2 f(\alpha, m_1) \\
&= 2 \arccos(2^{2 \alpha - 1} - 1) - 2 \arccos \bigg( \frac{m_1^{2\alpha} + 1 - (1-m_1)^{2 \alpha }}{2 m_1^\alpha} \bigg) \ ,
\end{align*}
where we have used that $f(m_1) \ge f(m_3)$ due to our assumption $m_1 \le m_3$ and that $h(\alpha, k)$ forms a minimum at $k=1/2$. For $m_1 = 1/4$, one finds that this expression is truly positive if and only if
\begin{align*}
9^\alpha < 2 \cdot 4^\alpha + 1 \ ,
\end{align*}    
which is fulfilled for all $\alpha \in [0,1)$ due to the subadditivity of the function $m \mapsto m^\alpha$, namely \mbox{$9^\alpha = (4 + 4 + 1)^\alpha < 4^\alpha + 4^\alpha +1$}. It follows that $\Gamma > 0$ also for all $m_1 > 1/4$ due to the monotonicity of $f$. 

\begin{figure}[b]
\centering 
\hspace*{-0.cm}
\includegraphics[scale=0.45]{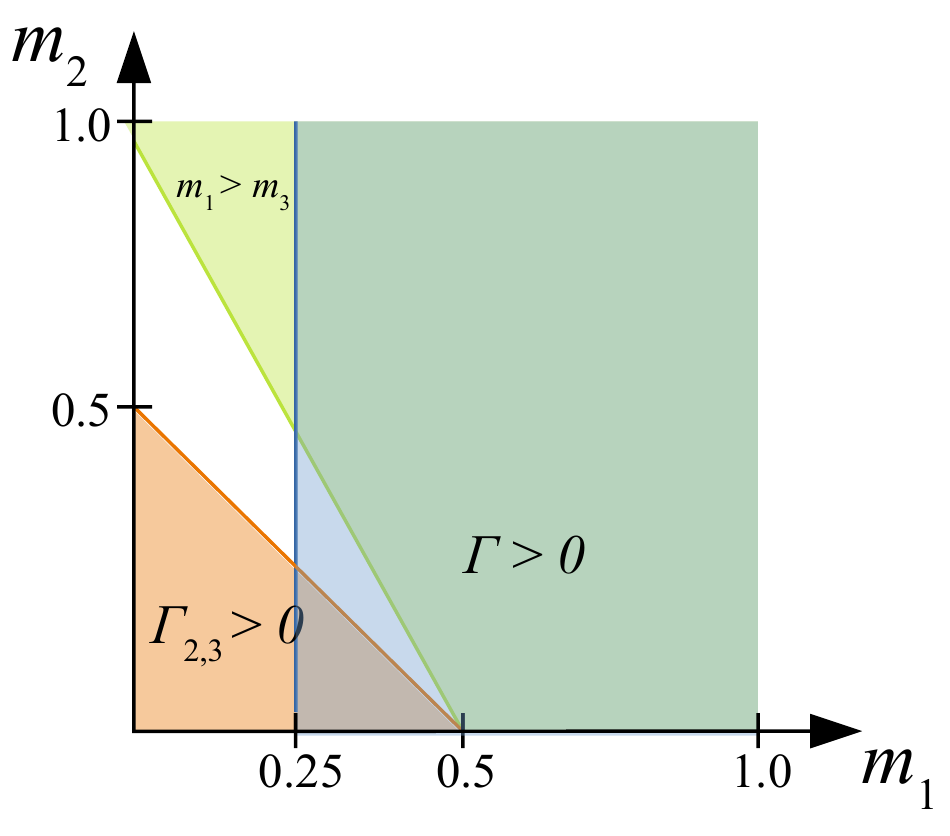} 
\caption{$m_1$-$m_2$ parameter space largely ruled out analytically. In the white region the inequalities are checked numerically.}  
\label{regions}
\end{figure} 

\paragraph{$\boldsymbol{\Gamma_{2,3} > 0} \ $ for $ \ \boldsymbol{m_3 \ge 1/2}$:} Inequalities $\Gamma_{2,3}$ is fulfilled if $m_3 \ge 1/2$, since
\begin{align}
\Gamma_{2,3}(\alpha, m_3, m_2) > h \Big(\alpha, \underbrace{\frac{m_3}{m_3 + m_2}}_{> \ m_3} \Big) - h(\alpha, m_3) \ge 0 \ \text{ for } \ m_3 \ge 1/2 \ ,
\end{align}
where we have used that for $\alpha > 0.5$, $h(\alpha, k)$ is monotonically increasing in $k$ for $k \in [0.5,1)$, as shown in Lem.~\ref{lem:dhdk}. The remaining parameter region, for which none of the inequalities have been shown yet, can be characterized by the following conditions:
\begin{align} \label{eq_remcond}
\alpha > 0.5 \, , \ \ m_1 < 0.25 \ \text{ and } \ m_2 \in [0.5 - m_1, 1 - 2 m_1] \ . 
\end{align}
The constraints for $m_3$ are implicitly represented, using that the normalization was chosen such that $m_1 +m_2 +m_3 = 1$. A visualization of the remaining region can be found in Fig.~\ref{regions}. Finally, we propose a simple and watertight numerical scheme, which can be used to rule out almost the entire remaining volume, already with little numerical effort.

\subsubsection{Numerical treatment of the remaining parameter space} 
\label{app:numer_ineq}
For the remaining parameter region characterized by the conditions in (\ref{eq_remcond}), we propose a numerical scheme which checks the inequality $\Gamma_{2,1}$ for all practically relevant parameter combinations of $m_1$, $m_2$ and $\alpha$. For this, we split the remaining volume into cuboids $ I = [\check \alpha, \hat \alpha] \times [\check m_1, \hat m_1] \times [\check m_2, \hat m_2]$ which are divided recursively into smaller cuboids based on the following octree scheme. For each cuboid, we determine a lower bound of $\Gamma_{2,1}$. This lower bound becomes tighter the smaller the cuboid $I$. If, for a cuboid, this lower bound is not yet positive, it is divided further into eight new cuboids by splitting each of the three intervals in half. This procedure is iterated until for all cuboids the lower bound is truly positive. For all $m_1, m_2, \alpha$ in a cuboid $I=[\check \alpha, \hat \alpha] \times [\check m_1, \hat m_1] \times [\check m_2, \hat m_2]$, the lower bound of $\Gamma_{2,1}$ is obtained by minimizing each summand individually:
\begin{align}
\Gamma_{2,1}(\alpha, m_1, m_2) \label{eq_lowerbound}
&\ge \underset{I}{\min} \ h \big(\alpha, \frac{m_1}{m_1 + m_2} \big) + \underset{I}{\min} \ f\big( \alpha, \frac{m_2}{1 - m_1} \big) - \underset{I}{\max} \ h(\alpha, m_1) \nonumber \\ 
&= h \Big(\hat \alpha, \frac{\hat m_1}{\hat m_1 + \check m_2} \Big)
+ f \Big( \hat \alpha,  \frac{ \hat m_2}{1 - \hat m_1} \Big)
- h(\check \alpha, \check m_1)
\end{align}
where we have used the crucial fact that for the remaining volume, characterized by (\ref{eq_remcond}), the functions $h(\alpha, k)$ and $f(\alpha, k)$ are monotonically decreasing in both arguments. The proof of these monotonicities can be found in App.~\ref{app_fh}. Note that the described procedure allows to rigorously confirm the inequality across a continuous region with finitely many evaluations.

For $\alpha \to 1$ or $m_1 \to 0$, the value of $\Gamma_{2,1}$ approaches zero, so that the above scheme cannot be used to proof the inequality for values arbitrarily close to these limits. However, if we restrict us to $m_1 > \delta$ and $\alpha < 1 - \epsilon$ for finite $\epsilon, \delta > 0$, the inequality $\Gamma_{2,1}$ can be shown for practically all parameter combinations with little numerical effort. For $\epsilon = \delta = 10^{-3}$, using the proposed scheme,  it was checked in only a few minutes that the lower bound in Eq.~(\ref{eq_lowerbound}) is larger than $10^{-4}$ everywhere in the remaining volume. To be numerically on the safe side, we have stopped splitting a cuboid not if the lower bound exceeded zero but set a suitable finite threshold, in this case $10^{-4}$. In other words, we have stopped splitting a cuboid if its respective lower bound was $>10^{-4}$. The smallest terms which occur during the arithmetic operations inside the functions $h$ and $f$ are of the order $\delta^2$, the largest terms are of order one. It is therefore safe to say that numerical errors at the order of the machine accuracy are negligibly small against the margin of $10^{-4}$ and we may say that all together globally optimal 4-BPs are ruled out, for all practical parameter combinations.  In principle, the presented scheme can be used to check the inequality $\Gamma_{2,1}$ up to even smaller $\epsilon$ and $\delta$. The Python code of the numerical scheme is made available at \url{https://github.com/hci-unihd/BranchedOT}.

\subsection{Non-optimality of five- and higher-degree branchings}
\label{app_higher}
In this section, we formally prove that globally optimal coupled $n$-BPs not coincident with a terminal can be ruled out in general, given that 4-BPs are not globally optimal. Lemma~\ref{lem:subprobs} states that a solution is not globally optimal if any subsolution is not globally optimal. It will therefore suffice to study the coupled BP as an isolated subproblem.
Let us start by proving the following corollary about the preservation of relative optimality under edge extensions for transient V- and L-branchings (see Def.~\ref{defn:trans}):\\

\begin{figure}[!htp]
\centering 
\hspace*{0.cm}
\includegraphics[scale=0.4]{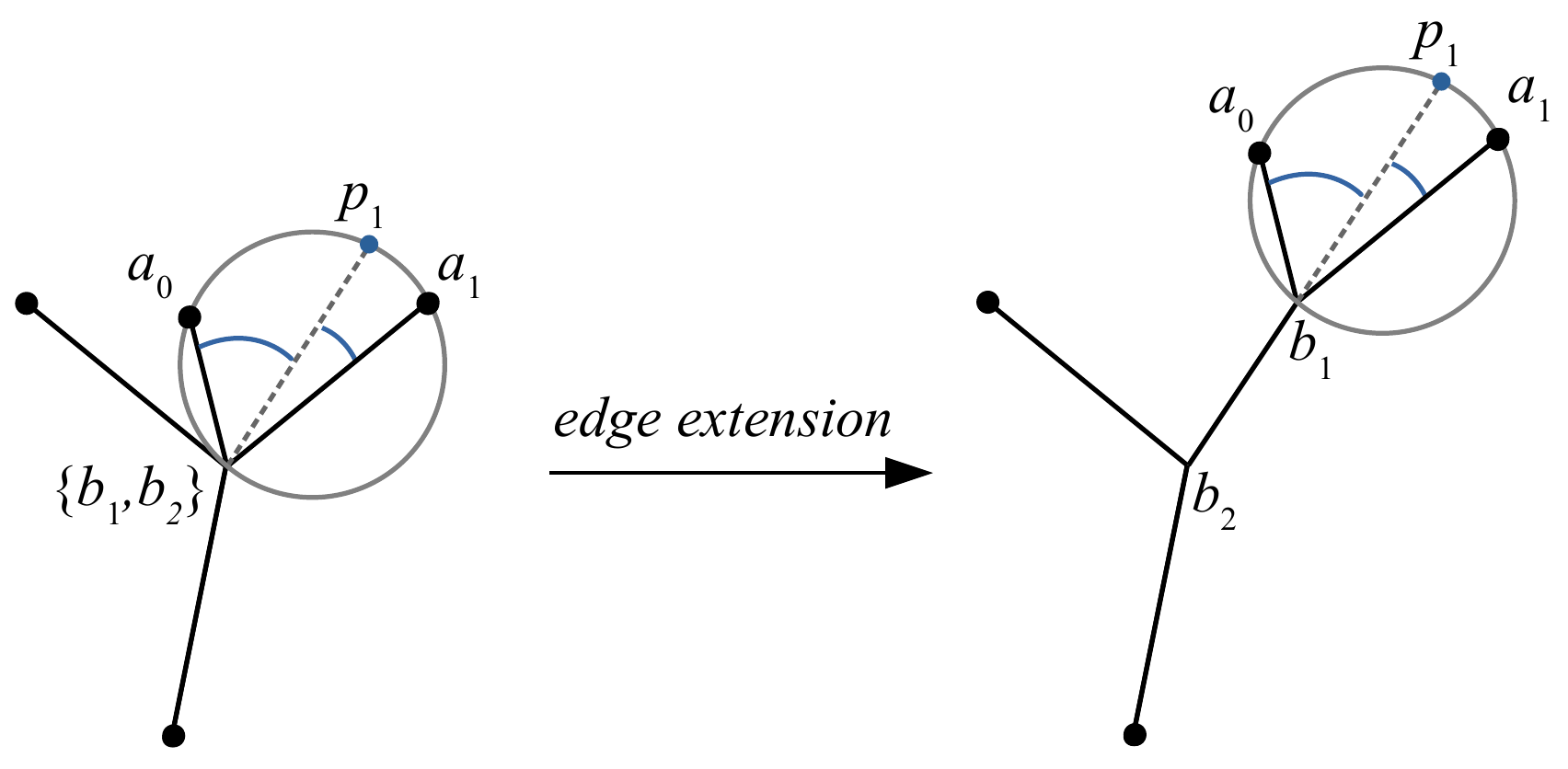} 
\caption{Extension of the edge $(b_1,b_2)$ in the direction of the pivot point $p_1$, preserving relative optimality.}  
\label{length_variation}
\end{figure}

\begin{cor} \label{cor:ext}
Consider a BOT problem with terminals $A=\{a_i\}$ and let the BP configuration $B = \{ b_i \}$ be relatively optimal for a given topology $T$. Let $B$ contain a transient V-branching between two terminals, say $a_0$ and $a_1$ and  denote the branching point connected to $a_0$ and $a_1$ by $b_1$. Otherwise, let $B$ not contain any strict L-or V-branchings. Let us further denote the branching point to which $b_1$ is coupled in a transient V-branching by $b_2$, as illustrated on the left side of Fig.~\ref{length_variation}. Then, there exists a direction in which the zero length edge $(b_1, b_2)$ can be extended to finite length $l>0$ (cf.~right side of Fig.~\ref{length_variation}) such that the new BOT problem (with shifted terminal positions) is solved relatively optimally by the new BP configuration (with shifted $b_1$).  
\end{cor}

\noindent \textit{Proof.} By assumption, the BP configuration of interest $B = \{b_i \}$ contains only Y-branchings and transient V- and L-branchings. Then, according to Lemma~\ref{lem_large_transient}, there exists a set of arbitrarily small displacements $\delta_i$, one for each branching point $b_i$, so that $B + \delta = \{ b_i + \delta_i \}$ is a non-degenerate BP configuration with arbitrarily small gradients $\left \| \nabla_{\! \! b_i} \, \mathcal{C}(B+\delta , A) \right \|$. Note that the notation $C(B,A)$ emphasizes that the cost function also depends on the terminal positions. Since the BP configuration $B + \delta$ is non-degenerate, any edge, in our case $(b_1,b_2)$, can be easily expanded in length without changing any of the branching angles, assuming that the extension preserves the direction of the edge and that the BPs and terminals are moved along correspondingly. As the gradient $\nabla_{\! \! b_i} \, \mathcal{C}$ depends only on the directions of the edges meeting at $b_i$ (i.e.~the branching angles), the gradient is not changed by this procedure. Let us summarize the shifted BPs by $B_{shift}$ and the shifted terminals by $A_{shift}$. Then, $\left \| \nabla_{\! \! b_i} \, \mathcal{C}(B + \delta , A) \right \| = \left \| \nabla_{\! \! b_i} \, \mathcal{C}(B_{shift} + \delta, A_{shift}) \right \|$ can also be made arbitrarily small as $\delta \to 0$. Let us prove by contradiction that $B_{shift}$ is the ROS of the BOT problem with terminals $A_{shift}$: Assuming that a different BP configuration $B_0 \neq B_{shift}$ was the ROS, there would exist a constant $\Delta>0$, so that $\mathcal{C}(B_{shift} , A_{shift}) = \mathcal{C}(B_0, A_{shift}) + \Delta$. However, since the cost is a convex function w.r.t.~the BPs, we have 
\begin{align*}
\mathcal{C}(B_0, A_{shift}) \ge \mathcal{C}(B_{shift} + \delta , A_{shift})
+ \langle \nabla_{\! \! b_i} \, \mathcal{C}(B_{shift} + \delta, A_{shift}), B_0 - (B_{shift} + \delta) \rangle \\
= \mathcal{C}(B_{shift} , A_{shift}) + \underbrace{O(\delta) + \langle \nabla_{\! \! b_i} \, \mathcal{C}(B_{shift} + \delta, A_{shift}), B_0 - (B_{shift} + \delta) \rangle}_{=:K} \, . 
\end{align*}
As $\delta \to 0$, the latter two terms summarized by $K$ can clearly be made arbitrarily small. In particular, there exists an $\delta > 0$, such that $\vert K \vert < \Delta/2$, which together with $\mathcal{C}(B_{shift} , A_{shift}) = \mathcal{C}(B_0, A_{shift}) + \Delta$ implies that
\begin{align*}
\mathcal{C}(B_0, A_{shift}) \ge \mathcal{C}(B_{shift}, A_{shift}) + K > \mathcal{C}(B_0, A_{shift}) + \Delta/2
\end{align*}
and we have thereby arrived at a contradiction, similarly to the reasoning in part~(c) of the proof in App.~\ref{sec:subopt}. \hfill\rlap{\hspace*{-1.em}$\qedsymbol$}\\

\begin{figure}[b]
\centering 
\hspace*{0.cm}
\includegraphics[scale=0.37]{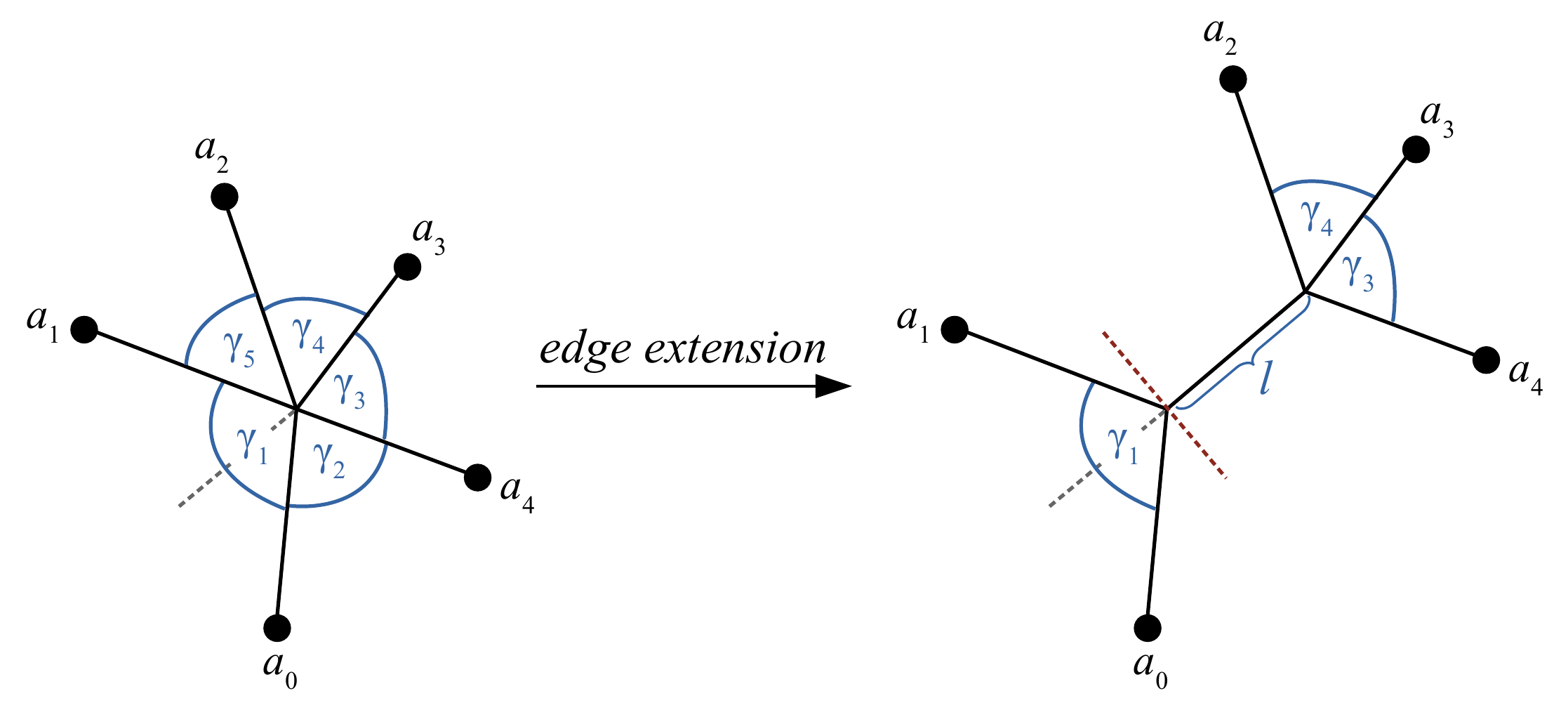} 
\caption{Edge extension to finite length $l > 0$ in a globally optimal and transient 5-branching with transient V-branching between $a_0$ and $a_1$. The  resulting ROS is split into two subsolutions, as indicted by the red dashed line.}  
\label{extended5}
\end{figure}

Coming back to the non-optimality of coupled $n$-BPs, let us, for concreteness, consider a globally optimal coupled 5-BP not coincident with a terminal. By induction, repeating the presented argument one can then rule out all globally optimal \mbox{$n$-BPs}. The proof is by contradiction, so let us assume a globally optimal 5-BP existed, between terminals $a_0$, $a_1$, $a_2$, $a_3$ and $a_4$. In terms of enclosed angles $\gamma_i$ (see Fig.~\ref{extended5}), it is a necessary condition that all $\gamma_i$ must exceed their respective lower bound, specified by the optimal V-branching conditions in Table~\ref{tab_corner} (Section~\ref{subsec:constr}). Again, for a 5-BP configuration to be globally optimal all possible full tree topologies must agree on the 5-BP configuration as their relative optimal solution. Starting from this configuration, one may continuously move one of the terminals, e.g.~$a_0$, such that the globally optimal coupled 5-BP starts to \textit{decouple}, meaning that $a_0$ is moved until for (at least) one of the possible topologies, say $\tilde T$, the 5-BP configuration no longer provides the ROS. This can always be achieved, for instance by bringing $a_0$ sufficiently close to $a_1$ so that a Y-branching between the two terminals becomes optimal. Similar to the 4-branching case, the ROS of such a topology $\tilde T$ in this moment becomes the globally optimal solution, as the other topologies are still in the 5-branching configuration which can only be globally optimal if \textit{all} topologies agree on it. The globally optimal solution of topology $\tilde T$ must exhibit one of the following two properties
\begin{enumerate}
\item it contains one Y-branching and a coupled 4-BP or
\item it is non-degenerate and contains only Y-branchings.
\end{enumerate}
In the first case, it would mean that a coupled 4-BP exists which is globally optimal as subgraph of a globally optimal solution (using the necessarily optimal substructure of Lem.~\ref{lem:subprobs}). This contradicts our assumption that 4-BPs are not globally optimal.

Regarding the second option, we proceed as follows. Let us move $a_0$ back to the point in which the coupled 5-BP configuration was still globally optimal but the ROS of $\tilde T$ is on the verge of decoupling into a non-degenerate branching configuration. Denote this special location of the terminals by $\{ a_i^* \}$. In this configuration an infinitesimal movement of $a_0$ away from $a_0^*$ can cause the ROS of $\tilde T$ to transition from coupled 5-branching to a non-degenerate BP configuration, very much analogous to the case of the transient 4-branching illustrated in Fig.~\ref{just-angles}. This means that in this configuration all V- and L-branchings are transient in the ROS of $T$ and WLOG we choose the labeling of the terminals such that the V-branching between $a_0^*$ and $a_1^*$ is transient, cf.~Fig.~\ref{extended5}. At this point, let us split the set of all possible full tree topologies $\mathcal T$ into the following subsets. The subset in which the terminals $a_0$ and $a_1$ are connected to a common branching point, say $b_1$, is denoted by $\mathcal{T}^{(0,1)}$. Let us label the BP to which $b_1$ is connected in these topologies by $b_2$. The ROS of all topologies $T \in \mathcal{T}^{(0,1)}$ for the current BOT problem contains a transient V-branching at branching point $b_1$. Moreover, let us single out a specific subset in $\mathcal{T}^{(0,1)}$, defined by the following condition:
\begin{align*}
\mathcal{T}^{(0,1)}_{trans} = \{T \in \mathcal{T}^{(0,1)} : \text{ROS of $T$ is transient if the terminals are located at $a_i^*$} \} \ .
\end{align*}
Visibly, the branching at $b_1$  appears as V-branching but it may also be seen as a Y-branching with a zero length stub. Let us now extend this zero length edge between $b_1$ and $b_2$ to finite length $l>0$ into the direction of the pivot point between $a_0^*$ and $a_1^*$, as explained in the proof of Corollary~\ref{cor:ext}. The two terminals $a_1$ and $a_2$ are shifted from $a_0^*$ and $a_1^*$ to $a_0(l)$ and $a_1(l)$ and Corollary~\ref{cor:ext} guarantees that the resulting BP configuration for all $T \in \mathcal{T}^{(0,1)}_{trans}$ solves the new BOT problem relatively optimally. For an illustration of the edge extension see Fig.~\ref{extended5}. Note that for all $T \in \mathcal{T}^{(0,1)}_{trans}$ this ROS is the same. Now split this ROS into two subsolutions as indicted in Fig.~\ref{extended5}. This induces two subproblems and subtopologies, as described in Def.~\ref{defn:split}. We focus on the upper right subproblem, consisting of four terminals. Note that any topology $T^{(4)}$ on this four terminal subproblem, may be induced as subtopology by a topology $T \in \mathcal{T}^{(0,1)}$. Let us distinguish the following two cases: a) On the four terminal subproblem the topology of the GOS, denoted by $T^{*(4)}$, is induced by a topology $T \in \mathcal{T}^{(0,1)}_{trans}$. Or b) $T^{*(4)}$ is induced by a topology $T \in \mathcal{T}^{(0,1)} \setminus \mathcal{T}^{(0,1)}_{trans}$. In case a) the ROS of $T^{*(4)}$ is given by the right subsolution in Fig.~\ref{extended5}, as the edge extension preserved the relative optimality for all $T \in \mathcal{T}^{(0,1)}_{trans}$. Hence, the globally optimal solution on the four terminal subproblem is given by a coupled 4-BP and we have arrived at a contradiction. \\
Otherwise, in case b), we do not know the ROS of $T^{*(4)}$ a priori, but it cannot be a coupled 4-branching configuration and must hence be non-degenerate. Crucially, it must be non-degenerate for any finite length extension $l > 0$, even if we consider the limit of $l \to 0$. But this means that the ROS of $T^{*(4)}$ already transitions from a coupled 4-BP into a non-degenerate ROS, if the terminals $a_0$ and $a_1$ are perturbed infinitesimally ($l > 0$ but infinitesimal). Consequently, for $l=0$ the ROS of $T^{*(4)}$ is transient. But this means that $T^{*(4)}$ can be induced as a subtopology of a transient topology $T \in \mathcal{T}^{(0,1)}_{trans}$ and we have arrived yet at another contradiction. \hfill\rlap{\hspace*{-1.em}$\qedsymbol$} \\

\section{BOT on two-dimensional Riemannian manifolds}
\label{app:mf}
In the following section, we prove that the optimal branching angles for Y-shaped branchings on two-dimensional manifolds are the same as the optimal branching angles derived for BOT in the Euclidean plane. An outline of the proof was given in Section~\ref{sec:approx-mf}. The same strategy, presented below for the optimal branching angles, can be used to generalize other necessary conditions for optimal BOT solutions to manifolds, as explained in Sect.~\ref{app:mf_other}. \\

\subsection{Optimal Y-branching on two-dimensional Riemannian manifolds}

For a Y-shaped branching at BP $b$ connecting the terminals $a_0$, $a_1$ and $a_2$ in Euclidean plane the cost function of BOT is given by
\begin{align*}
\mathcal C(b) = \sum_i m_i^\alpha \left \| a_i - b  \right \|_2 \ ,
\end{align*}
where the $m_i$ are the known edge flows. Let us consider a two-dimensional Riemannian manifold $\mathcal M$ embedded into $\mathbb{R}^3$, for which the metric is induced by the standard Euclidean inner product in $\mathbb{R}^3$. Let the geodesic distance be denoted by $d(x,y)$. The generalized cost function for 1-to-2 branching then reads
\begin{align*}
\mathcal C_M(b) = \sum_i m_i^\alpha  \, d( a_i, b) \ .
\end{align*} 
All points $b$ and $a_i$ now lie on the manifold and are assumed to have differing positions. In a solution which minimizes $\mathcal C_M$ the terminals $a_i$ are connected to $b$ via geodesics. For $b$ to be a valid solution to the BOT problem on the manifold, these geodesics must exist. We denote the geodesic which connects $b$ and $a_i$ by $v_i(\lambda) \in \mathcal{M}$, parametrized by the length $\lambda$. The tangent space at $b$ is denoted by $T_b \mathcal{M}$. Furthermore, let $\hat n_i$ be the unit tangent vectors of the geodesics $v_i(\lambda)$ at the branching point $b$, i.e.~$\hat n_i = \partial_\lambda v_i(\lambda) \big \vert_{\lambda = 0}$. WLOG, for all following considerations let us rotate and translate the manifold so that $b \in \mathbb{R}^3$ lies at the origin, i.e.~$b=0$,  and that the tangent space $T_b \mathcal{M}$ is equal to the $x_1$-$x_2$-plane of $\mathcal{R}^3$, i.e.~$T_b \mathcal{M} = \mathbb{R}^2 \times \{0\}$. \\

\paragraph{Restriction to a local subsolution on the manifold.} Since $\mathcal{M}$ is embedded into $\mathbb{R}^3$, there exists an $r > 0$ and an environment $U(r) \subset \mathbb{R}^3$ around $b=0$ such that the manifold $\mathcal{M} \cap U(r)$ can be represented as the graph of a function:
\begin{align*}
\mathcal{M} \cap U(r) = \{ (x, u(x)): x \in D(r) \subset \mathbb{R}^2 \} \ , 
\end{align*}    
where $u$ is a smooth, scalar function $u: D(r) \to \mathbb{R}$, defined on the disk of radius~$r$, denoted by $D(r) := \{ (x_1, x_2) \in \mathbb{R}^2 : \left \| (x_1,x_2)^T \right \|_2 \le r \}$. Note that due to the mentioned rotation and translation of $\mathcal{M}$, we have $u(0) = 0$ and $\nabla u(0) = 0$, where $\nabla u$ denotes the gradient of $u$. The existence of such a function $u$ is guaranteed by the implicit function theorem. A formal proof can be found in John M. Lee's book~\cite{lee2018introduction} in Proposition 8.24. Further, let us define the orthogonal projection~$\sigma$ from the manifold onto the $x_1$-$x_2$-plane as
\begin{align}
\sigma: \mathcal{M} \cap U(r) \to D(r) \times \{0\}, \ 
\begin{pmatrix} x_1 \\ x_2 \\ u(x_1,x_2) \end{pmatrix} \mapsto
\begin{pmatrix} x_1 \\ x_2 \\ 0 \end{pmatrix} \ .
\end{align}
WLOG, $r$ is chosen sufficiently small so that this projection is bijective. Now, Taylor's theorem states that $u$ can be approximated by the following expansion around $x = 0$:
\begin{align} \label{eq:u_quad}
u(x) = \underbrace{u(0)}_{= \ 0} + \langle \underbrace{\nabla u(0)}_{= \ 0}, x \rangle + O(\left \| x \right \|_2^2) \in O(r^2) \ ,
\end{align} 
where $\langle \cdot , \cdot \rangle$ is the standard Euclidean inner product and we have introduced the Big-O notation. A term is of order $O(r^2)$ if it goes to zero for $r \to 0$ at least as fast as $r^2$, or more formally: 
\begin{align*}
p(r) \in O(q(r)) \ \Leftrightarrow \ \lim_{r \to 0} \frac{p(r)}{q(r)} = c
\end{align*}  
for some finite constant $c$. Consequently, a point $a = (x,u(x))$ in $\mathcal{M} \cap U(r)$ and its projection onto the plane $\sigma(a)$ agree to first order, i.e.
\begin{align} \label{eq_projrel}
\left \| \sigma(a) - a \right \|_2 = u(x) \in O(r^2) \ .
\end{align} \\

One of the key ingredients when transferring BOT problems from two-dimensional surfaces to the tangent plane $T_b \mathcal{M}$ is the following Lemma about the difference between the Euclidean distance and the geodesic distance:

\begin{lem}[Relation between geodesic and Euclidean distance] \label{lem:dist-mf}
Let $r$ be a small radius, which characterizes the environment $U(r)$ around the origin $b = 0$ on a two-dimensional Riemannian manifold as described above. Let $a \in \mathcal{M} \cap U$ be a point in this environment, located at \mbox{$a = (x,u(x)) \in \mathbb{R}^3$} for some $x \in D(r)$ and $u$ as above. Then, the geodesic distance can be expressed through the Euclidean distance as
\begin{align}
d(a,b) = \left \| a - b \right \|_2 + O(r^3) \ . 
\end{align}  
\end{lem}
 
\noindent \textit{Proof.} 
Let us first note that $d(a,b) \ge  \left \| a - b \right \|_2$. It is hence sufficient to show that $d(a,b) \le \left \| a - b \right \|_2 + O(r^3)$. Let us consider the following curve $\gamma(t) = (tx, u(tx))$ on the manifold $\mathcal M$, which for $t \in [0,1]$ connects $a = (x,u(x))$ and $b = 0$. In general, $\gamma(t)$ is not a geodesic between $a$ and $b$. Thus, the length of $\gamma(t)$ provides an upper bound to $d(a,b)$. For the calculation of the length we use that $\gamma \hspace{0.03cm} '(t) = (x, \langle \nabla u(tx), x \rangle)$ and simply integrate $\left \| \gamma \hspace{0.03cm} '(t) \right \|_2$ along the curve in $\mathbb{R}^3$, since the metric of the embedding is induced by the standard Euclidean inner product in $\mathbb{R}^3$:
\begin{align*}
\frac{d(a,b)}{\left \| a - b \right \|_2} &\le \frac{1}{\sqrt{\left \|  x \right \|^2_2 + u^2(x)}} \int_0^1 \left \| \gamma \hspace{0.03cm} '(t) \right \|_2 \mathrm d t \\
&\le \frac{1}{\left \| x \right \|_2} \int_0^1 \sqrt{\left \|  x \right \|^2_2 + \vert \langle \nabla u(tx), x \rangle \vert^2} \ \mathrm d t \\
&= \int_0^1 \sqrt{1 + \left| \Big \langle \nabla u(tx), \frac{x}{\left \| x \right \|_2} \Big \rangle \right|^2} \ \mathrm dt \\
&\le 1 + \frac{1}{2} \int_0^1  \left| \Big \langle \nabla u(tx), \frac{x}{\left \| x \right \|_2} \Big \rangle \right|^2 \mathrm dt \\
&\le 1 + \frac{1}{2} \int_0^1 \left \| \nabla u(tx) \right \|_2^2 \ \mathrm dt \ .
\end{align*}
For the last step we have used the Cauchy-Schwarz inequality. Furthermore, we have used that $\sqrt{1 + x^2} \le 1 + x^2 / 2$. The only thing left to show is that the remaining integral is $O(r^2)$, as $\left \| a - b \right \| = ( \left \|  x \right \|^2_2 + u^2(x) )^{1/2} \in O(r)$, since $\left \| x \right \|_2 \le r$ and \mbox{$u(x) \in O(r^2)$}, cf.~Eq.~(\ref{eq:u_quad}). For that, let us consider the Taylor expansion of the $i$-th component of $\nabla u(tx)$ around $0$. For $i=1,2$ one has
\begin{align*}
\partial_i \, u(tx) = \underbrace{\partial_i \, u(0)}_{= \ 0 } + \partial_j \partial_i \, u(0) \ t x_j + O(r^2) \ ,
\end{align*} 
where the sum in $j$ is implicit and runs over both indices $j=1,2$. 
And thus:
\begin{align*}
\frac{1}{2} \int_0^1 \left \| \nabla u(tx) \right \|_2^2 \ \mathrm dt = \frac{1}{6} \, \partial_j \partial_i \, u(0) \ x_j \  \partial_k \partial_i \, u(0) \ x_k + O(r^2) \ ,
\end{align*}
where again the sum over all index pairs is implied. Since for all components $x_i$ we have $x_i \le r$, this completes the proof. \hfill\rlap{\hspace*{-1.em}$\qedsymbol$} \\

\begin{figure}[b]
\centering 
\hspace*{0.cm}
\includegraphics[width=0.35\textwidth]{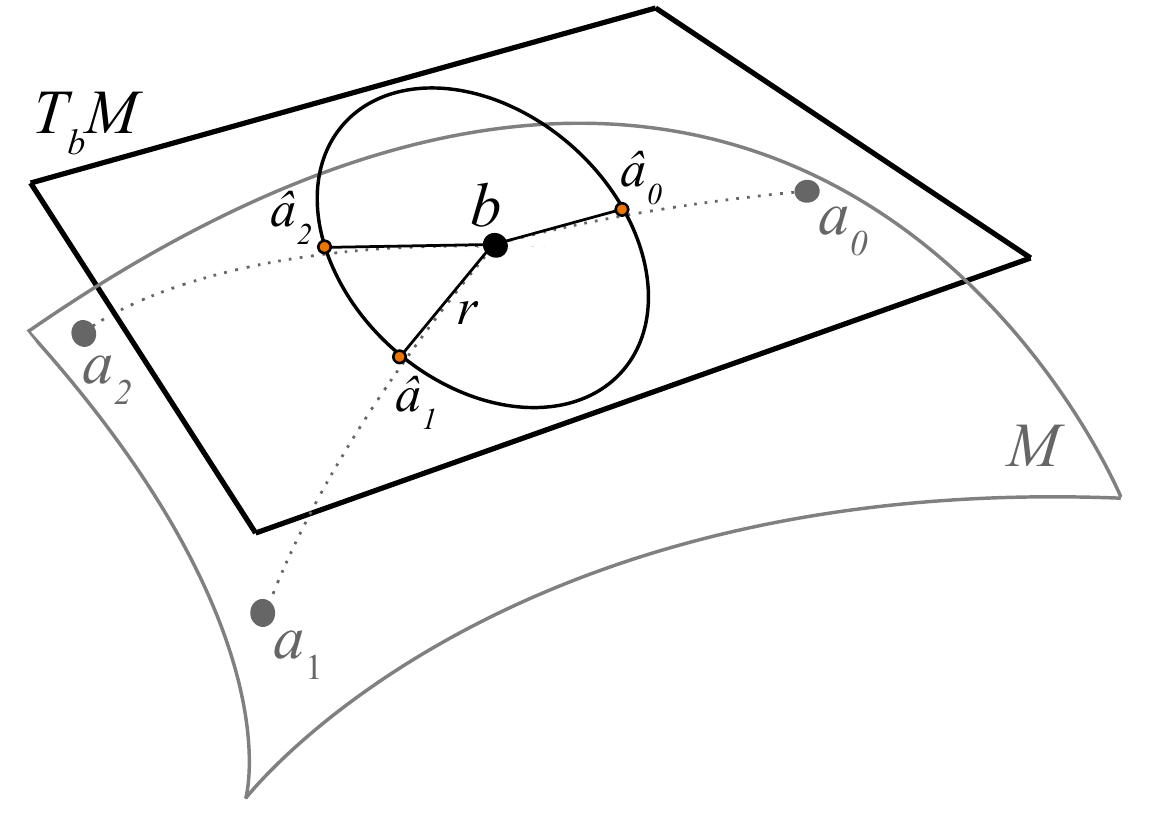} 
\caption{The tangent vectors of the geodesics (dotted) in $b$ define the position of the terminals $\hat a_i$ of the subproblem on the flat disk \mbox{$D(r) \subset T_b \mathcal{M}$}.}  
\label{tangent-plane2}
\end{figure}

\paragraph{Projection of a subproblem onto the tangent plane.} Coming back to the 1-to-2 BOT problem on the manifold, let us project the geodesics $v_i$ onto the $x_1$-$x_2$-plane using the orthogonal projection~$\sigma$ from above, in order to transfer the BOT problem from the manifold onto the tangent plane. WLOG, we choose the orientation of the $x_1$ and $x_2$ axis such that $\hat n_0$ points along the $x_1$-axis. Then, one may easily check that the implicit function theorem guarantees that for sufficiently small $r$ the projected geodesic~$\sigma(v_0)$ can be represented by a graph $(x_1, w(x_1))$ with smooth $w: (-\epsilon,r+\epsilon) \to \mathbb{R}$. Note, that the projected geodesic~$\sigma(v_0)$ may be considered on the open interval with $\epsilon >0$ such that all derivatives with respect to $x_1 \in [0,r]$ are well-defined. Due to our special orientation of the $x_1$-$x_2$-plane, we have $w(0)=0$ and $w'(0) = 0$. Let us now define the following two points inside the tangent plane: 
\begin{enumerate}
\item Define $\hat a_0$ as the point which lies in the direction of $\hat n_0$ at a distance $r$ away from the origin, i.e.~$\hat a_0 = r \cdot \hat n_0 = (r,0)$. Similarly, define $\hat a_i = r \cdot \hat n_i$. These will be the three terminals of the BOT subproblem of interest in the Euclidean plane. For an illustration see Fig.~\ref{tangent-plane2}. 
\item Define $a_{\perp,0} = (r, w(r))$ as special point on the projected geodesic $\sigma(v_0)$. From an analogous construction using the projected geodesics $\sigma(v_1)$ and $\sigma(v_2)$, we obtain the $a_{\perp,i}$ also for $i=1,2$. No explicit representation of these will be necessary.
\end{enumerate}
What is however important is that the two defined points agree up to linear order, in the sense that $\left \| \hat a_i - a_{\perp,i} \right \|_2 \in O(r^2)$ for all $i=0,1,2$. This is shown for $i=0$, by Taylor expansion of $w(x_1)$ around $x_1 = 0$, but holds of course equally true for $i=1,2$:
\begin{align} \label{eq:proj2}
\left \| \hat a_0 - a_{\perp,0} \right \|_2 = w(r)= \underbrace{w(0)}_{= \ 0} + \underbrace{w'(0)}_{= \ 0} \, r + O(r^2) \in O(r^2) .
\end{align}
Note that since the points $a_{\perp,i}$ lie on the projected geodesics, after an inverse projection, the $\sigma^{-1}(a_{\perp,i})$ mark special points on the unprojected geodesics $v_i \in \mathcal M$. These will be the terminals of the subproblem of interest on the manifold. Combining Eq.~(\ref{eq_projrel}) and (\ref{eq:proj2}), we can relate $\sigma^{-1}(a_{\perp, i}) \in \mathcal M$ to the terminals $\hat {a}_i$ of the BOT problem of interest in the plane by
\begin{align} \label{eq_terminals}
\left \| \sigma^{-1}(a_{\perp, i}) - \hat{a}_i \right \|_2 \in O(r^2) \ ,
\end{align}
Note that all steps above can be repeated for radii smaller than the chosen $r$. In the following we will decrease the scale $r$ of the two problems and therefore indicate the $r$-dependence in $\sigma^{-1}(a_{\perp, i}(r))$ and $\hat{a}_i(r)$ explicitly.    \\

\paragraph{Improved subsolutions on the manifold by projection from the tangent space.} Now, follows the main line of arguments to show that $b$ does not connect $a_i \in \mathcal M$ at minimal cost if the geodesic angles in the tangent space $T_b \mathcal{M}$ are different from the optimal angles in the Euclidean plane. We start by considering the subproblem on the manifold with terminals $\sigma^{-1}(a_{\perp, i}) \in \mathcal M$. The generalized cost function of this subproblem reads: 
\begin{align*}
\mathcal C_M(b) &= \sum_i m_i^\alpha \, d(\sigma^{-1}(a_{\perp,i}(r)), b) \ . 
\end{align*}
We can now use Eq.~(\ref{eq_terminals}) to express the terminals on the manifold through the terminals $\hat a_i$ in the tangent plane and express the geodesic distance in terms of the Euclidean distance, based on Lem.~\ref{lem:dist-mf}. All additional terms are at least of order $O(r^2)$ and we have
\begin{align} \label{eq_conthere}
\mathcal C_M(b) = \underbrace{\sum_i m_i^\alpha \left \| \hat{a}_i(r) -  b \right \|_2}_{=: \ \mathcal C_b(r)} + \ O(r^2) \ .
\end{align}  
Crucially, with $\mathcal C_b(r)$ we have arrived at the cost of the BOT problem which purely lives on the tangent space. It is the cost of the solution in which $b$ is connected via straight lines to the terminals $\hat a_i(r) = r \cdot \hat n_i$ at a distance $r$ away from $b$, as shown in Fig.~\ref{tangent-plane2}. By assumption, this solution is not relatively optimally since the angles (or equivalently the directions $\hat n_i$) are assumed to deviate from the optimal angles in the Euclidean plane. This means that a branching point $b^* \in D(r)\times \{0 \}$ exists which provides a better solution with cost denoted by $\mathcal C_{b^*}(r) < \mathcal C_b(r)$. From the definition of $\hat a_i = r \cdot \hat n_i$, we see that two BOT problems within disks of different radii $D(r)$ and $D(\nu \cdot r)$ are related simply by rescaling the coordinates. Under rescaling of coordinates the Euclidean distance between points and consequently also the cost functions changes proportionally, i.e.~$\mathcal{C} \to \nu \mathcal{C}$. Thus, there exist non-negative constants $M_1$ and $M_2$, so that $\mathcal C_b(r) = M_1 \cdot r$ and $\mathcal C_{b^*}(r) = M_2 \cdot r$. Let us distinguish the following two cases: \\

a) $\ \ \ \mathcal C_{b^*}(r) = \sum_i m_i^\alpha \left \| \hat{a}_i(r) -  b^* \right \|_2 = 0$. This special case looks unusual, but is in principle possible, e.g.~if all terminals $a_i$ and $b$ lie on a common geodesic. In the case, where $C_{b^*}(r) = 0$, we project $b^*$ onto the manifold using $\sigma^{-1}$ and go backwards in the above steps. We find that 
\begin{align*}
0 = \sum_i m_i^\alpha \left \| \hat{a}_i(r) -  b^* \right \|_2 = 
\underbrace{\sum_i m_i^\alpha \, d( \sigma^{-1}(a_{\perp,i}(r)) ,  \sigma^{-1}(b^*))}_{= \ \mathcal C_M(\sigma^{-1}(b^*)))} + O(r^2) \, . 
\end{align*}
Clearly, a sufficiently small $r > 0$ exists so that the terms contained in $O(r^2)$ are much smaller than $M_1 \cdot r$, so that
\begin{align*}
\mathcal C_M(\sigma^{-1}(b^*))) = 0 + O(r^2) < M_1 \cdot r + O(r^2) = \mathcal C_M(b) \ .
\end{align*}
This proves that the projection of $b^*$ onto the manifold provides a cheaper cost solution on the manifold than $b$.

b) $\ \ \ \mathcal C_{b^*}(r) > 0$, and thus $M_2 >0$. In this case, let us write the ratio of the two costs as $ \mathcal C_b(r)/ \mathcal C_{b^*}(r) = M_1 / M_2 =: 1 + \kappa$ for some $\kappa > 0$. Again, $b^*$ and its projection onto the manifold agree to first order, that is $\left \| \sigma^{-1}(b^*) - b^* \right \| \in O(r^2)$. We now show that $\sigma^{-1}(b^*)$ provides a better solution to the BOT problem on the manifold than~$b$. We proceed from Eq.~(\ref{eq_conthere}), using that $\mathcal C_{b^*}(r)$ is linear in $r$:
\begin{align*}
\mathcal C_M(b) &= C_b(r) + O(r^2)  = (1 + \kappa) \, \mathcal C_{b^*}(r) + O(r^2) \nonumber \\
&= \sum_i m_i^\alpha \left \| \hat{a}_i(r) -  b^* \right \|_2 + \kappa M_2 \cdot r + O(r^2) \nonumber \\
&= \underbrace{\sum_i m_i^\alpha \, d( \sigma^{-1}(a_{\perp, i}(r)) ,  \sigma^{-1}(b^*))}_{= \ \mathcal C_M(\sigma^{-1}(b^*))} + \kappa M_2 \cdot r  + O(r^2) \ .
\end{align*}
In the last step, we have projected inversely onto the manifold and replaced the Euclidean distance by the geodesic distance (all with differences at least of $O(r^2)$). We have therefore arrived at the cost of the new solution on the manifold. Since the term $\kappa M_2 \cdot r$ is positive, we conclude that $r$ may always be chosen sufficiently small, so that the linear term dominates over higher-order terms and we have:
\begin{align*}
\mathcal C_M(\sigma^{-1}(b^*)) <  \mathcal C_M(b) \ .
\end{align*}  
Together with Lem.~\ref{lem:subprobs} on the necessarily optimal substructure of all subsolutions, this concludes the proof that any optimally placed branching point on a two-dimensional embedded Riemannian manifold must exhibit the same optimal branching angles as in the Euclidean plane.\hfill\rlap{\hspace*{-1.em}$\qedsymbol$}

\subsection{Other local properties of optimal BOT solutions on manifolds}
\label{app:mf_other}
The logic of the proof outlined above can be easily extended to the conditions for optimal V- and L-branching. For BOT problems in the Euclidean plane, Table~\ref{tab_corner} lists these conditions under which the optimal BP position in a 1-to-2 branching coincides with one of the terminals. Let us now consider a 1-to-2 BOT solution on a two-dimensional manifold in which $b$ is located at the terminal $a_0$ and is connected to the terminals $a_1$ and $a_2$ via the geodesics $v_1$ and $v_2$ respectively. Imagine that in the tangent plane $T_b \mathcal{M}$ the angle enclosed by these geodesics does not fulfill the optimal V-branching criterion. In this case, one may again consider a sufficiently small disk of radius $r$ around $b = a_0$ in the tangent space $T_b \mathcal{M}$ and project the corresponding subproblem from the manifold onto this disk. In the Euclidean plane, an improved BP location $b^*$ must exists, since $b = a_0$ is the optimal solution if and only if the V-branching condition is fulfilled. All arguments about the scaling of the cost improvement apply as described above and the projection $b^*$ back to the manifold will provide an improved solution to the subproblem on the manifold if $r$ is chosen sufficiently small. 

Even more so, the same reasoning can also be applied to generalize our results regarding the non-optimality of higher-degree branchings. Given a coupled $n$-BP at position $b$ not coincident with a terminal, one may again consider a sufficiently small region around $b$ on the manifold and project the corresponding subsolution onto $T_b \mathcal{M}$. Exactly analogous to the previous arguments, improving the topology locally in the plane and projecting back to the manifold eventually results in an improved solution on the manifold.

\subsection{The practical side of BOT on manifolds}
\label{app:mf_practical}
 Although much of the theory generalizes nicely to Riemannian manifolds, the generalization of the practical algorithms is highly non-trivial. Unlike in Euclidean space, realizing the optimal branching angles is a necessary but no longer sufficient condition for relatively optimal solutions. For instance, on the sphere the meridians of three terminals located in the southern hemisphere at longitudes $0^\circ$, $120^\circ$, $240^\circ$ will intersect at both poles at angles of $120^\circ$, which is the optimal angle for $\alpha=0$. Nonetheless, only the south pole is the optimal branching point. In essence, the geometry optimization aims to assign simultaneously to each branching point the coordinates of the weighted geometric median of its neighbors, a problem that is considered in~\cite{fletcher2008robust}. The topology optimization presented in Sect.~\ref{sec:sim-ann} could be easily generalized to manifolds if the geodesic distance can be computed. Due to these obstacles, previous works of the Steiner Tree problem ($\alpha = 1$) have focused mostly on the sphere as important special case~\cite{dolan1991minimal}.

\section{Algorithms}
\label{appsec:algo}
\paragraph{Hardware and code availability.} \label{hardware}
Python code for all experiments can be found at \url{https://github.com/hci-unihd/BranchedOT}. For the different experiments, a single machine with 56 CPUs (Intel(R) Xeon(R) CPU E5-2660 v4 @ 2.00GHz) and 256GB RAM was used. Execution times for all described experiments lie (at most) in the order of hours. A more detailed estimate can be obtained from the performance statistics reported in Fig.~\ref{smith} and Fig.~\ref{iter_conv}. 

\subsection{Geometric construction of relatively optimal solutions for BOT with multiple sources}
\label{app:constr}

\begin{figure}[H]
\centering 
\hspace*{0.cm}
\includegraphics[width=0.14\textwidth]{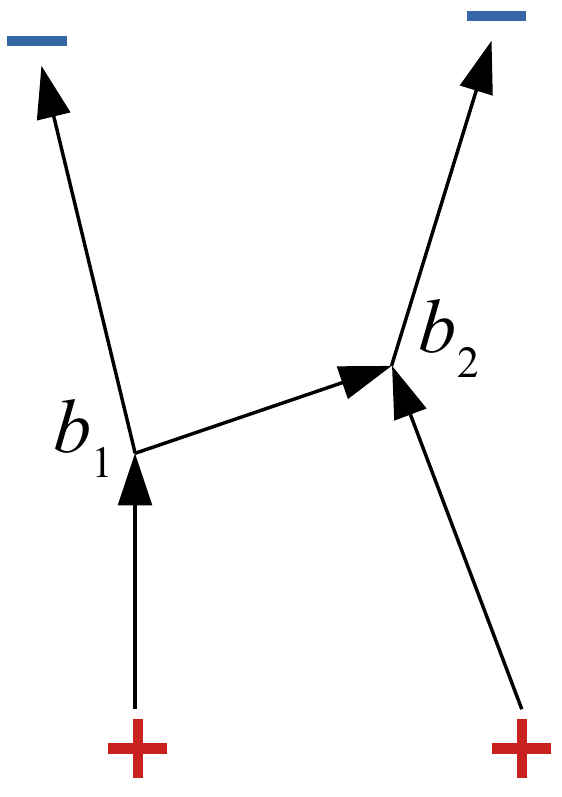}
\caption{Simple setup with symmetric branching and asymmetric branching, see Sect.~\ref{subsec:constr}.}  
\label{converg}
\end{figure}

In Sect.~\ref{subsec:constr} we have presented the the exact geometric construction of relatively optimal solutions based on~\cite{bernot2008optimal,gilbert1967minimum} and its generalization to the case of BOT with multiple sources. We have thereby solved the open problem 15.11 in~\cite{bernot2008optimal}, for which an example is illustrated in Fig.~\ref{converg}.

\begin{figure}[H]
	\centering 
     \begin{subfigure}[b]{0.3\textwidth}
         \centering
         \includegraphics[width=\textwidth]{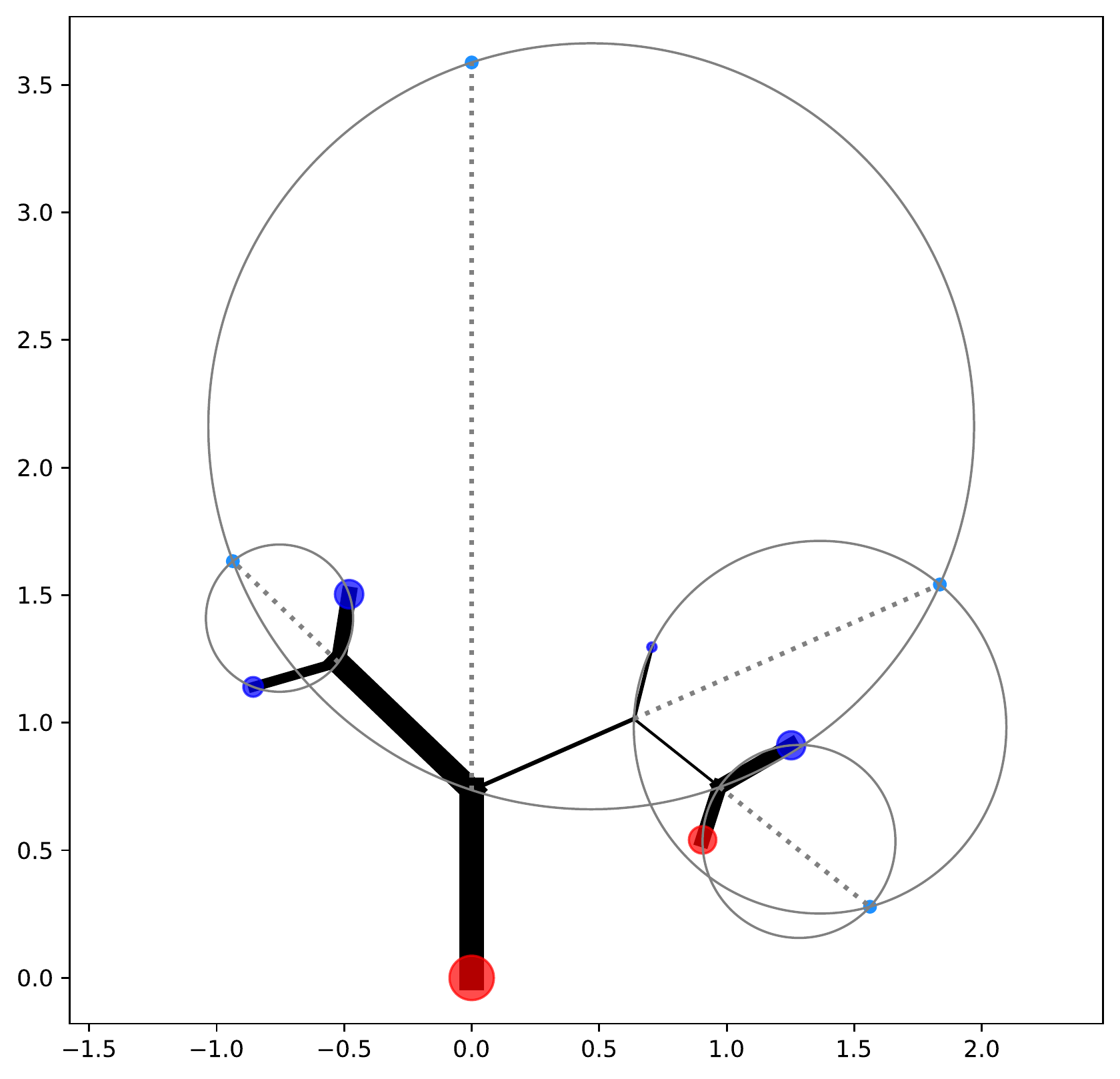}
         \caption{$\ \ \alpha = 0.1$}
     \end{subfigure}
     \hspace{1.5cm}
     \begin{subfigure}[b]{0.3\textwidth}
         \centering
         \includegraphics[width=\textwidth]{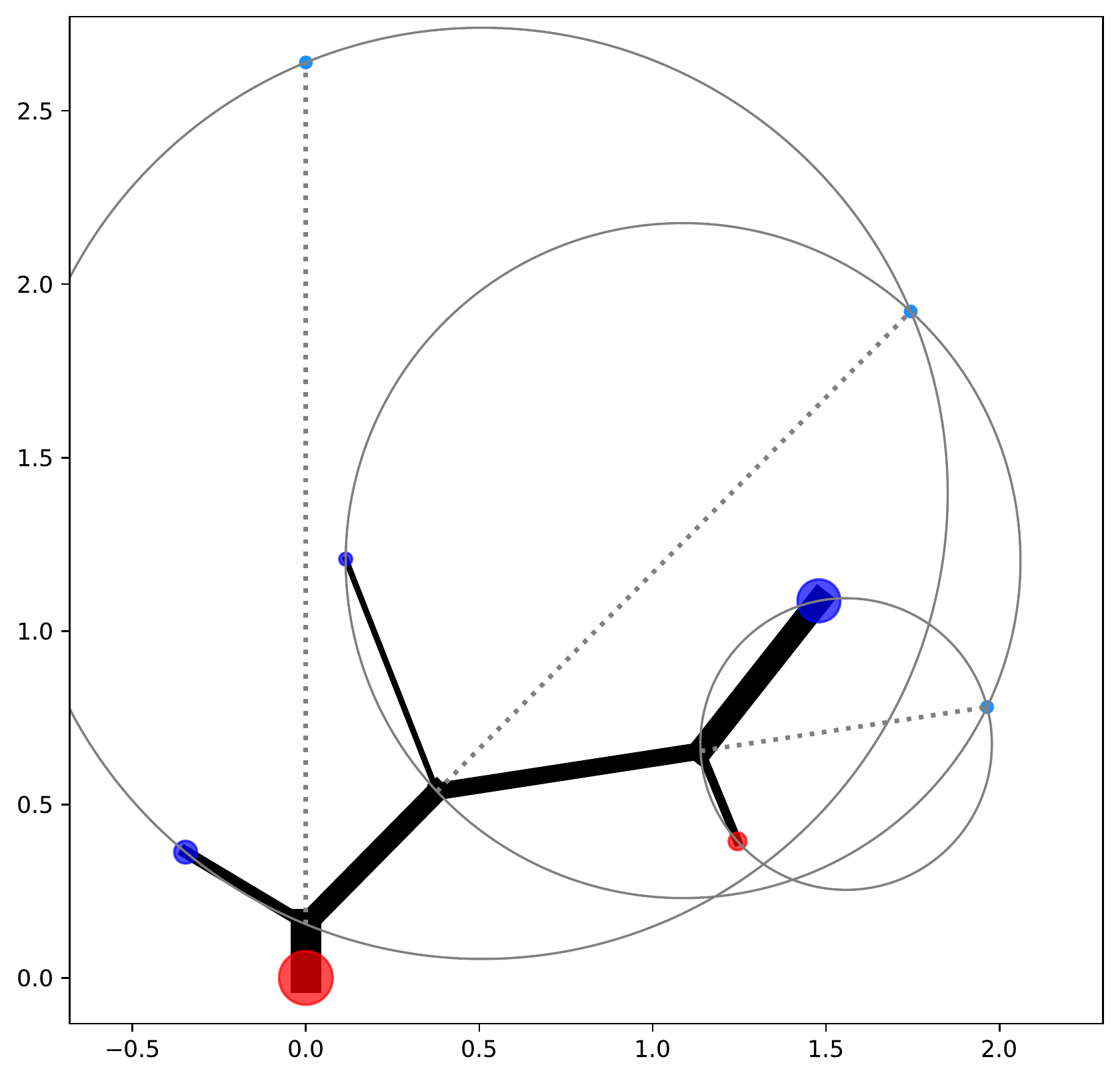}
         \caption{$\ \ \alpha=0.3$}
     \end{subfigure}
		\\
     \vspace{0.15cm}
     	\centering
     \begin{subfigure}[b]{0.3\textwidth}
         \centering
         \includegraphics[width=\textwidth]{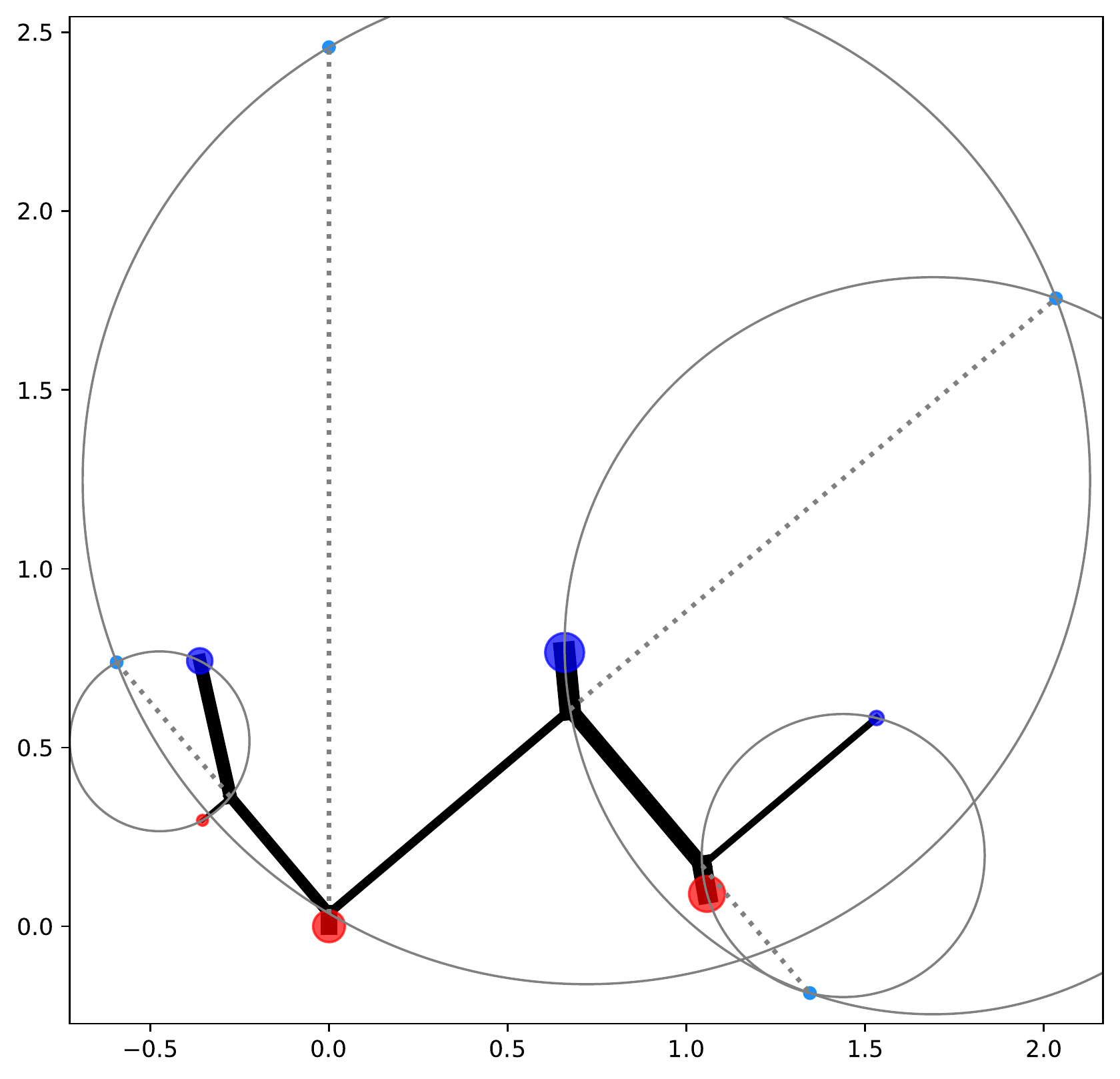}
         \caption{$\ \ \alpha = 0.5$}
     \end{subfigure}
     \hspace{1.5cm}
     \begin{subfigure}[b]{0.3\textwidth}
         \centering
         \includegraphics[width=\textwidth]{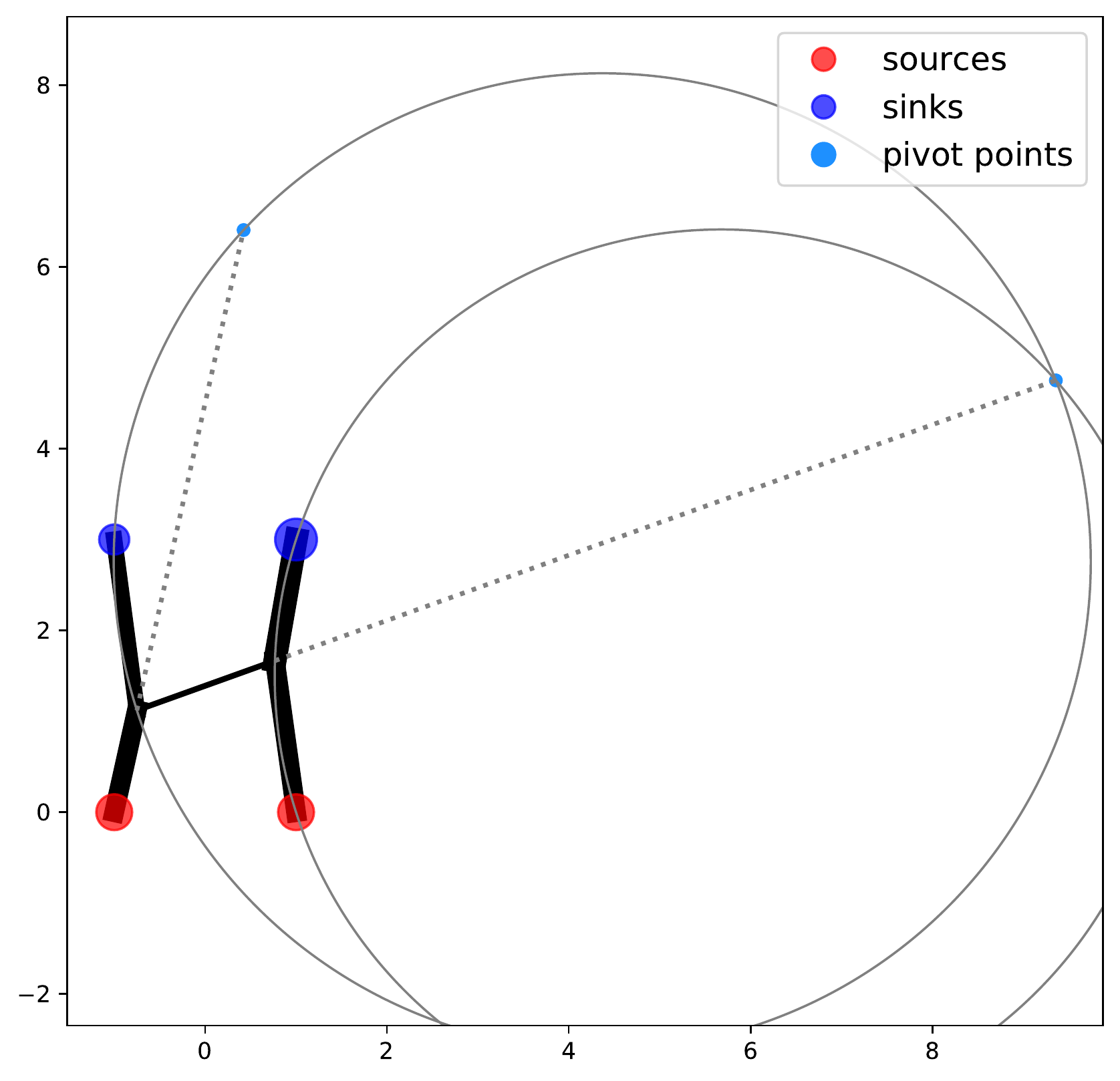}
         \caption{$\ \ \alpha=0.65$}
         \label{geom-solved}
     \end{subfigure}

        \caption{Examples for geometric construction of relatively optimal solutions with fixed topology for BOT problems with multiple sources. The construction is based on the optimal branching angles and uses one pivot circle and pivot point (light blue) per branching point. Sources are shown in red, sinks in blue.}
        \label{geo_ROS}   
        
\end{figure}

Figure~\ref{converg} shows a simple BOT problem with a chosen topology for which no root node can be chosen such that all branchings are symmetric (see Sect.~\ref{subsec:constr}). Choosing for instance the left source as a root node, the branching at $b_1$ is symmetric whereas the one at $b_2$ is asymmetric. However, after having solved the asymmetric branching case analogously to the symmetric one, the relatively optimal solution can now be constructed geometrically as shown in Fig.~\ref{geom-solved}. More examples of the geometric construction for a given topology applied BOT problems with multiple sources are shown in Fig.~\ref{geo_ROS}. 

\paragraph{Continuity of the optimal BP configuration.} The limitations of the geometric construction above are discussed in Sect.~\ref{subsec:constr}. Still, it forms the basis for our theoretical arguments in the paper. Moreover, from the construction of a single branching point based on the branching angles (see Fig.~\ref{1-to-2ybranching}), it can be seen that its optimal position changes continuously w.r.t.~the neighbor positions, the edge flows or $\alpha$. Namely, the Y-, V- and L- branching change continuously as the source $a_0$ is moved around. In addition, the optimal branching angles are given by continuous functions of $k$ and $\alpha$ (cf.~Eq.~(\ref{eq_fandh})), making the construction of the pivot point and pivot circle continuous. By transitivity and based on the optimal substructure in Lem.~\ref{lem:subprobs}, the continuity generalizes also to the construction of larger ROS for a given topology.   

\subsection{Numerical algorithm for geometry optimization}
\label{app:smith}
In this section, we provide theoretical details and practical experiments for the numerical geometry optimization presented in Sect.~\ref{sec:smith}. We have generalized this approached from the context of the ESTP~\cite{smith1992find} to BOT.

For a given tree topology $T$, the BP configuration is optimized by minimizing the following cost function:
\begin{align}
\label{eq_appcost}
\mathcal{C}(X) = \sum_{(i,j) \, \in \, T} m_{ij}^\alpha 
\left \| x_i - x_j  \right \|_2  ,
\end{align} 
where the $x_i$ for $1 \le i \le n$ are fixed terminals and the $x_i$ for $n+1 \le i \le n+m$ denote the variable branching point positions. All coordinates are collectively summarized by $X$. 

Starting from a non-optimal, non-degenerate BP configuration denoted $X^{(0)}$, e.g.~from a random initialization of the branching points, the algorithm iteratively solves the following \textit{linear} system of equations
\begin{align} \label{eq:iterapp}
x_i^{(k + 1)} = \!  \! \sum_{j \, : \, (i,j) \in T}  m_{ij}^\alpha 
\frac{x_j^{(k+1)}}{\vert x_i^{(k)} - x_j^{(k)}  \vert} 
 \  \Bigg  /  \sum_{j \, : \, (i,j) \in T}  \frac{m_{ij}^\alpha}{\vert x_i^{(k)} - x_j^{(k)}  \vert}, \hspace{0.4cm}\text{for } n+1 \le i \le n+m.
\end{align} 
In essence, this is an iteratively reweighted least squares (IRLS) approach~\cite{chartrand2008iteratively}. To see this, let us rewrite the cost function in Eq.~(\ref{eq_appcost}) into pseudo-quadratic from:
\begin{align}
\label{eq_app_q_form}
\mathcal{C}(X) = \sum_{(i,j) \, \in \, T} m_{ij}^\alpha 
\vert x_i - x_j  \vert^{(\beta-2) + 2} = \sum_{(i,j) \, \in \, T} \underbrace{m_{ij}^\alpha 
\vert x_i - x_j  \vert^{\beta-2}}_{=: w_{ij}(X)} \vert x_i - x_j  \vert^2 \, ,
\end{align}
with $\beta = 1$ for BOT. Clearly, the weights $w_{ij}$ depend on the branching point positions. However, the idea of IRLS is to insert the coordinates $X^{(k)} = \{x_i^{(k)} \}$ at $k$-th iteration into $w_{ij}$ to obtain a truly quadratic form:
\begin{align*}
Q^{(k)}(X) &= w_{ij}(X^{(k)}) \vert x_i - x_j  \vert^2 =
\sum_{ (i,j) \, \in \, T} m_{ij}^\alpha
\frac{ \vert x_i - x_j  \vert^2}{ \vert x_i^{(k)} - x_j^{(k)}  \vert} \, ,
\end{align*}
where we have plugged in $w_{ij}(X^{(k)})$ and $\beta = 1$. Indeed, minimizing this quadratic form yields $X^{(k+1)}$ as in  Eq.~(\ref{eq:iterapp}). The updated coordinates $X^{(k+1)}$ are then plugged into $w_{ij}(X)$ during the next iteration. This iterative updating of the weights $w_{ij}$ gives IRLS its name.    

\citet{smith1992find} proved in detail that this iterative solver converges to the minimum cost BP configuration. The reasoning in~\cite{smith1992find} is based on the following key argument: As $X^{(k+1)}$ is the minimizer of $Q^{(k)}(X)$, surely $Q^{(k)}(X^{(k)}) \ge Q^{(k)}(X^{(k+1)})$. Together with the fact that $\mathcal{C} (X^{(k)}) = Q^{(k)}(X^{(k)})$, we have
\begin{align*}
\mathcal{C}(X^{(k)}) &= Q^{(k)}(X^{(k)}) \ge Q^{(k)}(X^{(k+1)}) \\
&= \sum_{ (i,j) \, \in \, T: } m_{ij}^\alpha
\frac{\big(  \vert x_i^{(k)} - x_j^{(k)}  \vert + 
 \vert x_i^{(k+1)} - x_j^{(k+1)}  \vert -  \vert x_i^{(k)} - x_j^{(k)}  \vert  \big)^2}{ \vert x_i^{(k)} - x_j^{(k)}  \vert} \\
&= \mathcal{C}(X^{(k)}) + 2 [\mathcal{C}(X^{(k+1)}) - \mathcal{C}(X^{(k)}) ] \\
&\hspace{1.9cm} + \sum_{(i,j) \, \in \, T} m_{ij}^\alpha 
\frac{ \big( \vert x_i^{(k+1)} - x_j^{(k+1)}  \vert -  \vert x_i^{(k)} - x_j^{(k)}  \vert \big)^2}{ \vert x_i^{(k)} - x_j^{(k)}  \vert} \ .
\end{align*} 
Since the sum in the last expression is clearly non-negative, the inequality above implies that $\mathcal{C}(X^{(k)}) \ge \mathcal{C}(X^{(k+1)})$. This means that the cost of the BP configuration decreases with each iteration which, as shown in~\cite{smith1992find}, implies that the iterations defined by Eq.~(\ref{eq:iter}) converge to an absolute minimum of $\mathcal{C}(X)$.

As can be seen from the above derivation, the edge flows $m_{ij}^\alpha$ are constant coefficients which do not complicate the considerations in~\cite{smith1992find}. Consequently, all arguments presented there can be directly transferred from ESTP to BOT. For a detailed discussion on the complexity of the algorithm and suitable convergence criteria, we refer the reader to the work of Smith and only briefly state the results here. Based on an analytically tractable example, Smith claims that the algorithm requires at most $O(N/\epsilon)$ iterations to converge to a solution of the ESTP whose cost is within $\epsilon$ of the optimal cost. As a convergence criterion, Smith suggests to stop the iteration when the angles at all branching points are sufficiently close to the optimal angle conditions. In order to be able to apply the algorithm also to trees with higher-degree branching points where the optimal angle conditions are a necessary but not sufficient condition for the cost minimum, the experiments in this paper use a different criterion. The algorithm is considered to have converged if from one iteration to the next the relative cost improvement $\big(\mathcal{C}(X^{(k)}) - \mathcal{C}(X^{(k+1)}) \big) / \mathcal{C}(X^{(k+1)})$ has dropped below a certain threshold. The BP optimization routine for a given tree topology is summarized in Alg.~\ref{alg:bp-optim}. Note that, in practice, the denominators $\vert x_i^{(k)} - x_j^{(k)}  \vert$ in Eq.~(\ref{eq:iterapp}) are clipped to $10^{-7}$ to avoid numerical instabilities. 

\begin{algorithm}
\caption{BP optimization routine} \label{glo-bp-opt}
\label{alg:bp-optim}
\begin{flushleft}
\textbf{Input:} threshold $\eta$, tree topology $T$, BOT problem (terminal positions $x_{1:n}$, supplies/demands $\mu$, $\alpha$)  \\
\textbf{Output:} numerical minimal cost BP configuration  
\end{flushleft}
\begin{algorithmic}[1]
\State $F \gets$ \textit{get\_all\_edge\_flows}($T,\mu$) \hfill {$\triangleright \ $ uniquely determined
from flow constraints}
\State $x_{n+1:n+m} \gets$ randomly initialized. 
\State $C_{old} \gets \infty$ 
\State $C \gets$ \textit{BOT\_cost}$(T,x_{1:n+m},F, \alpha)$ 
\While {$\frac{\mathcal C_{old} - \mathcal{C}}{\mathcal C} > \eta$} 
\State $C_{old} \gets C$ 
\State $x_{1:n+m} \gets $ \textit{update\_BPs}($x_{1:n+m},F, \alpha$)
\hfill {$\triangleright \ $ solves linear system in Eq.~(\ref{eq:iterapp})}
\State $C \gets$ \textit{BOT\_cost}$(T,x_{1:n+m},F, \alpha)$
\EndWhile
\State \textbf{return} $x_{1:n+m}$
\end{algorithmic} 
\end{algorithm}

\paragraph{BOT with costs scaling non-linearly with the edge length.} 
Let us briefly consider BOT with a modified cost function of
\begin{align}
\label{eq_mod_cost}
\mathcal{C}(X) =  \sum_{(i,j) \, \in \, T} m_{ij}^\alpha 
\left \| x_i - x_j  \right \|^\beta_2 \,  
\end{align}
including one additional parameter $\beta$, regulating the scaling of the cost function w.r.t.~edge lengths. For $\beta \ge 1$ the cost function as a function of the branching points is still convex, as $x \mapsto x^\beta$ is convex and increasing and the Euclidean norm is convex. Thus, it has a unique minimum. As can be seen from Eq.~(\ref{eq_app_q_form}), the IRLS scheme completely absorbs the $\beta$ into the weights $w_{ij}$. Consequently, the geometry optimization as presented here is readily applicable also to the modified cost in Eq.~(\ref{eq_mod_cost}). 
However, for $\beta>1$ we are not aware of any theoretical convergence guarantee to the minimum, therefore, further investigation is required. Though, in the special case of $\beta=2$, it is clear that the geometry optimization can be solved to global optimality in just a single iteration.

\begin{figure}
\centering 
\hspace*{0.cm}
\includegraphics[scale=0.4]{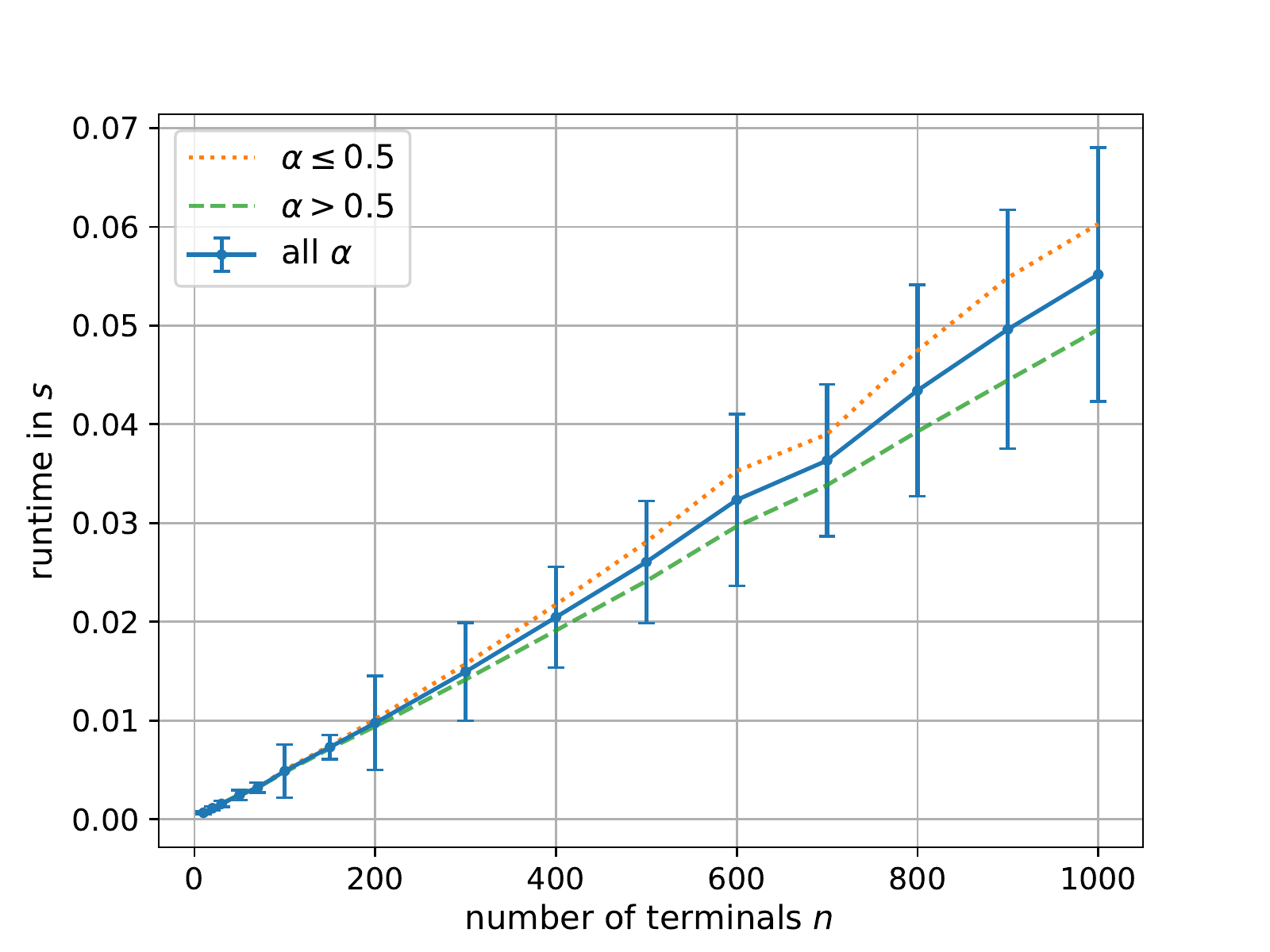} 
\caption{Average runtime of BP optimization routine applied to each 1000 random BOT problems of different size. Average of all problems with $\alpha \le 0.5$ in orange (dotted) and $\alpha > 0.5$ in green (dashed).}  
\label{smith}
\end{figure}

\paragraph{Performance test of numerical BP optimization.} In order to evaluate how long the BP optimization for a given topology takes, we have randomly generated 1000 BOT problems for different number of terminals using Alg.~\ref{alg:random}. For each of these problems, we have applied the BP optimization routine, given a uniformly sampled full tree topology. Regarding the generation of problem setups, note that BOT solutions and problems are invariant under rescaling of the total demand and supply as well as under global rescaling of the coordinates, see Eq.~(\ref{eq_cost}). WLOG, these scales are chosen to be 1. Figure~\ref{smith} shows the average runtime in seconds plotted against the number of terminals. For BOT problems with 10 to 1000 terminals our efficient C++ implementation of the geometry optimization takes just a fraction of a second. The relatively large error bars, indicating the standard deviation, are not due to an insufficient sample size but due to the natural run time variability for different problems that exists independent of the $n$-dependence. For instance, plotting the average runtime of all problems with $\alpha \le 0.5$ and $\alpha > 0.5$ separately reveals that the optimization on average requires more time for problems with higher $\alpha$. The reason for this was not investigated further, but we suspect that for larger $\alpha$ on average more V- and L-branchings occur which may require more iterations than Y-branchings for convergence. For this and all following experiments we chose the convergence threshold $\eta$ to be $10^{-6}$.

\begin{figure}
\centering 
\hspace*{0.cm}
\includegraphics[scale=0.44]{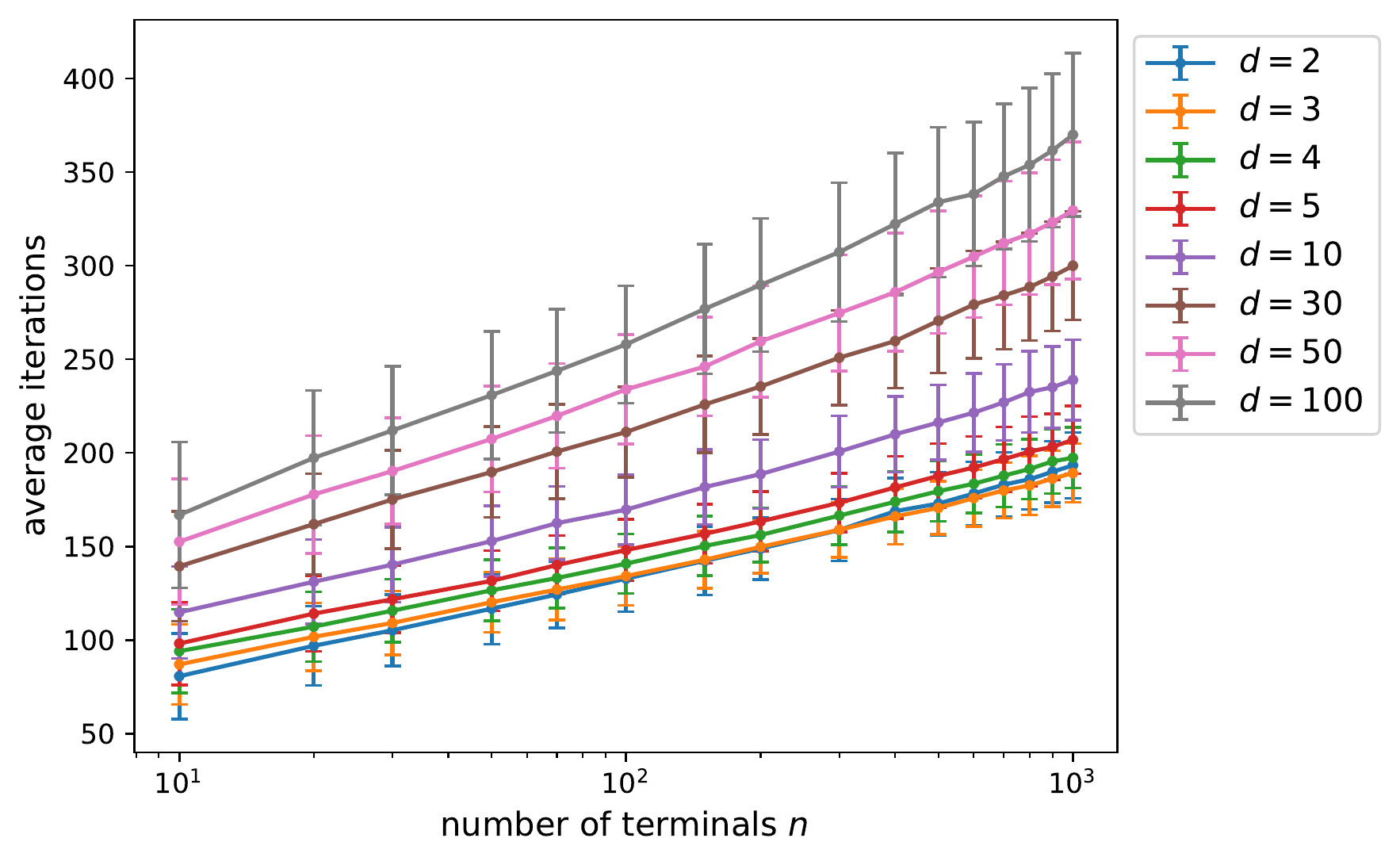} 
\caption{Average number of iterations required by the BP optimization routine until convergence for each 1000 random BOT problems of different sizes $n$ and different dimensionality $d$. The x-axis is in log-scale. The plot suggests that, for the investigated regime of $n$ and $d$, the number of iterations on average scales like $\log(n)$, though one cannot generalize this statement.}  
\vspace*{-0.3cm}
\label{smith_scaling}
\end{figure}

\paragraph{Scaling of the geometry optimization}
A single iteration of the geometry optimization, i.e. solving the linear system defined by Eq.~(\ref{eq:iter}) once, takes $O(nd)$ operations for a problem with $n$ terminals in $d$ spatial dimensions. The elimination scheme used in our efficient C++ implementation is based on the ``elimination on leaves of a tree'' found in~\cite{smith1992find}. From Eq.~(\ref{eq:iter}), one can easily see that the geometry optimization parallelizes over the spatial dimensions, as the linear system of the form $Ax = b$ shares the same matrix $A$ across the different dimensions and only the $b$ is different. Although a single iteration is of order $O(nd)$, it is a priori not clear how many iterations are required until convergence is reached. Paralleling the setup of Fig.~\ref{smith}, we have conducted an experiment where we report the number of iterations, for a batch of BOT problems. We performed the same experiment in Euclidean space of different dimensions ($d \in \{3,4,5,10,30,100 \}$). Figure~\ref{smith_scaling} shows the number of iterations until convergence plotted against the number of terminals. The plot suggests that, for the investigated regime of $n$ and $d$, the number of iterations on average scales like $\log(n)$. However, this is merely an empirical observation and theoretical investigations are an interesting subject for future research. 

\begin{algorithm}
\caption{Random BOT problem generation}
\label{alg:random} 
\begin{flushleft}
\textbf{Input:} number of terminals $n$ \\
\textbf{Output:} BOT problem with $n$ terminals located in $[0,1]\times[0,1]$,  total supply and demand equal to 1 
\end{flushleft} 
\begin{algorithmic}[1]
\State $\alpha \sim$ Uniform($[0,1]$) \hfill {$\triangleright \ $ sample uniformly from $[0,1]$ } 
\State $n^+ \sim$ Uniform($\{1,2,...,n-1\}$) \hfill {$\triangleright \ $ number of sources}
\State $n^- \gets n - n^+$ 
\State $\mu^+_{1:n^+} \sim$ Uniform($[0,1]$, size=$n^+$) \hfill {$\triangleright \ $ array of supplies}
\State $\mu^-_{1:n^-} \sim$ Uniform($[0,1]$, size=$n^-$) \hfill {$\triangleright \ $ array of demands}
\State $a_{1:n} \sim $ Uniform($[0,1]$, size=($n$,2)) \hfill{$\triangleright \ $ terminal coordinates}
\State \textbf{return} $a_{1:n}, \ \mu^+/\mathrm{sum}(\mu^+), \ \mu^-/\mathrm{sum}(\mu^-), \ \alpha$
\end{algorithmic} 
\end{algorithm}

\subsection{Greedy randomized heuristic for topology optimization}
\label{app:heuristic}

\begin{figure}
     \centering
     \begin{subfigure}[b]{0.29\textwidth}
         \centering
         \includegraphics[width=\textwidth]{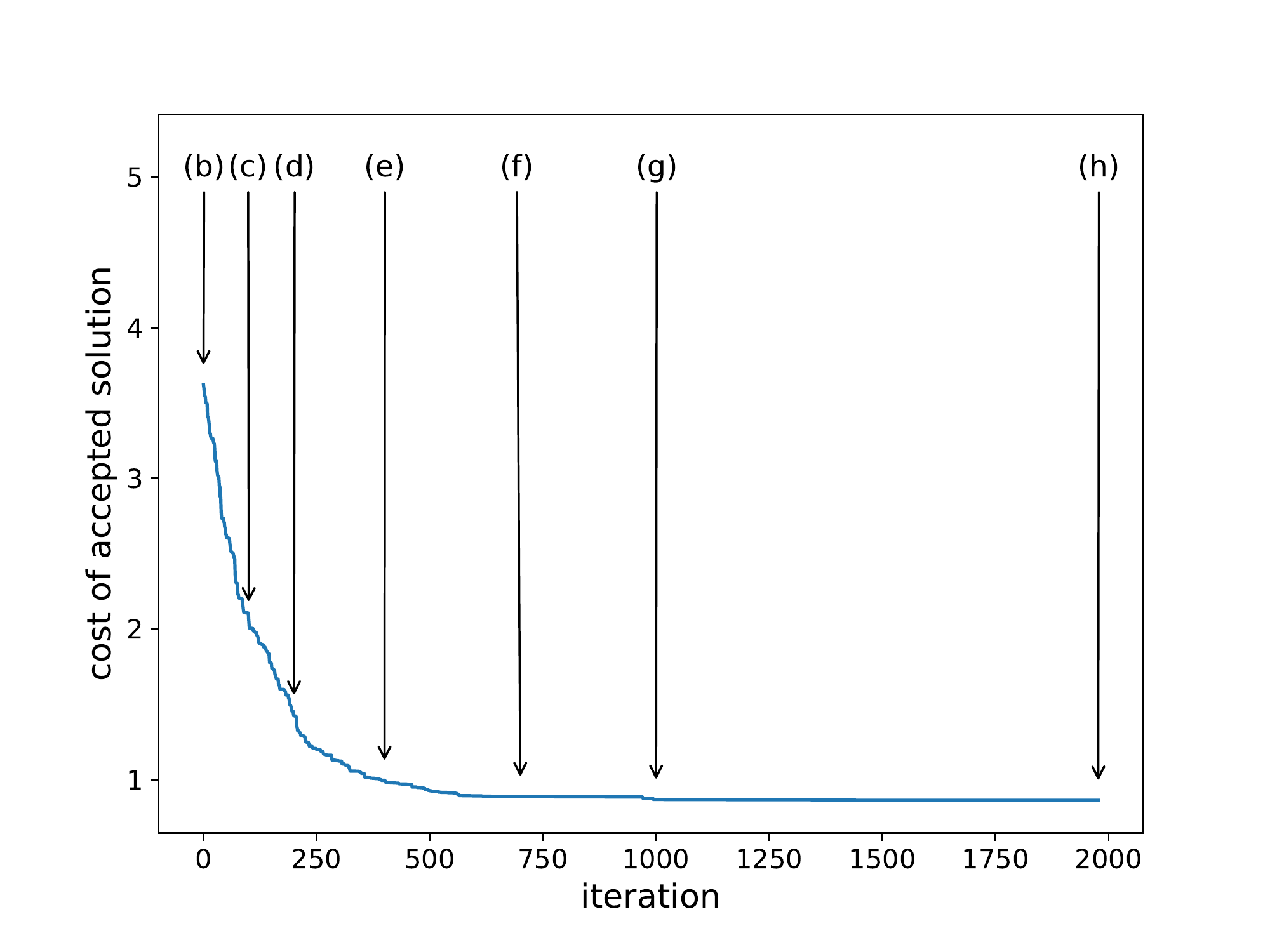}
         \caption{\label{fig:decrease}Decreasing cost vs. iterations}
     \end{subfigure}
     \hfill
     \begin{subfigure}[b]{0.23\textwidth}
         \centering
         \includegraphics[width=\textwidth]{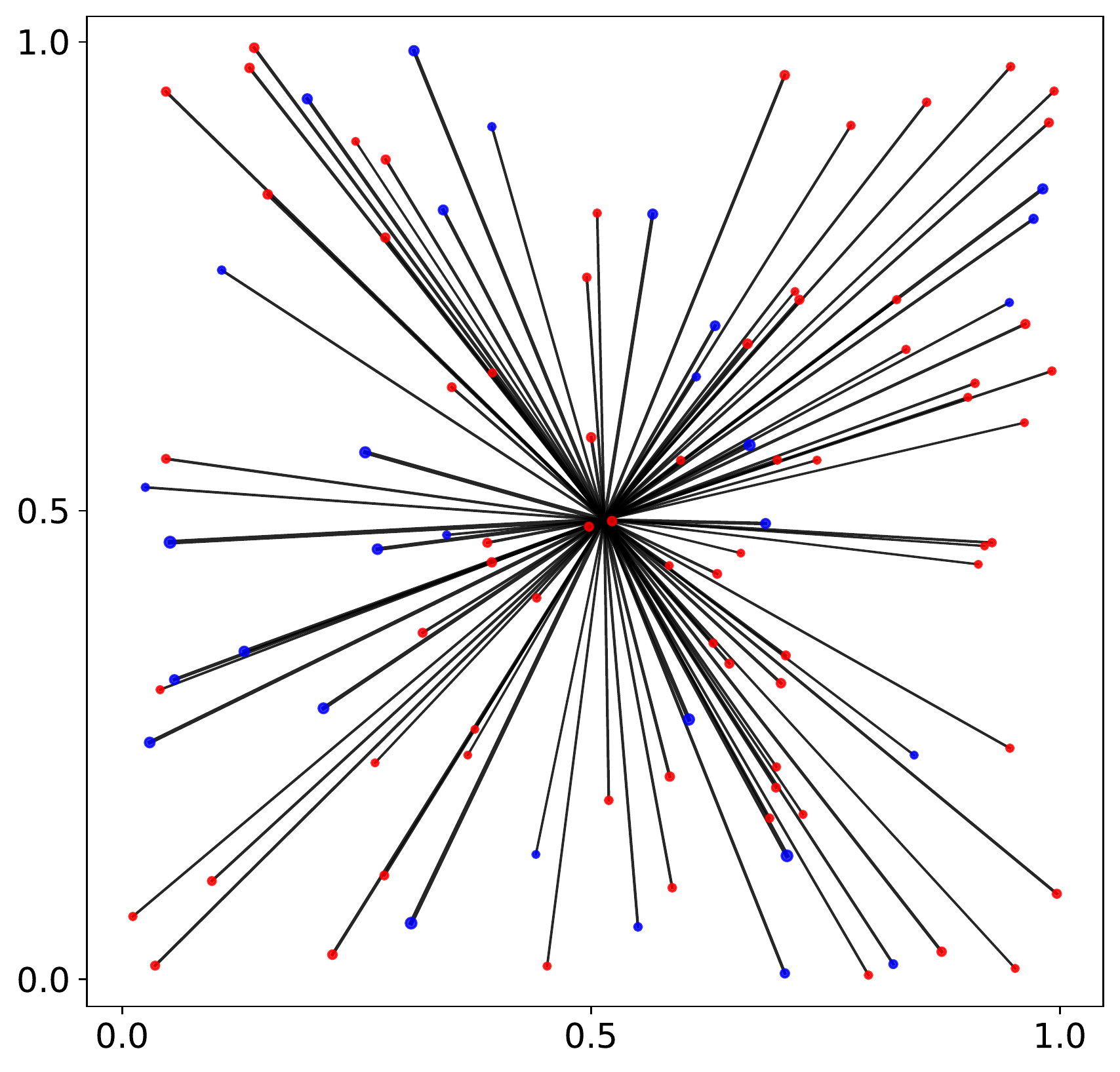}
         \caption{\label{fig:iter0}0 iterations.}
     \end{subfigure}  
     \hfill
	      \centering
     \begin{subfigure}[b]{0.23\textwidth}
         \centering
         \includegraphics[width=\textwidth]{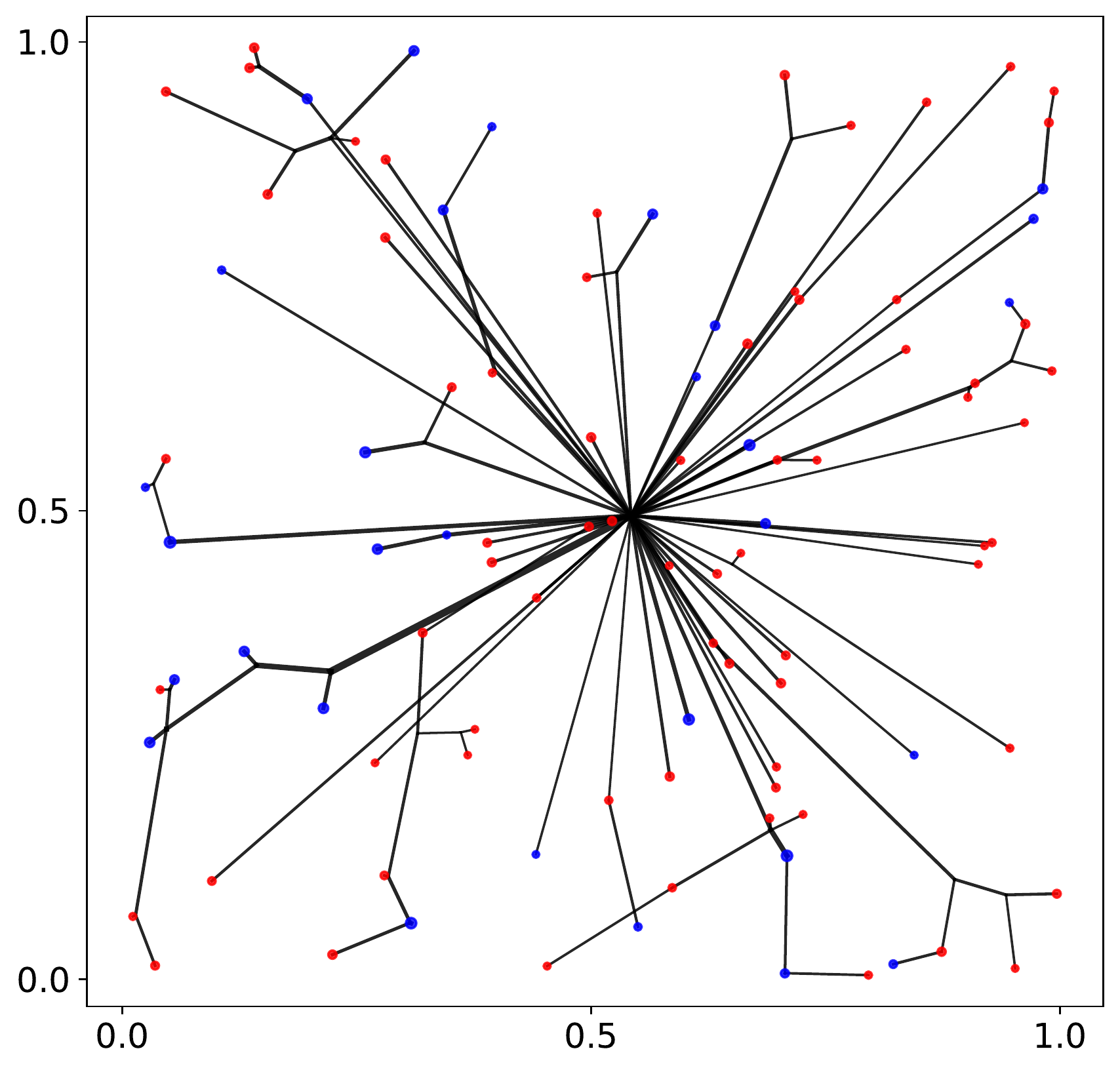}
         \caption{100 iterations.}
     \end{subfigure}
     \hfill
     \begin{subfigure}[b]{0.23\textwidth}
         \centering
         \includegraphics[width=\textwidth]{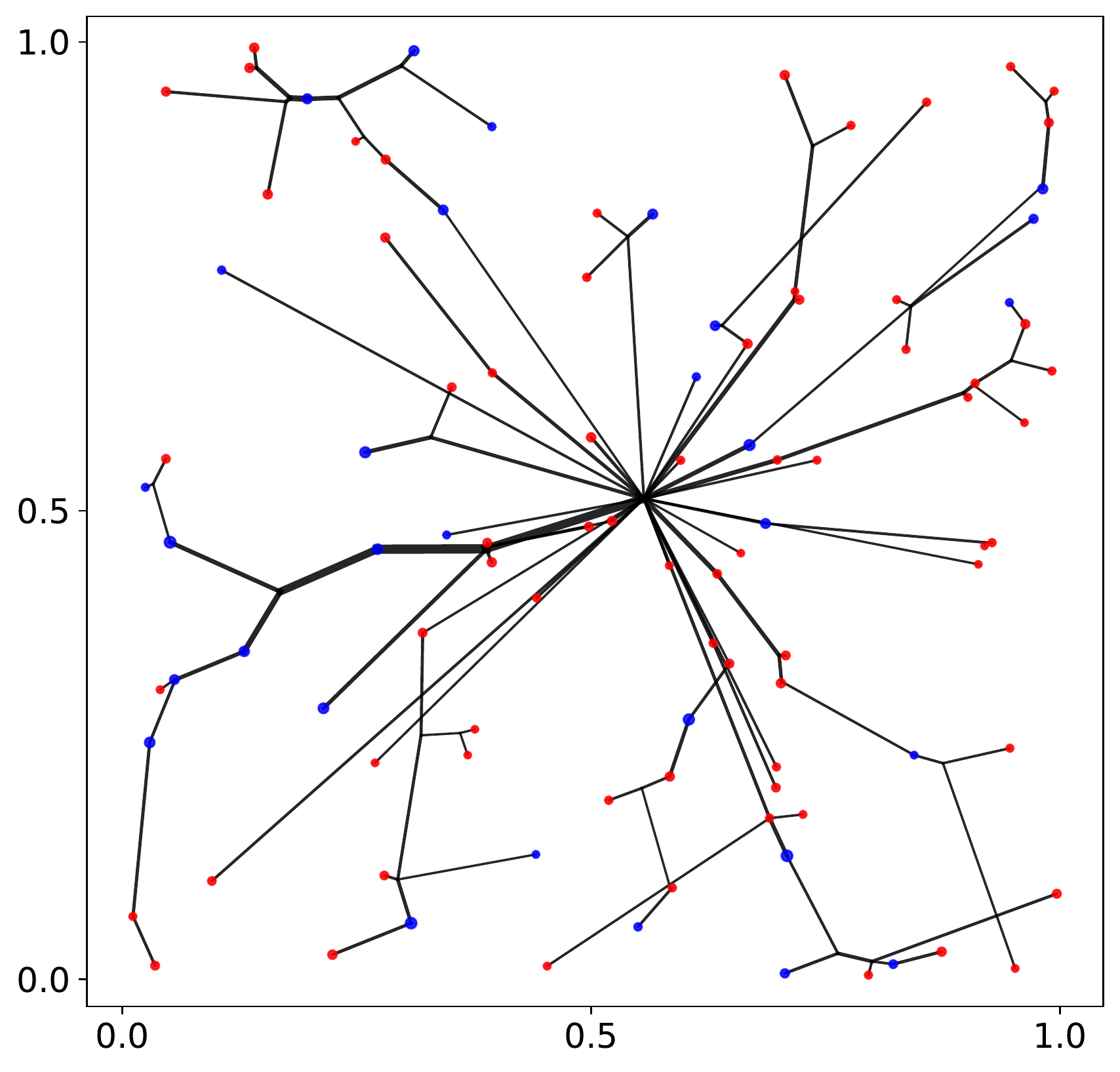}
         \caption{200 iterations.}
     \end{subfigure}
	 \\
     \vspace{0.45cm}
     \begin{subfigure}[b]{0.23\textwidth}
         \centering
         \includegraphics[width=\textwidth]{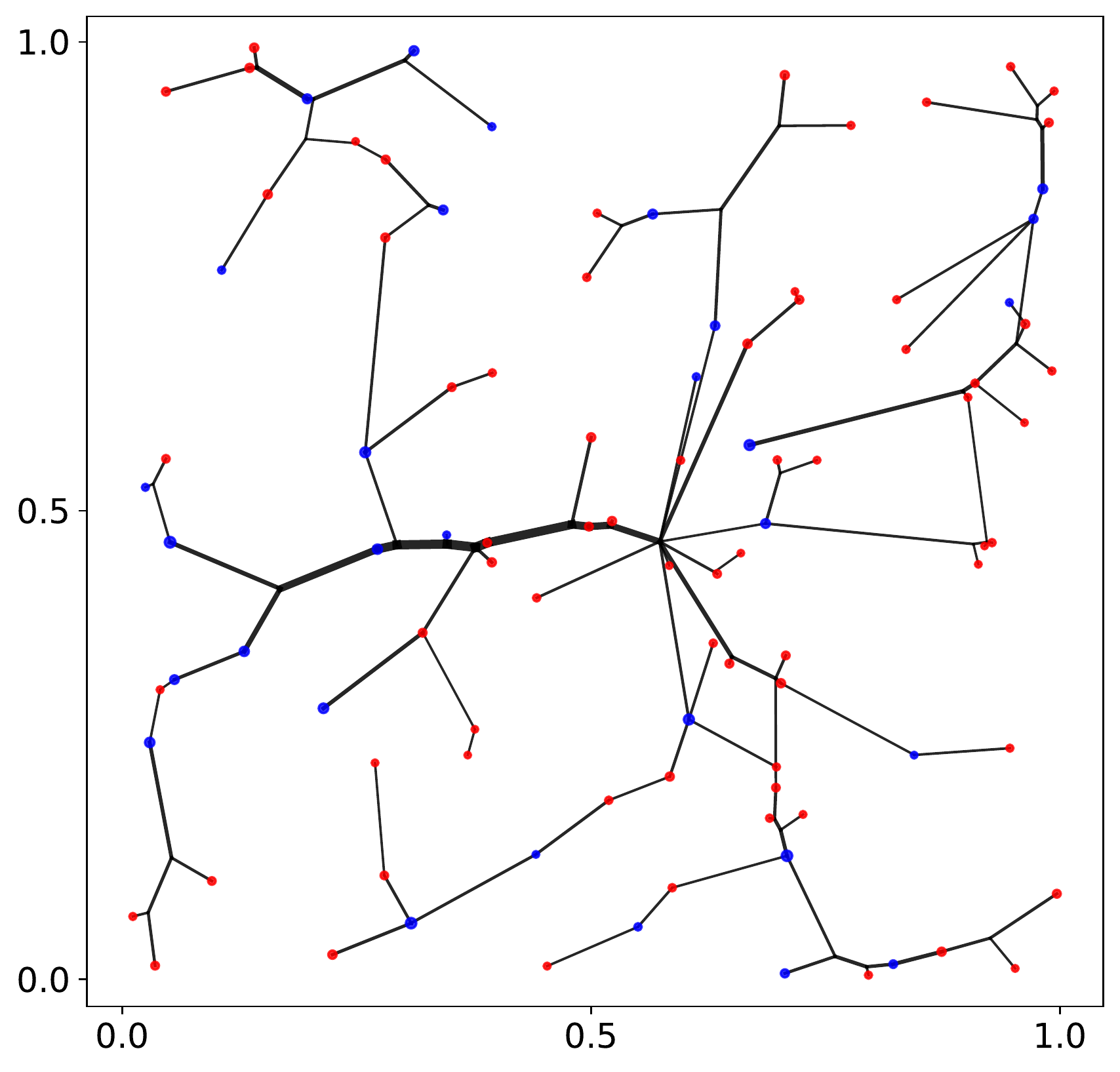}
         \caption{400 iterations.}
     \end{subfigure}  
    \hfill
	      \centering
     \begin{subfigure}[b]{0.23\textwidth}
         \centering
         \includegraphics[width=\textwidth]{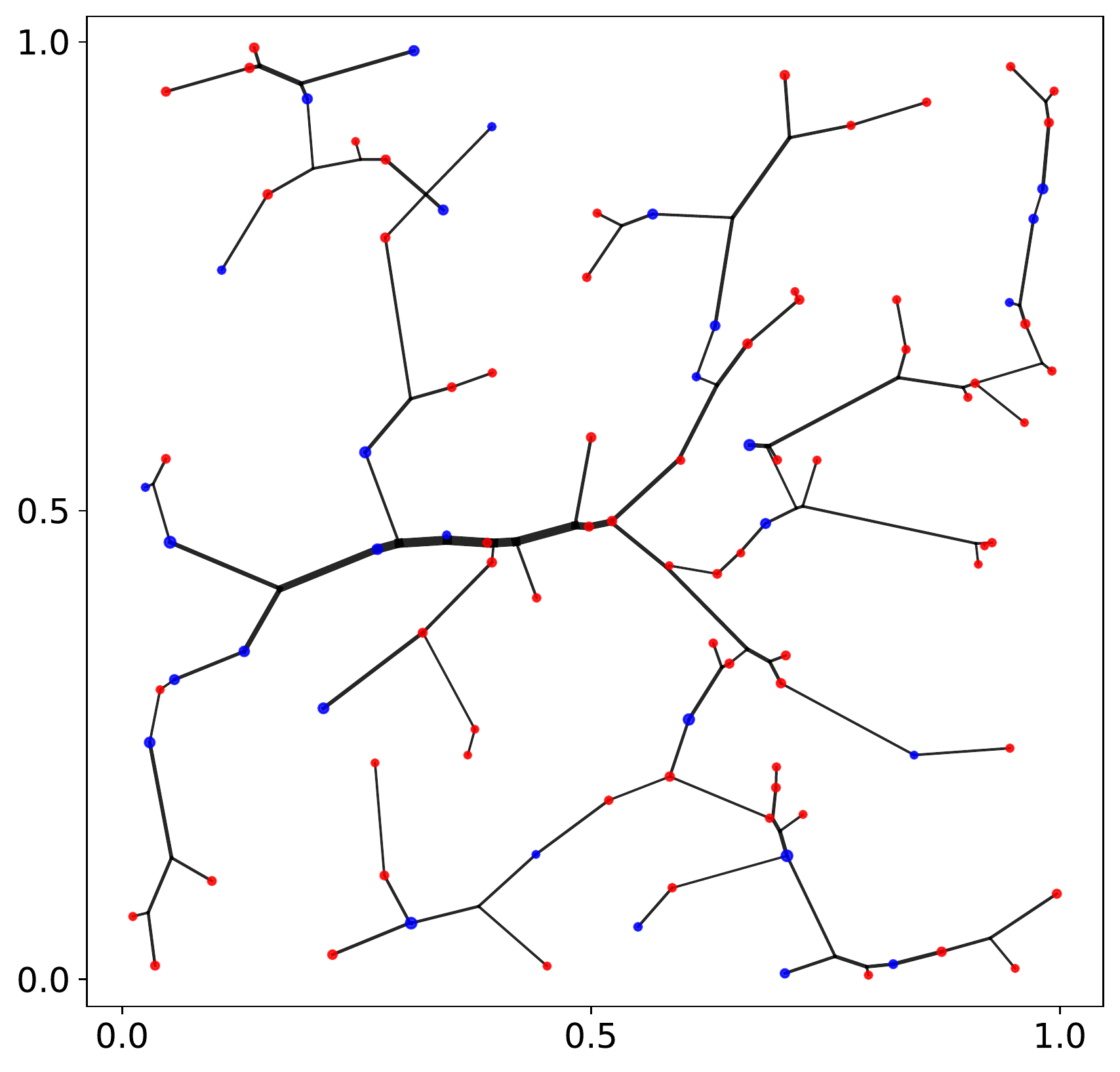}
         \caption{700 iterations.}
     \end{subfigure}
     \hfill
     \begin{subfigure}[b]{0.23\textwidth}
         \centering
         \includegraphics[width=\textwidth]{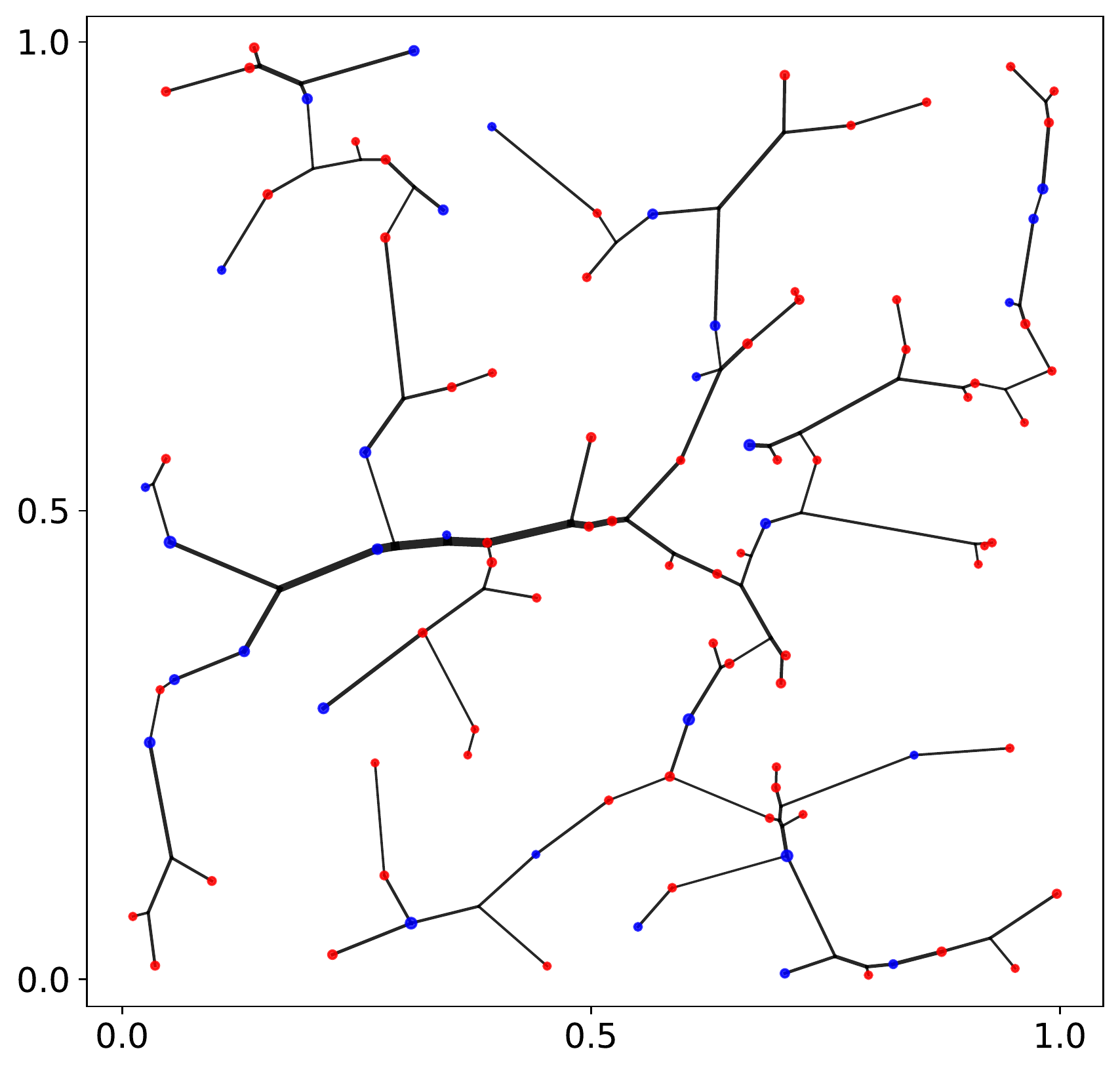}
         \caption{1000 iterations.}
     \end{subfigure}
	\hfill
     \begin{subfigure}[b]{0.23\textwidth}
         \centering
         \includegraphics[width=\textwidth]{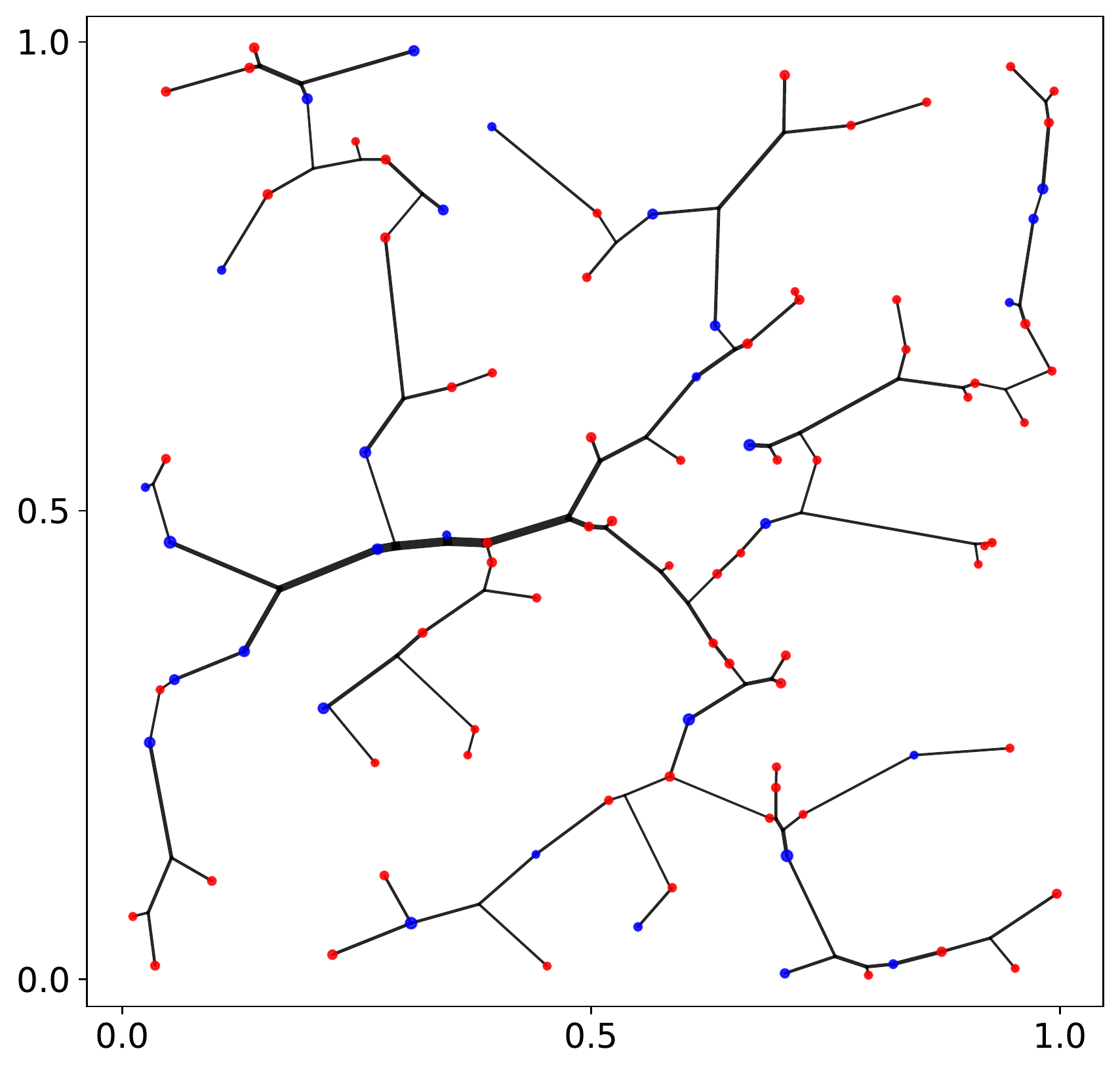}
         \caption{1977 iterations.}
     \end{subfigure}
     
        \caption{Our greedy BOT solver applied to a problem with 100 terminals (sources in red and sinks in blue), $\alpha = 0.58$.}
        \label{sim-ann2}   
        
\end{figure} 

In this section, we give the details of the greedy randomized heuristic presented in Section~\ref{sec:sim-ann} and show the results of some additional experiments conducted with it.

Starting from a tree topology $T$, a uniformly sampled edge $\hat e$ is removed. The node in the smaller connected component $\ell$ is connected via a new branching point to an edge in the other component. This edge is sampled according to $p(e) \propto \exp(-d(e, \ell)^2 / d_{min}^2)$, where $d(e, \ell)$ is the distance between an edge $e=(i,j)$ and node $\ell$, defined by 
\begin{align} \label{eq:e-distance}
d(e, \ell) = \underset{\lambda \in [0,1]}{\min} \ \left \| [x_i + \lambda (x_j - x_i) ] - \ell \right \|_2 \ 
\end{align}
and $d_{min}$ is the distance to the closest considered edge. The resulting new tree topology $T_{new}$ is accepted if it decreases the cost compared to the previous topology $T$ and the above procedure is iterated. This greedy optimization strategy is summarized in Alg.~\ref{alg:sim-ann} below.

\begin{algorithm}
\caption{Greedy randomized heuristic (zero temperature limit)}
\label{alg:sim-ann} 
\begin{flushleft}
\textbf{Input:} BOT problem $P$, tree $T$ interconnecting terminals $a_{1:n}$ (with help of BPs) \\
\textbf{Output:} heuristic BOT solution 
\end{flushleft}
\begin{algorithmic}[1]
\State $B \gets$ \textit{optimize\_BP\_configuration}($T, P$) \hfill {$\triangleright \ $ returns coordinates of BPs and terminals, see Alg.~\ref{alg:bp-optim}}
\State $\mathcal{C} \gets $ \textit{BOT\_cost}$(T,B,P)$
\State $E \gets$ list(T.edges())
\While{$E$ not empty}
\State $S \gets$ Store current state $(T,B)$
\State $\mathcal{C}_{old} \gets \mathcal{C}$ 
\State $\hat e \sim $ Uniform($E$) \hfill {$\triangleright \ $ sample uniformly from $E$}
\State $E$.remove($e$)
\State $T$.remove\_edge($\hat e$)
\State $C_b \gets$ subgraph component of $T$ with more nodes
\State $E_b \gets$  list($C_b$.edges)
\State $b \gets$ node of $\hat e$ in $C_b$
\State $\ell \gets$ node of $\hat e$ not in $C_b$
\If {degree($b$) == 2}
\State $n_1,n_2 \gets$ \textit{neighbors}($T, b$)
\State $T$.remove\_node($b$)  \hfill {$\triangleright \ $ remove unnecessary BPs with degree 2}
\State $T$.add\_edge($n_1,n_2$)
\State $E_b$.remove(($n_1,b$), ($n_2,b$))
\EndIf
\State $d \gets$ Initalize array of size \textit{length}($E_b$)
\For {edge $e \in E_b$}
\State $d[e] \gets $ \textit{get\_distance}($e,\ell$)  
\hfill {$\triangleright \ $calculate distance defined by Eq.~(\ref{eq:e-distance})}
\EndFor
\State $d_{min} \gets$ min($d$)
\State $e_c \sim \exp(-d^2/d_{min}^2)$ 
\hfill {$\triangleright \ $ sample edge according to distance kernel}  \label{line_kernel}
\State $T$.add\_node($b_{new}$) \hfill {$\triangleright \ $ Initialize a new BP}
\State $T$.add\_edges($(\ell,b_{new}),(b_{new}, e_c[0]),(b_{new}, e_c[1])$)
\State $B \gets$ \textit{optimize\_BP\_configuration}($T, P$)
\State $\mathcal{C} \gets $ \textit{BOT\_cost}$(T,X,P)$
\If {$ \mathcal{C} < \mathcal{C}_{old} $} 
\State $E \gets$ list(T.edges()) \hfill {$\triangleright \ $ accept new state}
\Else
\State $T,B \gets S$ 
\hfill {$\triangleright \ $ Restore old state $(T,B)$}  
\State $\mathcal C \gets \mathcal{C}_{old}$
\EndIf
\EndWhile
\State \textbf{return} $T, B$
\end{algorithmic} 
\end{algorithm}

Figure~\ref{sim-ann2} shows an example BOT problem with 100 terminals to which our greedy heuristic has been applied. As starting point, the topology with the least possible structure was used, a star-like tree centred around a single BP of degree 100, cf.~Fig.~\ref{fig:iter0}. It can be seen how the topology evolves over less than 2000 iterations to the final solution. Figure~\ref{fig:decrease} shows the decreasing transportation cost $\mathcal{C}$ plotted against the number of iterations. 

\begin{figure}[t]
\centering 
\hspace*{0.cm}
\includegraphics[scale=0.4]{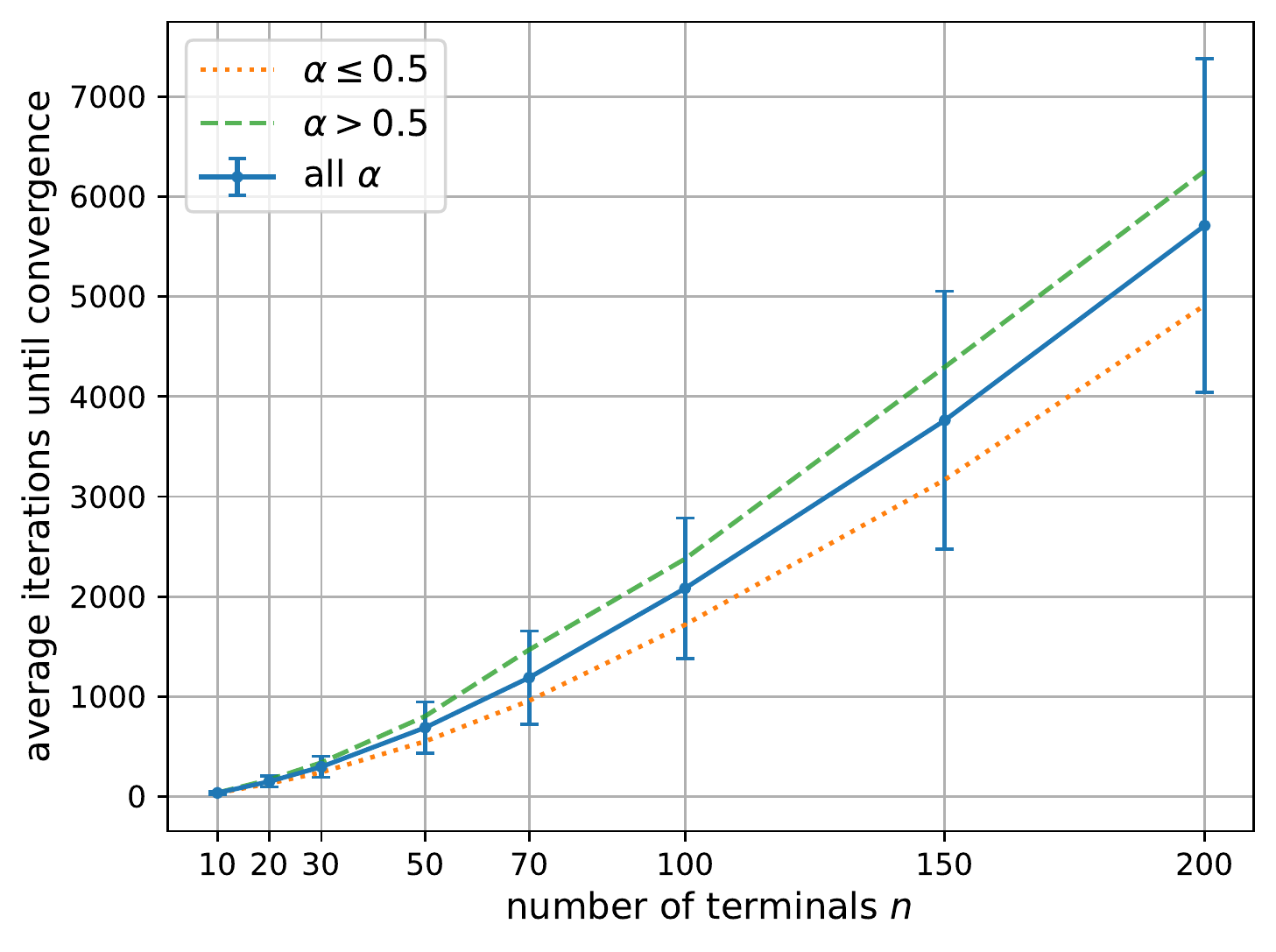} 
\caption{The greedy topology optimization applied to each 150 problems with $n$ terminals: Mean and standard deviation of required iterations plotted vs. $n$ in blue. Average for problems with $\alpha \le 0.5$ in green (dashed) and for problems with $\alpha > 0.5$ in orange (dotted).}  
\label{iter_conv}
\end{figure}

\paragraph{Performance test of heuristic topology optimization.} Finally, in a large experiment, we have investigated how many iterations the greedy heuristic requires on average to converge. For that, the greedy heuristic was applied to a number of BOT problems with $n$ terminals and the number of iterations until convergence were counted. The mean and standard deviation of the required iterations are plotted in Fig.~\ref{iter_conv}. The reason for the relatively large standard deviations are due to intrinsic variation of the sampled problems. To illustrate, for instance, the influence of the $\alpha$-value, the average number of iterations until convergence were plotted separately for all BOT problems with $\alpha \le 0.5$ and $\alpha > 0.5$. We find that the greedy heuristic systematically needs more iterations to converge for $\alpha > 0.5$. 

\begin{figure}
     \centering
     \begin{subfigure}[b]{0.54\textwidth}
         \centering
         \includegraphics[width=\textwidth]{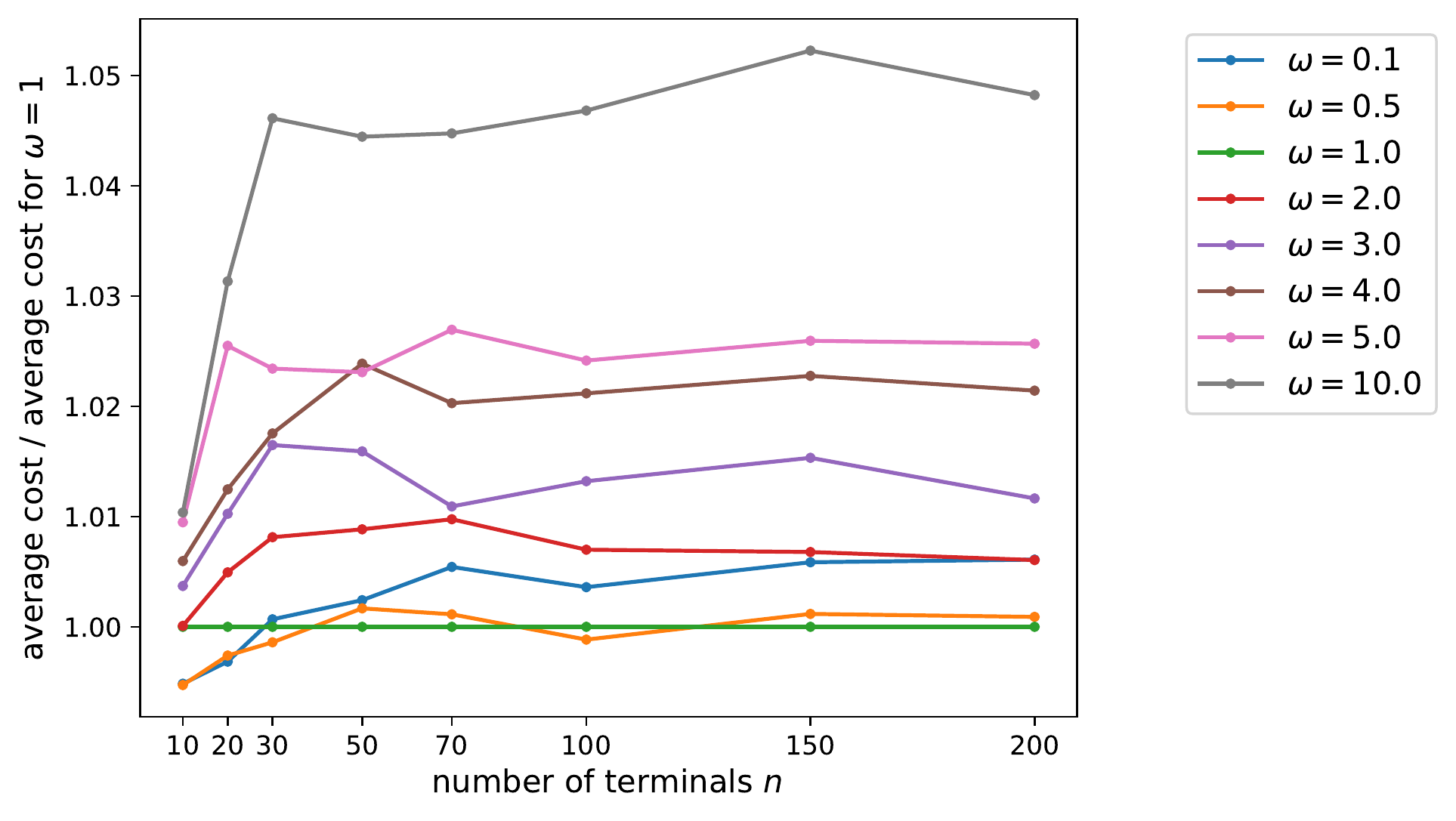}
         \caption{\phantom{ppppppppppp}}
         \label{width_cost}
     \end{subfigure}
     \hspace{0.3cm}
     \begin{subfigure}[b]{0.41\textwidth}
         \centering
         \includegraphics[width=\textwidth]{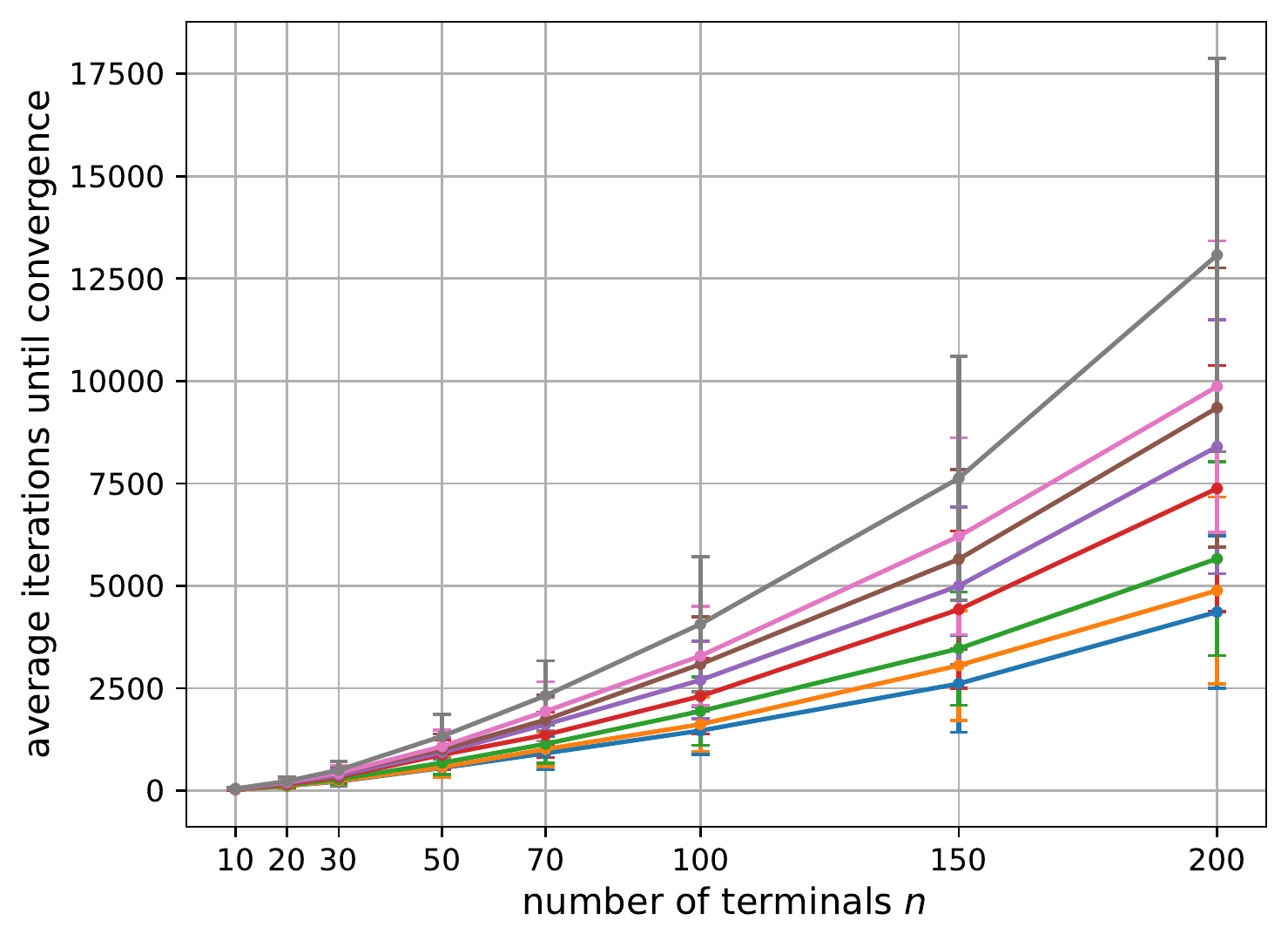}
         \caption{}
         \label{width_iters}
     \end{subfigure}
        \caption{Influence of kernel width in topology optimization based on 150 random problems of various sizes $n$: \textbf{(a)} Ratio of the average cost of the heuristic solution with kernel width factor $\omega$ to the default of $\omega=1$.  \textbf{(b)} Average number of iterations until convergence for different kernel width factors $\omega$. Depending on the kernel width the number of iterations on average scales between $O(n^{1.5})$ and $O(n^{1.7})$.}
        \label{kernel_width}
\end{figure}

\paragraph{Influence of the edge sampling kernel.}
Our heuristic topology optimization involves a number of design choices which may affect its performance. A systematic investigation is beyond the scope of this work and left for future research. However, one hyperparameter of particular interest is the kernel (chosen to be Gaussian) and its width (chosen to be $d_{min}$
), which together define the replacement probability of the edges, see l.~\ref{line_kernel} in Alg.~\ref{alg:sim-ann}. To study its influence on the performance we varied the width of the Gaussian kernel $\exp(-d^2/(\omega d_{min})^2)$ by tuning the parameter $\omega$. Based on 150 random problems of various sizes $n$, we calculated the average cost ratio of the heuristic with different~$\omega$ to the default of $\omega = 1$ (cf. Fig.~\ref{width_cost}). Indeed, Fig.~\ref{width_cost} shows that the default choice of $\omega = 1$ is quite strong and relatively robust given that $\omega = 0.5$ or $\omega = 0.1$ 
work similarly well. Clearly, wider kernels lead to larger (i.e. less local) changes of the topology. At later stages of the algorithm most of these topology changes will be unfavorable, explaining why the algorithm for wider kernels terminates with comparatively less optimal solutions. 
Furthermore, we have investigated the influence of the kernel width on the number of iterations required (cf. Fig.~\ref{width_iters}). Qualitatively, Fig.~\ref{width_iters} confirms that for wider kernels, which encourage exploration, more iterations are required. Fitting a power law of the form $x \mapsto ax^b$ to the curves in Fig.~\ref{width_iters}, one finds that, depending on the kernel, the greedy heuristic on average scales between $O(n^{1.5})$ and $O(n^{1.7})$. Again, this is a purely empirical statement. A careful theoretical analysis to obtain guarantees, also for problem sizes not covered in our experiments, is beyond the scope of this work.

\begin{figure}[H]
\centering 
\hspace*{0.cm}
\includegraphics[scale=0.4]{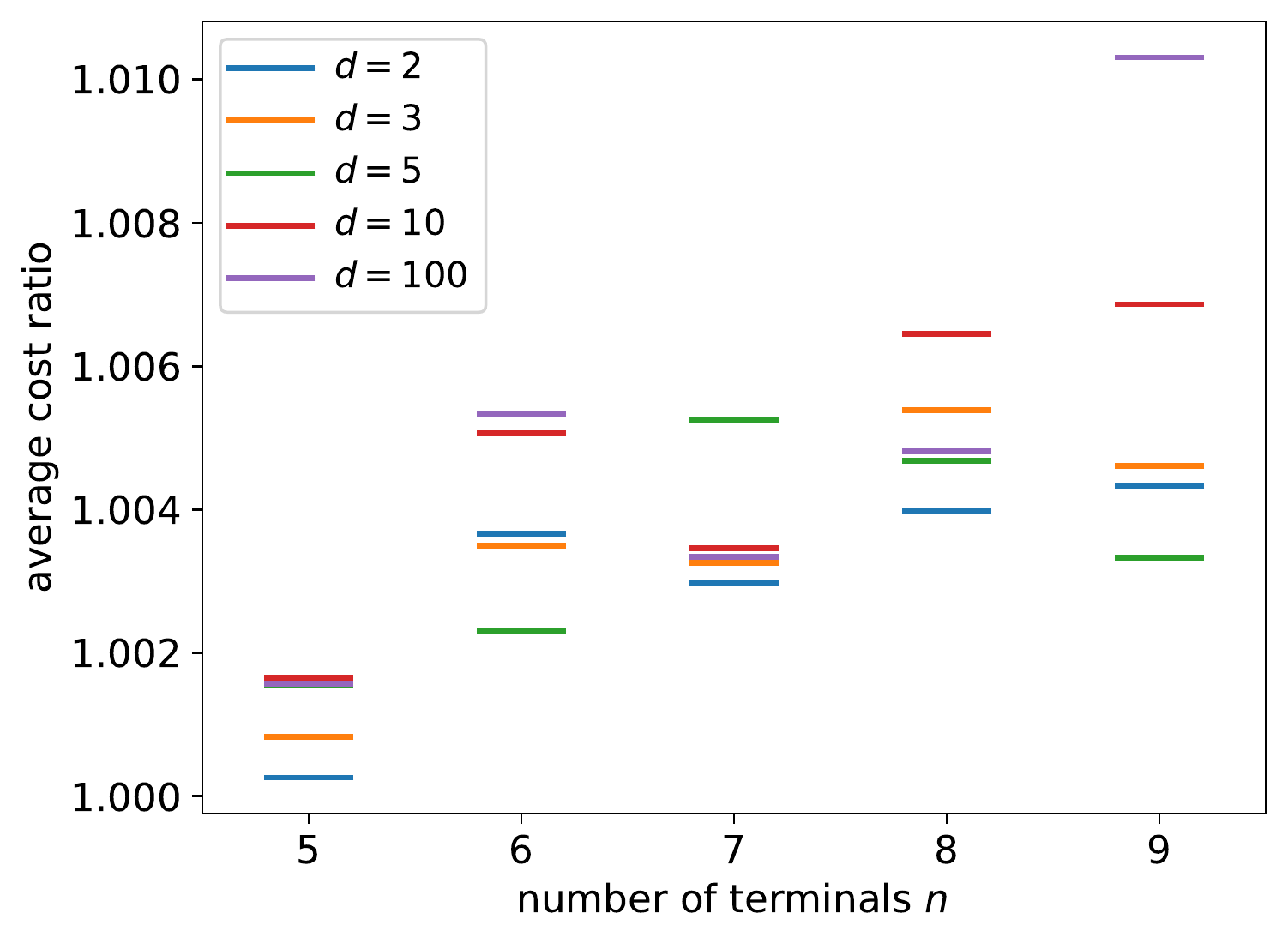} 
\caption{Average cost ratios of our greedy heuristic and brute-force solutions (the closer to 1 the better) for different number of terminals $n$ and dimensions $d$. For each $n$, we uniformly sampled 100 different BOT problems. Though the average cost ratio increases slightly with $n$, our approximate BOT solver compares very well against the ground truth solutions, independently of the dimensionality~$d$.}  
\label{bf_all}
\end{figure}

\paragraph{Greedy topology optimization vs. brute-force search for higher-dimensional BOT.} 
Both the numerical geometry optimization and the greedy algorithm for the topology optimization presented in Sect.~\ref{sec:practical} are readily applicable to BOT problems in $\mathbb{R}^d$. 
Paralleling the experiment presented in Fig.~\ref{MCratio}, we have compared our heuristic topology optimization against brute-forced solutions for each 100 problems of size $n=5$ to $n=9$ and spatial dimension $d$. Although Fig.~\ref{bf_all} suggests that the average cost ratio increases slightly with $n$, our approximate BOT solver again compares very well against the ground truth solutions, independently of the dimensionality $d$.

\end{document}